\documentclass{article}

\PassOptionsToPackage{numbers}{natbib}
\usepackage[main,preprint]{neurips_2026}

\usepackage{graphicx}
\usepackage{subcaption}
\usepackage[utf8]{inputenc}
\usepackage[T1]{fontenc}
\usepackage{hyperref}
\usepackage{url}
\usepackage{booktabs}
\usepackage{amsmath,amssymb,amsfonts}
\usepackage{microtype}
\usepackage{xcolor}
\usepackage{enumitem}
\usepackage{textcomp}
\usepackage{multirow}
\usepackage{algorithm}
\usepackage{algorithmic}
\usepackage{wrapfig}
\usepackage{colortbl}
\usepackage{longtable}
\usepackage{placeins}
\usepackage{needspace}
\hypersetup{hidelinks}

\def \G{\mathrm{G}}
\def \F{\mathrm{F}}
\def \U{\mathrm{U}}

\newcommand{\LFGAS}{\textsc{LF-GAS}}
\newcommand{\GAGAS}{\textsc{GA-GAS}}
\newcommand{\APHIQL}{\textsc{AP-HIQL}}

\newtheorem{lemma}{Lemma}
\newtheorem{proposition}{Proposition}
\newtheorem{definition}{Definition}

\begin{document}

\title{SAGAS: Semantic-Aware Graph-Assisted Stitching for Offline Temporal Logic Planning}

\author{%
  Ruijia Liu \qquad Ancheng Hou \qquad Xiang Yin\thanks{Corresponding author: Xiang Yin (\texttt{yinxiang@sjtu.edu.cn}). Project page: \href{https://cps-sjtu.github.io/SAGAS/}{\nolinkurl{cps-sjtu.github.io/SAGAS}}.} \\
  School of Automation and Intelligent Sensing \\
  Shanghai Jiao Tong University, Shanghai 200240, China \\
  \texttt{\{liuruijia,hou.ancheng,yinxiang\}@sjtu.edu.cn}
}

\maketitle

\begin{abstract}
Linear Temporal Logic (LTL) provides a rigorous framework for specifying long-horizon robotic tasks, yet existing approaches face a trade-off: model-based synthesis relies on accurate labeled transition systems, whereas learning-based methods often require online interaction, task-specific rewards, or specification-conditioned training.
We study LTL-specified robotic planning and execution in an offline, model-free setting, where the agent is given only fixed, task-agnostic trajectory fragments, with no dynamics model, task demonstrations, or online data collection.
To address this setting, we propose \textbf{SAGAS}, a framework that combines the compositionality of symbolic synthesis with the data-driven reachability structure learned from offline trajectories.
SAGAS first learns a reusable latent reachability graph and a frozen goal-conditioned executor from fragmented offline data.
For each new LTL formula, it performs task-time semantic graph augmentation to ground state-defined propositions on the learned graph and applies B\"uchi product search to synthesize a cost-aware accepting prefix--suffix waypoint plan executed by the frozen executor.
This separation of learning and planning enables zero-shot generalization to unseen LTL formulas and new predicates supported by the offline data, without task-specific rewards, additional data collection, or policy retraining.
Experiments on LTL task suites constructed from OGBench locomotion domains show that SAGAS achieves higher success rates and lower execution costs than baselines on diverse unseen tasks, with further evaluation in OGBench Scene demonstrating its applicability to robotic manipulation.
\end{abstract}

\section{Introduction}
Linear Temporal Logic (LTL) provides a rigorous framework for specifying long-horizon robotic tasks whose requirements go beyond goal-reaching, such as sequential assembly, persistent surveillance, recurrence, and safety constraints~\cite{kloetzer2008fully,baier2008principles,scher2020warehouse}.
Automata-theoretic planning handles such specifications by reducing satisfaction to search over the product of a labeled transition system and a B\"uchi automaton~\cite{fainekos2005temporal,smith2011optimal,ren2024ltl}.
This offers a compositional and interpretable route to temporal-logic planning, but it assumes access to an accurate transition model whose states carry task-relevant proposition labels and whose transitions are dynamically feasible.
In many robotic domains, such a labeled transition model is unavailable. Learning-based alternatives relax this modeling assumption by encoding logical objectives into rewards~\cite{hasanbeig2018logically,voloshin2023eventual,shahltl}, augmenting policies with automaton states, or decomposing specifications into reusable skills and subgoals~\cite{qiu2023instructing,jackermeierdeepltl,liu2024skill,guo2025one}.
Some of these methods support compositional or zero-shot generalization to new temporal specifications by reusing learned skills, goal-conditioned policies, or symbolic subgoal interfaces.
However, this generalization is typically tied to a prescribed skill, goal, or proposition library, or to a distribution of specifications seen during training.
Such an interface can be restrictive when new propositions are defined directly over the robot state at test time, or when the available data consist only of task-agnostic trajectory fragments rather than demonstrations or rollouts for the relevant logical tasks (see Appendix~\ref{app:related} for extended related work).

We study LTL-specified robotic task planning and execution in an offline, model-free setting.
The agent is given only a fixed dataset of task-agnostic trajectory fragments, with no dynamics model, task demonstrations, or online data collection.
Rather than assuming a pre-specified library of temporal-logic skills or proposition-conditioned policies, we aim to learn reusable motion structure directly over the state space and compose it at test time according to a newly provided LTL formula.
This calls for an abstraction that captures dataset-supported motion connectivity while allowing task-time propositions to be grounded through the state interface.

To address this setting, we propose \textbf{SAGAS}, which extends Graph-Assisted Stitching (GAS)~\cite{baek2025graph} from single-goal trajectory stitching to temporal-logic synthesis over offline data.
Similar to GAS, SAGAS first learns a temporal-distance representation from fragmented offline trajectories, constructs a latent reachability graph in the learned space, and trains a goal-conditioned low-level policy to execute local transitions between graph waypoints.
This offline stage produces a reusable proxy for dataset-supported reachability, while the low-level policy provides a model-free executor for short-range waypoint tracking.
Crucially, these reusable components are learned without LTL task information, task rewards, or a pre-specified library of symbolic skills.

Given a new LTL task, SAGAS instantiates the symbolic layer only at test time. It augments the latent reachability graph with task-relevant proposition semantics evaluated on the stored state support, converting the reusable motion graph into a data-supported labeled abstraction suitable for automata-guided synthesis. SAGAS then searches the product of this semantic graph and the B\"uchi automaton to synthesize an accepting prefix--suffix waypoint plan, where graph costs serve as learned reachability surrogates and automaton transitions encode logical progress. The resulting high-level plan is executed by the pretrained frozen low-level policy. Through this decoupling, formula-specific reasoning is handled by semantic graph augmentation and symbolic product search, while the learned reachability proxy and executor are reused across tasks. Consequently, SAGAS generalizes zero-shot to unseen LTL formulas and new state-defined propositions in both planning and execution, provided that the propositions can be evaluated at task time and are sufficiently supported by the offline data. No task-specific reward design, data collection, or policy retraining is required.

Our contributions are summarized as follows.
First, we propose \textbf{SAGAS}, a framework that extends GAS-style trajectory stitching to LTL planning and execution by separating reachability learning on fixed, task-agnostic trajectories from task-time symbolic synthesis.
Second, we introduce a task-time semantic graph augmentation and cost-aware product-synthesis procedure over the learned latent graph, grounding state-defined propositions through preserved state support and optimizing accepting prefix--suffix waypoint plans using dataset-supported reachability costs.
Finally, on LTL task suites constructed from the OGBench AntMaze and HumanoidMaze domains, we show that SAGAS synthesizes and executes prefix--suffix behaviors zero-shot from fragmented offline data, improving finite-lasso execution success and execution cost over three baselines, especially on hard specifications. Further experiments in OGBench Scene demonstrate its applicability to robotic manipulation.

\section{Preliminaries}\label{sec:preliminaries}

\subsection{System model}
We consider a discrete-time dynamical system with unknown transition dynamics, defined over a continuous state space $\mathcal{S}\subseteq\mathbb{R}^{d_s}$ and an action space $\mathcal{U}_{\mathrm{act}}\subseteq\mathbb{R}^{d_a}$. The system evolves as $s_{t+1}=f(s_t,a_t)$,
where $f:\mathcal{S}\times\mathcal{U}_{\mathrm{act}}\rightarrow\mathcal{S}$ is an \emph{unknown} deterministic transition function. To relate the system state to high-level tasks, we define a projection map $\Pi : \mathcal{S} \to \mathcal{X}$, where $\mathcal{X}\subseteq\mathbb{R}^{d_t}$ denotes the \emph{task space} (e.g., robot workspace).
While the control inputs act on the high-dimensional state space $\mathcal{S}$, the logical predicates defining the task objectives are specified over the task space $\mathcal{X}$.

\subsection{Linear temporal logic task}\label{sec:LTL}
\textbf{Atomic propositions.}
For a test-time LTL specification $\phi$, let $\mathcal{AP}_\phi=\{\ell_1,\dots,\ell_m\}$ denote the finite set of atomic propositions appearing in $\phi$.
Each proposition $\ell\in\mathcal{AP}_\phi$ is associated with an evaluable predicate over the task space; for concreteness, we represent it as a measurable labeled region $\mathcal{R}_\ell\subseteq\mathcal{X}$.
The induced labeling function is
\(
L_\phi(s)=\{\ell\in\mathcal{AP}_\phi\mid \Pi(s)\in\mathcal{R}_\ell\}.
\)
The set can be empty when $\Pi(s)$ lies outside all labeled regions.
A state can satisfy several propositions simultaneously, and the same labeling interface describes predicates over joint object configurations.

\textbf{Linear temporal logic.}
We specify high-level tasks using Linear Temporal Logic without the ``next'' operator (LTL$_{-\mathrm{X}}$), a stutter-invariant fragment in which finite repetitions of the same proposition assignment do not affect formula satisfaction. This fragment is widely adopted for robotic tasks in continuous space, where moving between labeled regions takes a variable number of time steps~\cite{kloetzer2008fully,luo2021abstraction}. The syntax of LTL$_{-\mathrm{X}}$ is recursively defined as:
\[
\phi ::= \text{true} \mid \ell \in \mathcal{AP}_\phi
\mid \neg \phi
\mid \phi_1 \wedge \phi_2
\mid \phi_1 \vee \phi_2
\mid \phi_1\, \mathrm{U}\, \phi_2,
\]
where $\neg$, $\wedge$, and $\vee$ denote logical negation, conjunction, and disjunction, respectively, and $\mathrm{U}$ is the ``until'' operator. Common temporal operators such as ``eventually'' ($\F \phi = \text{true}\, \mathrm{U}\, \phi$) and ``always'' ($\G \phi = \neg \F \neg \phi$) are derived as standard abbreviations. Given a trajectory $\tau = s_0 s_1 s_2 \cdots$, the sequence of labels $L_\phi(s_0)L_\phi(s_1)L_\phi(s_2)\cdots$ constitutes an infinite word $\sigma \in (2^{\mathcal{AP}_\phi})^\omega$.
We write $\tau \models \phi$ if the induced word $\sigma$ satisfies $\phi$ according to standard LTL semantics~\cite{baier2008principles}.
Any LTL formula $\phi$ can be translated into a Nondeterministic B\"uchi Automaton (NBA) $\mathcal{B}_\phi = (Q, Q_0, \Sigma, \delta, F)$, where $\Sigma=2^{\mathcal{AP}_\phi}$ is the alphabet. An infinite run $q_0 q_1 \dots$ is accepting if it visits the set of accepting states $F$ infinitely often. If the automaton admits an accepting run, it admits one in lasso form: a finite \emph{prefix} leads from an initial state to an accepting state, and a nonempty \emph{suffix} cycle returns to that state and repeats indefinitely~\cite{baier2008principles,smith2011optimal,vasile2013sampling}.

\subsection{Problem setting}

We consider an \emph{offline} setting in which the agent has no access to the transition map $f$ and no additional environment interaction is allowed for data collection, policy learning, or planning.
Learning and planning must rely solely on a fixed dataset
\(
\mathcal{D}=\{\tau^{(i)}\}_{i=1}^N,
\tau^{(i)}=\big(s^{(i)}_0,a^{(i)}_0,s^{(i)}_1,\dots,a^{(i)}_{H_i-1},s^{(i)}_{H_i}\big),
\)
collected by unknown behavior policies.
The trajectories are task-agnostic and fragmented: they are not assumed to individually demonstrate the target LTL specification, nor are they labeled with the propositions that may appear in downstream tasks.

At test time, the agent is given an initial state $s_0$ and an LTL$_{-\mathrm{X}}$ specification $\phi$ over a finite set of state-evaluable propositions $\mathcal{AP}_\phi$.
The associated task-space regions or predicate evaluators define a labeling function $L_\phi:\mathcal{S}\to 2^{\mathcal{AP}_\phi}$ through the projection $\Pi$.
The formula, propositions, and labeled regions are introduced only at test time. As in standard offline learning, reliable synthesis is limited by the support contained in the fixed dataset $\mathcal{D}$, as further discussed in Appendix~\ref{app:limitations}.

We seek a hierarchical solution consisting of a reusable low-level executor learned from $\mathcal{D}$ and a task-specific high-level prefix--suffix plan
\(
\Gamma=(\Gamma_{\mathrm{pre}},\Gamma_{\mathrm{suf}}).
\)
Following the standard prefix--suffix representation used in automata-based LTL planning~\cite{smith2011optimal,vasile2013sampling,kantaros2018sampling,kantaros2020stylus,luo2021abstraction}, $\Gamma_{\mathrm{pre}}$ is a finite sequence of high-level subgoals executed once to reach an accepting automaton state, and $\Gamma_{\mathrm{suf}}$ is a finite cycle of subgoals repeated thereafter.
Let $\tau_\Gamma$ denote the physical trajectory induced by executing $\Gamma$ from $s_0$ with the learned executor.
The desired outcome is
\(
\tau_\Gamma \models \phi,
\)
where satisfaction is evaluated using the test-time labeling function $L_\phi$.

Among satisfying prefix--suffix plans, we seek to minimize the execution-cost objective
\begin{equation}\label{formula:problem}
J_{\mathrm{exec}}(\tau_\Gamma;\Gamma)
= \lambda\, T_{\mathrm{pre}}(\tau_\Gamma;\Gamma)
+ (1-\lambda)\, T_{\mathrm{suf}}(\tau_\Gamma;\Gamma),
\end{equation}
where $T_{\mathrm{pre}}$ is the number of environment steps used to complete the prefix, $T_{\mathrm{suf}}$ is the number of steps used to complete one suffix cycle, and $\lambda\in[0,1]$ balances transient and recurrent performance.
Since the true execution cost is unavailable during offline planning, the method below optimizes a graph-based surrogate of \eqref{formula:problem} using dataset-supported reachability costs and evaluates execution cost from realized rollouts.
This formulation separates task-time symbolic planning from reusable execution: new formulas and labeling functions determine the high-level prefix--suffix plan $\Gamma$, while the low-level executor is learned once from $\mathcal{D}$ and reused without task-specific retraining.

\begin{figure*}[tbp]
    \centering
    \includegraphics[width=0.85\textwidth]{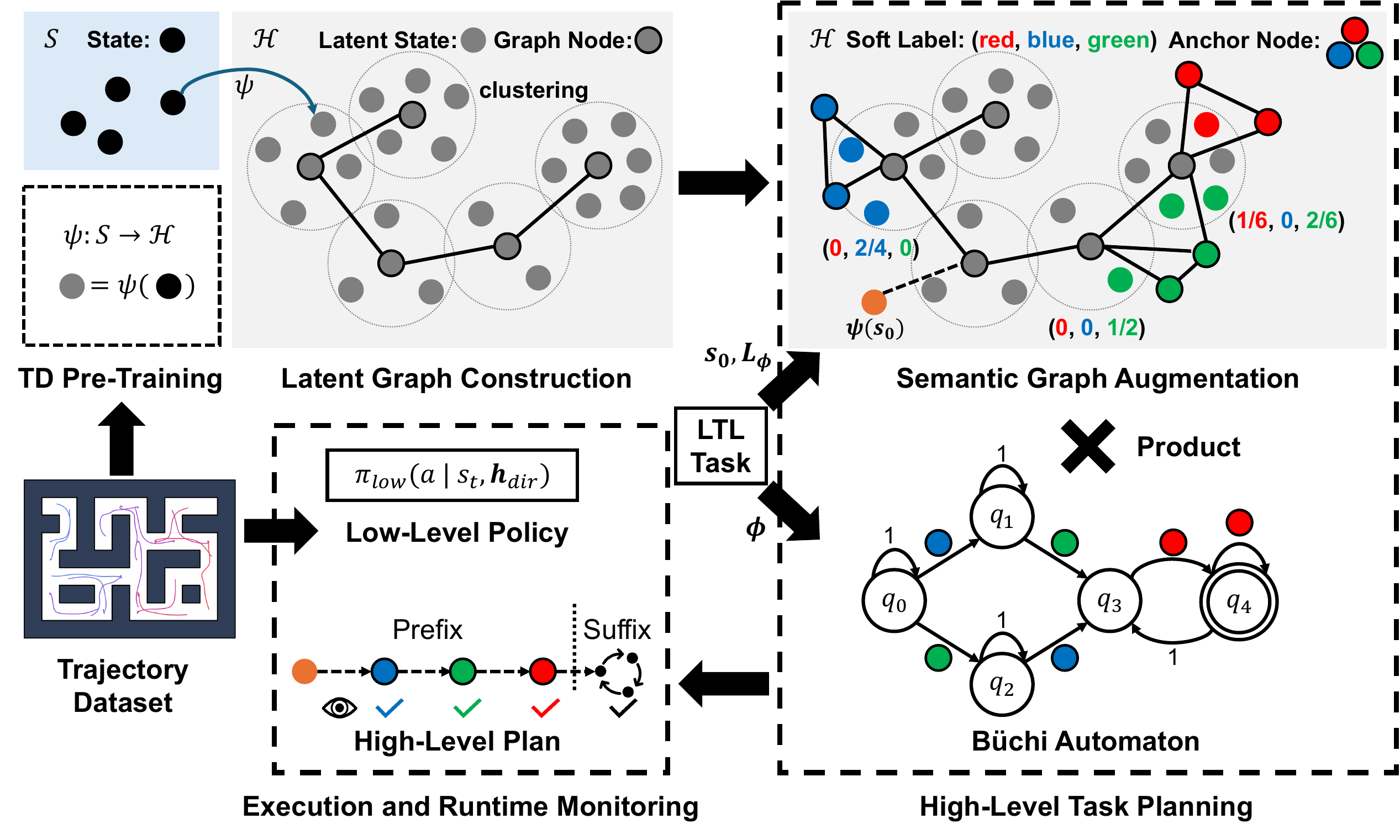}
    \caption{
Overview of \textbf{SAGAS}. A reusable reachability interface is built from
task-agnostic offline data. Given a new LTL formula and labeling function,
semantic graph augmentation and B\"uchi-product search produce an accepting
prefix--suffix plan, executed by the frozen low-level policy under runtime
monitoring.
}
    \label{fig:framework}
    \vspace{-1em}
\end{figure*}

\section{Our method}
\label{sec:method}
\subsection{Overview}
\label{sec:overview}
To address this problem, SAGAS separates offline motion learning from task-time symbolic synthesis (Figure~\ref{fig:framework}). The offline phase constructs a formula-agnostic reachability interface from the fixed trajectory dataset. This interface consists of three reusable objects: a weighted latent reachability graph $\mathcal{H}_{\mathrm{graph}}=(\mathcal{V},\mathcal{E},w)$, raw-state support sets $\{\mathcal{D}_v\}_{v\in\mathcal{V}}$ for graph nodes, and a frozen low-level policy $\pi_{\mathrm{low}}$ for executing local graph transitions. The graph and its edge weights summarize dataset-supported local connectivity, while the support sets preserve the raw-state evidence needed to ground future propositions. The low-level policy realizes the selected waypoint transitions in the physical system. Throughout the paper, a \emph{waypoint} denotes a selected latent-graph node used as a high-level subgoal, not a hand-specified task-space point. These components are built once from the task-agnostic dataset and do not depend on downstream LTL formulas, proposition regions, task rewards, or automaton states.

At task time, SAGAS receives a task instance $(s_0,\phi,L_\phi)$, specifying the initial state, LTL formula, and labeling function. The augmentation stage uses $L_\phi$ to ground propositions on the preserved state support and incorporates $s_0$ into a task-specific semantic graph (Section~\ref{sec:graph_aug}). This graph is the input to the product planner, which couples it with the task automaton and uses learned reachability costs to guide synthesis of an accepting prefix--suffix plan (Section~\ref{sec:planning}). The plan is then passed to the execution layer: its waypoints provide goals for the frozen policy, while its semantic obligations guide runtime monitoring of the realized trajectory (Section~\ref{sec:execution}). Algorithm~\ref{alg:sagas-main} summarizes this task-time workflow, with detailed algorithms provided in Appendix~\ref{app:algorithms}.

\begin{algorithm}[t]
\caption{SAGAS: task-time planning and execution}\label{alg:sagas-main}
\begin{algorithmic}[1]
\REQUIRE Frozen $(\mathcal H_{\mathrm{graph}},\{\mathcal D_v\},\psi,\pi_{\mathrm{low}})$ and offline states $\mathcal D_S$; task $(s_0,\phi,L_\phi)$; prefix budget $K$
\ENSURE Product plan $\rho^\star$, trajectory $\tau$, and execution outcome, or search failure
\STATE Translate $\phi$ to the original NBA $\mathcal B_\phi$ and normalize its guard terms $\mathcal G_\phi$
\STATE Compute soft labels $\widehat P$ from $\{\mathcal D_v\}$ and $L_\phi$
\STATE Retrieve witness states from $\mathcal D_S$ for individual labels and joint terms in $\mathcal G_\phi$, forming connected anchor nodes $\mathcal A$ with full assignments $L_\phi(s)$
\STATE Insert $\mathcal A$ and $s_0$ into the graph to form $(\mathcal H_\phi^{\mathrm{sem}},v_0)$ with $\widehat P$
\STATE $Q_{\mathrm{aligned}}\leftarrow\bigcup_{q_0\in Q_0}\delta(q_0,L_\phi(s_0))$
\STATE Restrict future guards by support; form $(\widetilde{\mathcal B}_\phi,\mathcal Z_0)$ from $Q_{\mathrm{aligned}}$
\STATE $\mathcal C_{\mathrm{pre}}\leftarrow\operatorname{PrefixSearch}(\mathcal Z_0,\mathcal V_\phi\times\widetilde F,K)$; $\mathcal C\leftarrow\varnothing$
\FORALL{$\rho_{\mathrm{pre}}\in\mathcal C_{\mathrm{pre}}$}
\STATE $z_f\leftarrow\operatorname{LastState}(\rho_{\mathrm{pre}})$; $\rho_{\mathrm{suf}}\leftarrow\operatorname{AcceptingSuffix}(z_f)$
\IF{a valid suffix $\rho_{\mathrm{suf}}$ is found}
\STATE $\mathcal C\leftarrow\mathcal C\cup\{(\rho_{\mathrm{pre}},\rho_{\mathrm{suf}})\}$
\ENDIF
\ENDFOR
\STATE \textbf{if} $\mathcal C=\varnothing$ \textbf{return} no accepting lasso found within the candidate budget
\STATE $\rho^\star\in\arg\min_{\rho\in\mathcal C}J_{\mathrm{graph}}(\operatorname{proj}_{\mathcal V_\phi}(\rho))$
\STATE $(\tau,\mathrm{outcome})\leftarrow\operatorname{ExecuteAndMonitor}(\rho^\star,\mathcal H_\phi^{\mathrm{sem}},\psi,\pi_{\mathrm{low}},\mathcal B_\phi,L_\phi)$
\RETURN $\rho^\star$, $\tau$, and $\mathrm{outcome}$
\end{algorithmic}
\end{algorithm}
\FloatBarrier

\subsection{Offline reachability interface}
\label{sec:offline_graph}

We instantiate the support-preserving interface with a GAS-style temporal-distance graph and low-level executor~\cite{baek2025graph}. These objects expose reusable motion structure that can be labeled and searched once a formula is specified. After the offline stage, the graph, support, and executor are frozen and reused across downstream tasks. Temporal-distance learning, filtering, clustering, and executor training are detailed in Appendix~\ref{app:offline-backbone}.

\paragraph{Latent temporal-distance graph.}
Let $\psi:\mathcal{S}\to\mathcal{H}$ denote an embedding learned from the offline trajectories, where $\mathcal{H}\subseteq\mathbb{R}^{d_h}$ is the latent representation space.
The embedding is trained so that Euclidean distances between embeddings approximate temporal distance, or step-to-go reachability, between the underlying system states.
SAGAS uses this latent distance as a proxy for local reachability cost.

Using $\psi$, SAGAS first applies the GAS temporal-efficiency rule to select trajectory states whose observed motion aligns with the learned reachability geometry.
The retained states are then clustered in the latent space at a scale set by $H_{\mathrm{TD}}$ to form a weighted graph
\(
\mathcal{H}_{\mathrm{graph}}=(\mathcal{V},\mathcal{E},w).
\)
Each node $v\in\mathcal{V}$ corresponds to a latent cell $\mathcal{C}_v\subseteq\mathcal{H}$, i.e., a cluster of embedded dataset states.
With a slight abuse of notation, we also use $v$ to denote the representative point of this cell in latent space.
The temporal-distance scale $H_{\mathrm{TD}}$ also controls local connectivity: nodes within this distance are connected, and nearest-node links are added between disconnected components.
For distinct nodes, the local connection rule and edge costs are
\begin{equation}\label{eq:graph-construction}
\|u-v\|_2\le H_{\mathrm{TD}}\ \Longrightarrow\ (u,v)\in\mathcal E,
\qquad w(u,v)=\|u-v\|_2\quad\forall(u,v)\in\mathcal E.
\end{equation}
The resulting graph summarizes motion connectivity learned from full-state trajectories, with edge weights providing empirical estimates of transition cost.

\paragraph{Raw-state support map.}
The latent graph alone is insufficient for temporal-logic planning because LTL propositions are evaluated in the original state or task space, not directly in the learned latent space.
SAGAS therefore retains the raw dataset support associated with each graph node.
Let $\mathcal{D}_{S}$ denote the set of states appearing in the offline dataset.
For each node $v$, we store a support set $\mathcal{D}_v\subseteq\{s\in\mathcal{D}_{S}\mid\psi(s)\in\mathcal{C}_v\}$.
This support map connects each latent cell to observed physical states, allowing test-time predicates to be evaluated through $\Pi$ and grounded on the graph.

\paragraph{Frozen low-level executor.}
SAGAS learns a low-level policy $\pi_{\mathrm{low}}$ from the same dataset and temporal scale $H_{\mathrm{TD}}$ to execute local graph transitions. Conditioned on the target waypoint's latent direction, it tracks that waypoint until the execution rule declares completion or the tracking budget is exhausted.

\subsection{Semantic graph augmentation}
\label{sec:graph_aug}

The offline graph $\mathcal{H}_{\mathrm{graph}}$ encodes dataset-supported reachability, but it is not yet a labeled abstraction for LTL synthesis.
Given a task instance with initial state $s_0$, test-time formula $\phi$ over propositions $\mathcal{AP}_\phi$, and labeling function $L_\phi:\mathcal{S}\to 2^{\mathcal{AP}_\phi}$, SAGAS performs semantic graph augmentation according to the supplied proposition predicates.
The augmentation first estimates empirical labels for existing graph nodes and then inserts semantically explicit nodes for the task-relevant labeled regions.

\paragraph{Soft labels for empirical graph semantics.}
The latent graph provides a compressed representation of the offline dataset, with each node representing a cluster of embedded states.
As the embedding $\psi$ has no inverse decoder, a node's latent coordinate cannot be directly converted into a physical state for proposition evaluation.
SAGAS therefore evaluates test-time propositions on the node's preserved raw-state support to estimate how strongly the node is associated with each proposition.
For an original graph node $v$, the empirical label estimate, or \emph{soft label}, is written in weighted form as
\begin{equation}\label{eq:soft-label-main}
\widehat{P}(\ell \mid v) =
\frac{
\sum_{s\in\mathcal{D}_v}K_v(s)\mathbf{1}\{\ell\in L_\phi(s)\}
}{
\sum_{s\in\mathcal{D}_v}K_v(s)
}.
\end{equation}
Here $K_v(s)\geq 0$ is a support-state weight.
Intuitively, $\widehat P(\ell\mid v)$ measures the weighted proportion of the node's preserved support that satisfies proposition $\ell$: values near one indicate that most support states satisfy it, whereas values near zero indicate little empirical association.
These soft labels provide graph search with a semantic reference for deciding which nodes to traverse.
They are particularly useful for negative literals: when a guard forbids $\ell$, the planner uses the estimate to avoid nodes strongly associated with that proposition.
Specifically, an original graph node $v$ passes this negative-literal check only when its support has positive total weight and $\widehat P(\ell\mid v)\le\tau_{\mathrm{soft}}$.
Appendix~\ref{app:soft-labels} details the screening mechanism.

\paragraph{Anchor nodes for labeled regions.}
Soft labels characterize an original graph node through its preserved support, which can contain states both inside and outside a proposition region.
Consequently, tracking the node's latent representative may reach a state where the desired proposition is false.
To provide an explicit semantic target, SAGAS retrieves label-satisfying dataset states and embeds them as \emph{anchor nodes}.
Each anchor $a_s=\psi(s)$ retains the full label set $L_\phi(s)$ of its observed witness $s$ and is connected to the graph using the same temporal-distance scale $H_{\mathrm{TD}}$.

The stored label set allows the planner to determine which logical transitions an anchor can witness.
For a conjunctive guard term $\eta$ with positive and negative proposition sets $P_\eta$ and $N_\eta$, a connected anchor is compatible when its witness satisfies $P_\eta\subseteq L_\phi(s)$ and $N_\eta\cap L_\phi(s)=\varnothing$.
For example, a transition requiring $a\wedge b$ uses an anchor whose witness state satisfies both propositions simultaneously.
SAGAS retains anchors for individual propositions and retrieves additional witnesses as needed for joint requirements.
If the required witness cannot be found in the available data or connected to the graph, SAGAS reports the guard term as unsupported by the current abstraction and disables that alternative.
Appendix~\ref{app:anchor-nodes} details witness retrieval and insertion.

\paragraph{Semantic interface for product search.}
The resulting semantic graph
\(\mathcal{H}^{\mathrm{sem}}_{\phi}=(\mathcal V_\phi,\mathcal E_\phi,w_\phi)\)
combines soft labels, anchor witnesses, and inherited reachability costs. It also embeds and connects a temporary start node for $s_0$; if local insertion yields no connection, the nearest graph node provides an initialization heuristic.
This separation keeps the offline reachability graph compact and formula-agnostic.
Once a task is specified, soft labels and anchors add semantic resolution only where needed for its propositions, avoiding uniform refinement of the entire graph.
Appendix~\ref{app:successor-oracle} gives the full construction.
\subsection{High-level planning over the semantic graph}
\label{sec:planning}

With this semantic interface in place, SAGAS turns the test-time LTL task into a search problem over the product of the semantic graph and the task automaton.
The graph supplies dataset-supported connectivity and learned reachability costs, while its soft labels and anchors provide semantic evidence for enabling automaton transitions.
The resulting product search couples logical progress with dataset-supported reachability, using the learned costs to guide the synthesis of an accepting prefix--suffix plan.

\textbf{B\"uchi automaton preprocessing.}
Given the LTL formula $\phi$, SAGAS translates it into a Nondeterministic B\"uchi Automaton (NBA) $\mathcal{B}_{\phi}=(Q,Q_0,\Sigma,\delta,F)$.
We represent $\delta$ symbolically: for each pair $(q,q')$, the Boolean guard $g_{q,q'}$ is satisfied by exactly those proposition assignments $\sigma\in\Sigma$ for which $q'\in\delta(q,\sigma)$.
Before search, SAGAS converts these guards into disjunctive normal form (DNF), so that each conjunctive alternative can be checked independently against the semantic graph.
Contradictory terms are discarded, and a term containing positive literals is retained only if a connected anchor witnesses the complete term.
The remaining alternatives define the support-restricted automaton $\widetilde{\mathcal B}_\phi$, with states $\widetilde Q\subseteq Q$ and accepting set $\widetilde F=F\cap\widetilde Q$.
Appendix~\ref{app:automaton-preprocess} details preprocessing.

\textbf{Product search and prefix--suffix synthesis.}
SAGAS follows the classical prefix--suffix approach to B\"uchi acceptance: a prefix reaches an accepting product state, and a suffix returns to that state so that its repetition yields infinitely many accepting visits~\cite{smith2011optimal}.
Both stages operate on the implicit product system
\[
\mathcal{P}_{\phi}
=
\mathcal{H}^{\mathrm{sem}}_{\phi}\otimes\widetilde{\mathcal{B}}_{\phi},
\qquad
\mathcal{Z}_{\phi}=\mathcal{V}_{\phi}\times\widetilde Q .
\]
Here $\mathcal V_\phi$ is the node set of the semantic graph, and each product state $(v,q)\in\mathcal Z_\phi$ pairs a graph node with a state of the support-restricted automaton.

\emph{Prefix search.} The prefix search starts from the product states associated with the initial system state $s_0$ and aims to reach the accepting set $\mathcal V_\phi\times\widetilde F$.
To establish these starting states, SAGAS first reads the observed assignment $L_\phi(s_0)$ in the automaton, obtaining the possible automaton states $Q_{\mathrm{init}}$ and the initial product set $\mathcal Z_0$:
\[
Q_{\mathrm{init}}
=
\{q'\in\widetilde Q\mid \exists q_0\in Q_0,\ q'\in\delta(q_0,L_\phi(s_0))\},
\qquad
\mathcal{Z}_0=\{(v_0,q):q\in Q_{\mathrm{init}}\}.
\]
Here $v_0$ is the temporary start node introduced during semantic graph augmentation. The initialization uses the original NBA's initial-state set $Q_0$ and transition relation $\delta$, consuming $L_\phi(s_0)$ once.

SAGAS then explores the product from $\mathcal Z_0$ using A*-style search.
Each expansion combines a neighboring graph node $v'$ with an outgoing automaton transition $q\to q'$ whose guard is supported by the semantic evidence at $v'$.

The semantic check follows the soft-label and anchor rules introduced above.
For a guard term $\eta$ with positive and negative proposition sets $P_\eta$ and $N_\eta$, $\operatorname{PosOK}(v',\eta)$ requires $v'$ to be an anchor whose witness satisfies all propositions in $P_\eta$; it is automatically true when $P_\eta$ is empty.
The check $\operatorname{NegOK}(v',\eta)$ requires every forbidden proposition to have soft label at most $\tau_{\mathrm{soft}}$ at an original graph node with positive total support weight, or to be absent from the observed assignment at an anchor or start node.
It is automatically true when $N_\eta$ is empty; an original graph node without valid support fails a nonempty negative check.
Writing $\widetilde g_{q,q'}$ for the retained guard and $\mathcal E_\phi$ for the semantic graph's edge set, the movement successors satisfy
\begin{equation}\label{eq:product-transition-main}
\medmuskip=2mu \thickmuskip=3mu
((v,q),(v',q'))\in\mathcal E_{\mathcal P}^{\mathrm{mov}}\Longleftrightarrow(v,v')\in\mathcal E_\phi\land\exists\eta\in\operatorname{DNF}(\widetilde g_{q,q'}):\operatorname{PosOK}(v',\eta)\land\operatorname{NegOK}(v',\eta).
\end{equation}
Each movement edge inherits the cost $w_\phi(v,v')$ of its semantic-graph edge.
SAGAS generates successors from this relation as needed during search, using accumulated edge costs and a graph-distance heuristic to guide expansion toward accepting states. This on-demand construction uses the same transition rules as an explicitly constructed product graph.
When an accepting state is reached, the planner reconstructs its prefix and retains it for suffix construction.
Because different endpoints can admit different return cycles, SAGAS retains up to $K$ low-cost prefixes, stopping when this budget is reached or the search frontier is exhausted.

\emph{Suffix search.} At each retained endpoint $z_f=(u,q_f)$, SAGAS first checks whether $q_f$ has a self-loop whose guard is enabled at $u$. If so, the suffix can stay at the same waypoint with zero graph cost.
Otherwise, SAGAS searches for a nonempty path back to $z_f$ using the same movement transitions and edge costs as in prefix search. Only prefixes with a valid suffix enter the final cost comparison.
The complete successor relation and search procedures are given in Appendices~\ref{app:successor-oracle} and~\ref{app:prefix-suffix-search}.

\textbf{Graph surrogate objective.}
Each product plan $\rho=(\rho_{\mathrm{pre}},\rho_{\mathrm{suf}})$ has a waypoint projection $\Gamma=\operatorname{proj}_{\mathcal V_\phi}(\rho)=(\Gamma_{\mathrm{pre}},\Gamma_{\mathrm{suf}})$, obtained by retaining the graph-node component of each product state. SAGAS selects the candidate $\rho^\star$ minimizing the graph surrogate cost
\begin{equation}
J_{\mathrm{graph}}(\Gamma)
=
\lambda\,\mathrm{Cost}(\Gamma_{\mathrm{pre}})
+
(1-\lambda)\,\mathrm{Cost}(\Gamma_{\mathrm{suf}}),
\end{equation}
where $\mathrm{Cost}(\cdot)$ sums the edge weights of graph movements and $\lambda\in[0,1]$ trades off transient and recurrent efficiency.
This objective serves as a graph-level surrogate for the execution objective in Eq.~\eqref{formula:problem}, replacing the environment-step counts for the prefix and one suffix traversal with sums of learned reachability costs. Since the actual step counts are unavailable during planning, SAGAS ranks the generated candidates using this surrogate.
Detailed search and monitoring procedures are given in Algorithms~\ref{alg:product-search} and~\ref{alg:execution}.

\subsection{Plan execution and runtime monitoring}
\label{sec:execution}

SAGAS follows the waypoint projection of $\rho^\star$ with the frozen low-level policy, using runtime monitoring to decide when to advance.
Waypoints inherited from the offline graph guide intermediate motion and may be skipped when locally reachable within $H_{\mathrm{TD}}$.
The task-time anchors selected by the plan instead serve as mandatory semantic checkpoints, where both latent proximity and the required positive labels under $L_\phi$ must be verified.
As the policy tracks each selected target, the monitor starts from the confirmed source automaton state, $S=\{q_i\}$, and updates $S\leftarrow\bigcup_{q\in S}\delta(q,L_\phi(s_t))$ on each new observation using the original NBA.
Reaching the target advances the plan only if its planned automaton state $q_j$ belongs to $S$ and the guard checks pass.
On confirmation, the monitor resets to $S=\{q_j\}$ and processes subsequent observations for the next segment.

Throughout this motion, the monitor also checks negative obligations until their release.
To help the executor avoid violations between waypoints, navigation tracking adds a bounded repulsive bias away from nearby anchors associated with active forbidden labels.
If an anchor is missed, tracking times out, or an active negative obligation is violated, the planned product transition is reported as unrealized.
After completing the prefix, the executor repeats the suffix under the same monitoring rules.
A cycle is counted as complete only when the suffix endpoint is reached and the monitor confirms a return to the planned accepting state from the labels observed during that cycle.
Appendix~\ref{app:execution-details} gives the full execution procedure.

\begin{proposition}[Soundness of synthesis and monitored execution]\label{prop:conditional-main}
For every product plan $\rho$ returned by SAGAS, the initial label and the guards along its prefix and repeated suffix define a nonempty set of infinite label words accepted by the original B\"uchi automaton $\mathcal B_\phi$. During execution, the labels observed between consecutive targets confirmed by the monitor admit a run of this automaton connecting their planned automaton states. If the prefix is completed and the monitor confirms completion of infinitely many suffix cycles, then $\tau_\Gamma\models\phi$.
\end{proposition}
\noindent\emph{Proof sketch.} Each selected guard term is satisfiable and implies an original NBA guard. Any label word consistent with the initial label and selected guard sequence therefore admits an accepting run along the plan's automaton states. During execution, induction on the monitor updates shows that each confirmed segment admits an automaton run between its planned source and target states, allowing successive runs to be concatenated at their shared states. This run follows the observed physical labels and visits the accepting state on every confirmed suffix cycle, so infinitely many such cycles imply $\tau_\Gamma\models\phi$. Appendix~\ref{app:conditional-proof} gives the full proof.
\Needspace{8\baselineskip}

\section{Experiments}\label{sec:experiments}
Our experiments examine the reliability, efficiency, and generality of zero-shot LTL planning from fixed offline data.
\textbf{Q1:} Does semantic product planning improve LTL satisfaction under a shared offline backbone?
\textbf{Q2:} Does joint graph--automaton cost optimization reduce capped execution cost relative to decoupled logic-first planning?
\textbf{Q3:} How does performance vary across dataset regimes, maze scales, and robot dynamics?
\textbf{Q4:} Can the same semantic interface handle overlapping regions and joint object configurations?
\subsection{Experimental setup}\label{sec:exp_setup}

\textbf{Environments and datasets.}
We evaluate on OGBench~\cite{park2025ogbench} AntMaze, HumanoidMaze, and Scene using the official task-agnostic offline datasets, collected independently of any LTL formula.
Navigation covers eight AntMaze settings across the \texttt{navigate}, \texttt{stitch}, and \texttt{explore} regimes and six HumanoidMaze settings across \texttt{navigate} and \texttt{stitch}, with multiple maze sizes in each domain.
Scene contains a robot arm, a cube, drawer, window, and two buttons that toggle the drawer and window locks; its dataset consists of scripted interactions with these objects.
Within each setting, all graphs, representations, and policies are frozen and reused across tasks, without task-specific data collection or retraining.

\textbf{Compared methods.}
To compare high-level planning strategies, \LFGAS{} and \GAGAS{} share SAGAS's graph and frozen executor.
\LFGAS{} first selects an automaton route and then connects its transitions through the graph; \GAGAS{} greedily advances automaton transitions through nearest semantic anchors without product search.
We also compare with \APHIQL{}, which adapts LTL-instructed goal reaching~\cite{qiu2023instructing} using a separate HIQL policy~\cite{park2023hiql}.

\textbf{Task suites.}
For navigation, we generate unseen LTL$_{-\mathrm{X}}$ tasks by randomly instantiating and composing reachability, sequencing, safety, and recurrence templates.
Easy tasks use one template, Medium tasks conjoin up to three, and Hard tasks up to four; Easy and Medium use at most five regions, while Hard allows eight.
Regions are sampled to be disjoint to simplify feasibility control.
All methods receive the same tasks and initial states: 100 tasks per setting and difficulty, totaling 4,200 cases.
We also evaluate 320 overlapping-region tasks across AntMaze Giant-Stitch and Large-Explore, and 120 Scene tasks with joint object-state predicates.
The Scene predicates describe joint object configurations in a 7D mixed discrete--continuous space.

\textbf{Evaluation.}
We report finite-lasso execution success rate (SR), determined by independent post-rollout verification of the labels along each saved state trajectory.
For recurrent tasks, success requires prefix completion followed by $M=2$ suffix traversals in both navigation suites and $M=1$ in Scene.
Navigation efficiency is measured by normalized capped cost (NCC): $\min\{T_{\mathrm{pre}}+T_{\mathrm{suf}},2C_{\max}\}/(2C_{\max})$ when the prefix and first suffix traversal are completed, and $1$ otherwise, where $C_{\max}$ is the configured cost budget.
Lower NCC therefore reflects both shorter executions and fewer capped failures.
Appendix~\ref{app:exp-details} gives task designs, baselines, and suite-specific parameters.
\subsection{Results and analysis}
\label{sec:large_scale_exp}
Table~\ref{tab:ltl-domain-summary} summarizes the navigation comparison as means and cross-setting standard deviations; Appendix~\ref{app:exp-results} reports the per-setting results.

\begin{table}[tbp]
\centering
\small
\caption{Aggregate results over 8 AntMaze and 6 HumanoidMaze environments. SR is finite-lasso execution success rate; NCC is normalized capped cost. Entries are means $\pm$ cross-setting standard deviations. Dark/light gray shading marks the best/second-best method within each domain--difficulty group; higher SR and lower NCC are better.}
\label{tab:ltl-domain-summary}
\resizebox{\linewidth}{!}{%
\begin{tabular}{llrrrrrr}
\toprule
Domain & Method & \multicolumn{2}{c}{Easy} & \multicolumn{2}{c}{Medium} & \multicolumn{2}{c}{Hard} \\
\cmidrule(lr){3-4} \cmidrule(lr){5-6} \cmidrule(lr){7-8}
 &  & SR (\%) & NCC & SR (\%) & NCC & SR (\%) & NCC \\
\midrule
AntMaze & SAGAS & \cellcolor{gray!30}$75.4\pm10.6$ & \cellcolor{gray!30}$0.26\pm0.11$ & \cellcolor{gray!30}$75.1\pm14.0$ & \cellcolor{gray!30}$0.26\pm0.13$ & \cellcolor{gray!30}$65.0\pm13.0$ & \cellcolor{gray!30}$0.38\pm0.13$ \\
 & \LFGAS{} & \cellcolor{gray!12}$73.5\pm11.5$ & \cellcolor{gray!12}$0.28\pm0.12$ & \cellcolor{gray!12}$71.5\pm12.6$ & \cellcolor{gray!12}$0.30\pm0.13$ & \cellcolor{gray!12}$55.9\pm14.4$ & \cellcolor{gray!12}$0.46\pm0.14$ \\
 & \GAGAS{} & $67.1\pm9.0$ & $0.36\pm0.09$ & $66.1\pm12.3$ & $0.36\pm0.13$ & $53.6\pm12.7$ & $0.49\pm0.13$ \\
 & \APHIQL{} & $56.0\pm24.8$ & $0.45\pm0.25$ & $50.7\pm24.8$ & $0.50\pm0.25$ & $39.8\pm20.4$ & $0.63\pm0.20$ \\
\midrule
HumanoidMaze & SAGAS & \cellcolor{gray!30}$64.8\pm14.1$ & \cellcolor{gray!30}$0.37\pm0.11$ & \cellcolor{gray!30}$57.8\pm13.7$ & \cellcolor{gray!30}$0.45\pm0.14$ & \cellcolor{gray!30}$49.5\pm16.9$ & \cellcolor{gray!30}$0.54\pm0.16$ \\
 & \LFGAS{} & \cellcolor{gray!12}$63.0\pm14.9$ & \cellcolor{gray!12}$0.40\pm0.14$ & \cellcolor{gray!12}$56.7\pm14.2$ & \cellcolor{gray!12}$0.46\pm0.13$ & \cellcolor{gray!12}$43.2\pm14.0$ & \cellcolor{gray!12}$0.60\pm0.13$ \\
 & \GAGAS{} & $59.5\pm14.3$ & $0.44\pm0.13$ & $50.2\pm10.2$ & $0.53\pm0.09$ & $41.2\pm14.5$ & $0.63\pm0.12$ \\
 & \APHIQL{} & $55.7\pm27.1$ & $0.49\pm0.25$ & $50.8\pm27.7$ & $0.54\pm0.25$ & $37.7\pm21.7$ & $0.67\pm0.19$ \\
\bottomrule
\end{tabular}
}
\vspace{-1.5em}
\end{table}

\textbf{Q1: LTL satisfaction.}
SAGAS achieves the highest aggregate SR in every domain--difficulty group, improving Hard-task SR over \LFGAS{} by 9.1 percentage points in AntMaze and 6.3 in HumanoidMaze.
These gains support joint reasoning about logical progress and physical reachability when several events, avoidance constraints, and suffix cycles must be coordinated.
Paired comparisons on all 4,200 cases confirm significant aggregate SR gains over all three baselines in both domains after Holm correction (largest corrected $p=0.032$; Appendix~\ref{app:paired-statistics}).

\textbf{Q2: Execution cost.}
SAGAS also obtains the lowest aggregate NCC in every domain--difficulty group.
Relative to \LFGAS{}, Hard-task NCC decreases from $0.46$ to $0.38$ in AntMaze and from $0.60$ to $0.54$ in HumanoidMaze.
Because NCC also penalizes failures, we separately compare execution length on tasks solved by both methods.
Across all difficulties, SAGAS uses on average 103.3 fewer environment steps in AntMaze and 210.0 fewer in HumanoidMaze for the prefix plus the first suffix traversal, supporting the use of joint graph--automaton costs to select shorter executable plans (Appendix~\ref{app:common-success}).
Appendix~\ref{app:search-scale} examines planning time and candidate-budget sensitivity; Appendix~\ref{app:surrogate-cost} relates graph cost to execution length.

\textbf{Q3: Dataset regime, maze size, and robot dynamics.}
In AntMaze, \APHIQL{} is most competitive on the longer, coherent Navigate trajectories.
SAGAS's SR advantage grows from 4.11 percentage points on Navigate to 22.44 on Stitch and 52.17 on Explore, consistent with the value of stitching behavior from fragmented or exploratory data (Appendix~\ref{app:regime-analysis}).
Within AntMaze Navigate, SAGAS achieves $63.7\%$ SR on the Large maze and $55.3\%$ on Giant.
Across all settings, its aggregate SR is $71.8\%$ in AntMaze and $57.4\%$ in HumanoidMaze, where it leads in each difficulty group.
Most SAGAS failures occur during execution. Appendices~\ref{app:failure-analysis} and~\ref{app:ablations-diagnostics} examine these failures and the effects of soft-label screening and guard-aware steering.

\textbf{Q4: Predicate generalization.}\label{sec:predicate-generalization}
SAGAS handles simultaneous proposition requirements in the overlap suite, finding 292/320 accepting lassos and achieving $81.3\%$ SR (260/320).
The same interface also supports joint object configurations in Scene: it finds 120/120 lassos and achieves $67.5\%$ SR (81/120) while coordinating manipulation under the objects' lock dependencies.
Table~\ref{tab:predicate-generalization} in the appendix summarizes the results; Appendices~\ref{app:overlap-protocol} and~\ref{app:scene-protocol} give the protocols, and Appendix~\ref{app:predicate-results} provides confidence intervals and per-family analyses.

\Needspace{30\baselineskip}

\subsection{Qualitative case study}
\label{sec:case_study}
We use a patrol task in \texttt{antmaze-giant-stitch} to illustrate how SAGAS plans and executes a long-horizon route from short offline trajectory fragments.
The robot must repeatedly patrol three regions in order and visit three others in sequence at least once. The middle region of this second sequence is off limits until the robot visits a designated checkpoint. Writing $e_i$ for the proposition that the robot is in region $i$, the task is
\begin{equation}\label{formula:LTL_task1}
\underbrace{\G(\F(e_1 \land \F(e_2 \land \F e_3)))}_{\text{Infinite Patrol}}
\land \underbrace{\F(e_4 \land \F(e_5 \land \F e_6))}_{\text{Sequential Visit}} \land \!\!\!\underbrace{(\neg e_5\U e_7)}_{\text{Conditional Safety}}\!\!\!.
\end{equation}

\setlength{\intextsep}{6pt plus 1pt minus 1pt}
\begin{wrapfigure}[14]{r}{0.46\textwidth}
    \vspace{-0.8em}
    \centering
    \includegraphics[width=\linewidth]{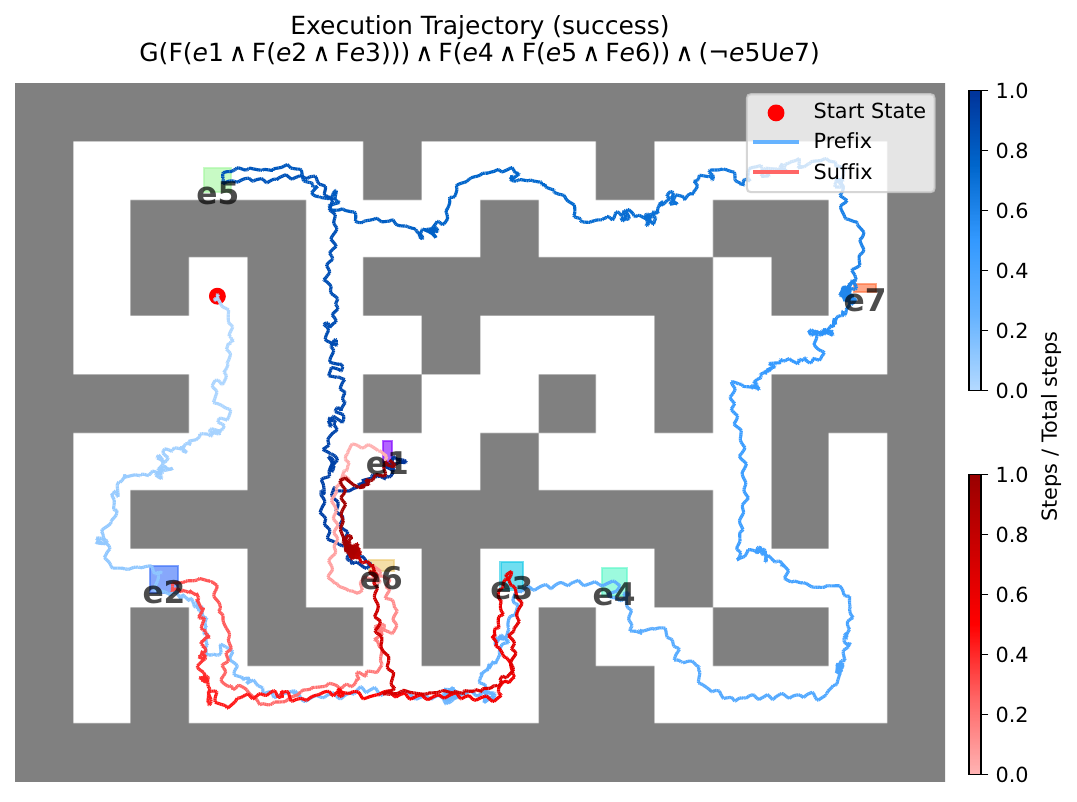}
    \caption{Execution trajectory for Eq.~(\ref{formula:LTL_task1}), with the prefix shown in blue and the recurrent suffix shown in red.}
    \label{fig:case_study}
\end{wrapfigure}
Figure~\ref{fig:case_study} shows the resulting trajectory. Along the blue prefix, the robot reaches $e_4$, detours to the checkpoint $e_7$, and only then enters $e_5$ and continues to $e_6$. With the one-time sequence complete, it follows the red patrol loop through $e_1$, $e_2$, and $e_3$.
No individual trajectory in the offline dataset demonstrates this complete temporal pattern; SAGAS constructs it by stitching fragments together.
Additional qualitative case studies are provided in Appendix~\ref{app:case-studies}.
\setlength{\intextsep}{12pt plus 2pt minus 2pt}

\section{Conclusion}
\label{sec:conclusion}

We presented \textbf{SAGAS}, an offline, model-free framework for LTL planning and execution from fixed, task-agnostic trajectory fragments.
Task-time semantic graph augmentation grounds new predicates on preserved state support, while cost-aware B\"uchi product search synthesizes plans for the frozen executor.
This separation enables zero-shot generalization across unseen formulas and data-supported predicates without task-specific data collection or policy retraining.
Experiments in fixed-topology OGBench environments show higher finite-lasso execution success rates and lower execution costs than baselines in locomotion, with further evaluations demonstrating support for overlapping predicates and robotic manipulation.
Future work will investigate whether this generalization can extend beyond tasks and predicates to unseen layouts and changing topology.
Appendix~\ref{app:limitations} discusses the assumptions and limitations in detail.

\clearpage
\begin{ack}
This work was supported by the National Science and Technology Major Project (2025ZD1600700) and the National Natural Science Foundation of China (62573291, 62533017).
The authors declare no competing interests.
\end{ack}

\bibliographystyle{unsrtnat}
\bibliography{ifacconf}

\appendix
\numberwithin{equation}{section}
\numberwithin{figure}{section}
\numberwithin{table}{section}
\numberwithin{algorithm}{section}

\clearpage
\section*{Appendix}

\begingroup
\setlength{\parskip}{3pt}
\setlength{\parfillskip}{0pt}
\newcommand{\appendixentry}[3]{%
  \noindent\makebox[1.8em][l]{\textbf{#1}}%
  #2\nobreak\hspace{0.6em}%
  \leaders\hbox to 0.65em{\hss.\hss}\hfill
  \nobreak\hspace{0.6em}\makebox[2em][r]{#3}\par
}
\newcommand{\appendixsubentry}[1]{%
  {\small\noindent\hspace*{1.8em}\makebox[2.6em][l]{\ref{#1}}%
  \hyperref[#1]{\nameref*{#1}}\nobreak\hspace{0.6em}%
  \leaders\hbox to 0.65em{\hss.\hss}\hfill
  \nobreak\hspace{0.6em}\makebox[2em][r]{\pageref{#1}}\par}%
}
\appendixentry{\ref{app:related}}{\hyperref[app:related]{Related work}}{\pageref{app:related}}
\appendixentry{\ref{app:algorithms}}{\hyperref[app:algorithms]{Algorithmic details}}{\pageref{app:algorithms}}
\appendixsubentry{app:offline-backbone}
\appendixsubentry{app:soft-labels}
\appendixsubentry{app:anchor-nodes}
\appendixsubentry{app:automaton-preprocess}
\appendixsubentry{app:successor-oracle}
\appendixsubentry{app:prefix-suffix-search}
\appendixsubentry{app:execution-details}
\appendixsubentry{app:complexity}
\appendixentry{\ref{app:conditional-proof}}{\hyperref[app:conditional-proof]{Soundness of synthesis and monitored execution}}{\pageref{app:conditional-proof}}
\appendixentry{\ref{app:limitations}}{\hyperref[app:limitations]{Scope and limitations}}{\pageref{app:limitations}}
\appendixentry{\ref{app:exp-details}}{\hyperref[app:exp-details]{Experimental protocols and implementation details}}{\pageref{app:exp-details}}
\appendixsubentry{app:navigation-environments}
\appendixsubentry{app:navigation-task-generation}
\appendixsubentry{app:compared-methods}
\appendixsubentry{app:evaluation-metrics}
\appendixsubentry{app:implementation-details}
\appendixsubentry{app:overlap-protocol}
\appendixsubentry{app:scene-protocol}
\appendixentry{\ref{app:exp-results}}{\hyperref[app:exp-results]{Detailed task results}}{\pageref{app:exp-results}}
\appendixsubentry{app:navigation-results}
\appendixsubentry{app:paired-statistics}
\appendixsubentry{app:common-success}
\appendixsubentry{app:regime-analysis}
\appendixsubentry{app:predicate-results}
\appendixentry{\ref{app:failure-analysis}}{\hyperref[app:failure-analysis]{Failure-mode analysis}}{\pageref{app:failure-analysis}}
\appendixentry{\ref{app:ablations-diagnostics}}{\hyperref[app:ablations-diagnostics]{Component and sensitivity analysis}}{\pageref{app:ablations-diagnostics}}
\appendixsubentry{app:semantic-steering-ablation}
\appendixsubentry{app:surrogate-cost}
\appendixsubentry{app:horizon-sensitivity}
\appendixsubentry{app:grounding-sensitivity}
\appendixentry{\ref{app:search-scale}}{\hyperref[app:search-scale]{Planning time and search complexity}}{\pageref{app:search-scale}}
\appendixsubentry{app:navigation-planning-time}
\appendixsubentry{app:graph-automaton-sizes}
\appendixsubentry{app:reachable-products}
\appendixsubentry{app:search-accuracy}
\appendixentry{\ref{app:case-studies}}{\hyperref[app:case-studies]{Case studies}}{\pageref{app:case-studies}}
\appendixsubentry{app:decoupled-case-study}
\appendixsubentry{app:predicate-case-studies}
\endgroup

\section{Related work}
\label{app:related}
\paragraph{Automata-based planning and formal synthesis.}
Classical temporal-logic planning constructs a finite transition system, forms its product with an automaton for the LTL specification, and searches for an accepting prefix--suffix run~\cite{kloetzer2008fully,fainekos2005temporal,smith2011optimal,vasile2013sampling}. Some planners sample continuous configurations rather than discretizing the workspace in advance~\cite{karaman2012sampling,luo2021abstraction}; others sample states of a discrete product rather than constructing it in full~\cite{kantaros2018sampling,kantaros2020stylus}. SAGAS uses the same product-search principle with a semantic graph built from fixed offline data.

\paragraph{Motion planning with reusable motions and controls.}
Motion planners can reuse feasible motions or controls to reduce the work required for each new query. State-lattice planners precompute motions between neighboring lattice states~\cite{pivtoraiko2009lattices}, while db-A* allows bounded gaps between successive motion primitives that are subsequently closed by trajectory optimization~\cite{honig2022dbastar}. Bundles of Edges~\cite{shome2021bundles} reuses sampled controls rather than fixed connections: for each goal, it grows a new tree by forward propagation and provides an asymptotic optimality guarantee. SAGAS instead reuses a reachability graph learned from fixed trajectories to plan for new LTL specifications without query-time propagation.

\paragraph{LTL-guided reinforcement learning and reward machines.}
Learning-based approaches often encode logical progress through reward shaping, automaton states, or reward machines~\cite{hasanbeig2018logically,icarte2018reward,bagatelladirected,voloshin2023eventual,shahltl}. Transition-centric decomposition methods also use automaton transitions as reusable subtasks~\cite{miao2026t4nmtd}. These approaches support exploration and policy learning for temporal objectives, typically through online interaction or training for a target specification or task distribution. SAGAS instead learns a reusable low-level executor and handles each new specification through graph search at test time.

\paragraph{Zero-shot LTL generalization and policy composition.}
Zero-shot LTL methods reuse learned behavior under new specifications. Goal-conditioned agents can follow LTL instructions without additional training over the LTL task space~\cite{qiu2023instructing}, while DeepLTL conditions B\"uchi-structure-aware policies on truth-assignment sequences to handle finite- and infinite-horizon formulas~\cite{jackermeierdeepltl}. Related universal policies use Boolean-formula sequences or semantically labelled automata~\cite{jackermeier2026zeroshot,abate2026semantically}. PlatoLTL generalizes across predicate instances by encoding their task parameters in the goal embedding~\cite{cloete2026platoltl}.

Other methods compose subgoal solvers or task primitives. GenZ-LTL decomposes unseen specifications into reach-avoid subgoals solved through safe RL~\cite{guo2025one}. Comp-LTL combines minimum-violation task primitives by constructing a pruned transition system and searching its product with the specification automaton~\cite{bergeron2024compltl}. Skill-transfer and modular policy methods likewise reuse learned primitives across temporal tasks~\cite{liu2024skill}.

SAGAS reuses a latent reachability graph and low-level executor learned from fixed task-agnostic data, handling each specification through task-time semantic graph augmentation and product search. This also separates predicate learning from planning: learned grounding methods map sub-symbolic observations to propositions~\cite{pannacci2026grounding}, whereas SAGAS takes a supplied labeling function and evaluates it on preserved states to construct anchors and soft labels. New predicates can therefore be introduced at task time wherever the offline data provide sufficient support.

\paragraph{Offline goal-conditioned RL and graph-based stitching.}
Hindsight experience replay makes a trajectory useful for goals other than the one originally pursued by relabeling it with a goal the agent achieved~\cite{andrychowicz2017hindsight}. Offline RL methods such as implicit Q-learning instead address how to learn a policy when the data are fixed~\cite{kostrikovoffline}. For goal reaching, contrastive RL estimates goal-conditioned values from state--action and goal embeddings~\cite{eysenbach2022contrastive}; HIQL and foundation policies based on Hilbert representations learn goal-reaching behavior directly from offline trajectories~\cite{park2023hiql,park2024foundation}.
Fragmented demonstrations make long-horizon execution difficult, motivating hierarchical methods that generate subgoals or stitch short trajectories~\cite{shin2023guide,park2025temporal,kim2024stitching}. GAS builds a latent reachability graph from a temporal-distance representation and searches it for single-goal reaching~\cite{baek2025graph}. TTGS likewise combines learned policies with test-time graph search to bridge long horizons~\cite{opryshko2025test}. SAGAS adds a support-preserving semantic interface to the GAS backbone: proposition-grounded anchors and soft labels enable cost-aware B\"uchi product synthesis with positive and negative literals and accepting suffix cycles.

\paragraph{Diffusion-based planning from offline data.}
For goal-reaching tasks, SSD~\cite{kim2024stitching} stitches together sub-trajectories generated by a diffusion model conditioned on the goal and a target value.
DOPPLER~\cite{feng2025diffusion} uses diffusion-generated trajectories as candidate options for LTL tasks. It learns to score each option given the current state and the part of the formula that remains to be satisfied.
In SAGAS, the formula enters after training: the planner searches the product of the labeled reachability graph and the formula's B\"uchi automaton for an accepting prefix--suffix plan.

\section{Algorithmic details}
\label{app:algorithms}

\subsection{Offline graph construction and executor training}
\label{app:offline-backbone}

The offline interface consists of a weighted reachability graph, preserved state support, and a local policy for tracking selected graph nodes. We instantiate the graph and policy with GAS~\cite{baek2025graph}, using a shared temporal-distance representation to define graph connections and policy subgoals. The preserved support links each node to observed physical states, allowing SAGAS to evaluate new predicates after the representation, graph, and policy have been frozen.

\paragraph{Temporal-distance representation.}
The temporal-distance representation measures proximity in terms of reachability: two states that are close in task space may still require many steps to connect. In the learned embedding $\psi:\mathcal{S}\rightarrow\mathcal{H}$, with $\mathcal{H}\subseteq\mathbb{R}^{d_h}$, the distance
\begin{equation}
d_\psi(s,g)=\|\psi(s)-\psi(g)\|_2
\end{equation}
estimates the minimum number of environment steps needed to reach $g$ from $s$. We parameterize the goal-conditioned value as
\begin{equation}
V(s,g)=-d_\psi(s,g),
\end{equation}
and fit $\psi$ to offline transitions with an IQL-style expectile temporal-difference objective~\cite{kostrikovoffline,park2024foundation}.

\paragraph{TD-aware graph construction.}
To reduce the influence of detours in the recorded trajectories, graph construction first selects states whose observed motion is directionally consistent at the local scale $H_{\mathrm{TD}}$. Let $\mathcal{F}(s_t,d)$ denote the first subsequent state on the same trajectory at latent distance at least $d$ from $s_t$. For each $s_t$, we compare two latent displacements along that trajectory: one to $\mathcal{F}(s_t,H_{\mathrm{TD}})$ and the other to the state recorded $H_{\mathrm{TD}}$ steps later. Their cosine similarity defines the temporal-efficiency score
\begin{equation}
\theta_{\mathrm{TE}}(s_t)
=
\mathrm{cos}\!\left(
\psi(\mathcal{F}(s_t,H_{\mathrm{TD}}))-\psi(s_t),
\psi(s_{t+H_{\mathrm{TD}}})-\psi(s_t)
\right).
\end{equation}
We retain $s_t$ for clustering when the two displacement vectors are closely aligned.

A retained state joins the nearest cluster if the latent distance to its representative is at most $H_{\mathrm{TD}}/2$; otherwise it starts a new cluster. Each cluster mean becomes a graph node. Clustering is performed jointly across trajectories, so a node can represent nearby latent states from different recorded fragments. We connect distinct nodes within latent distance $H_{\mathrm{TD}}$ of one another. If the graph is disconnected, we repeatedly add the shortest link from the growing main component to any remaining component.

Each edge $(u,v)\in\mathcal E$ receives weight $w(u,v)=\|u-v\|_2$, the distance between the two cluster means. This weight serves as a surrogate for the number of environment steps needed to move between the represented states. The product planner uses the same cost convention for local edges and links between components.

\paragraph{Support-preserving graph interface.}
A cluster representative cannot be directly evaluated by state-defined predicates. SAGAS preserves raw states from the cluster to supply the corresponding semantic evidence:
\begin{equation}
\mathcal{D}_v\subseteq\{\,s\in\mathcal{D}_{S}\mid \psi(s)\in\mathcal{C}_v\,\}.
\end{equation}
Here $\mathcal{D}_{S}$ is the set of states in the offline dataset, and $\mathcal D_v$ may contain either all states assigned to the cluster or a bounded subset.

\paragraph{Low-level executor training and deployment.}
The low-level policy learns to move toward subgoals specified by their latent direction. For each training state $s_t$, we select $s_{\mathrm{sub}}=\mathcal{F}(s_t,H_{\mathrm{TD}})$ from the same recorded trajectory, using the graph's temporal-distance scale rather than a fixed number of steps. The policy receives the normalized direction toward this subgoal:
\begin{equation}
\vec h_{\mathrm{dir}} =
\frac{\psi(s_{\mathrm{sub}})-\psi(s_t)}
{\|\psi(s_{\mathrm{sub}})-\psi(s_t)\|}.
\end{equation}
We train the direction-conditioned critic with the intrinsic reward
\begin{equation}
r^{\mathrm{dir}}(s_t,s_{t+1},\vec h_{\mathrm{dir}})
=
\left\langle \psi(s_{t+1})-\psi(s_t),\vec h_{\mathrm{dir}}\right\rangle,
\end{equation}
which measures progress along the requested direction in the embedding. The DDPG+BC actor objective combines high estimated value with a behavior-cloning term that keeps actions close to those in the dataset, following the same regularization principle as TD3+BC~\cite{fujimoto2021minimalist}. Unless noted otherwise, we use the training settings of GAS~\cite{baek2025graph}.

At execution time, the selected graph waypoint $v$ replaces the training subgoal. The frozen policy receives the same form of directional input, recomputed from the current state at each step:
\begin{equation}
\vec h_{\mathrm{dir}}(s_t,v)
=
\frac{v-\psi(s_t)}
{\|v-\psi(s_t)\|_2}.
\end{equation}

\subsection{Soft-label screening for negative literals}
\label{app:soft-labels}

For an original graph node $v$, the soft label measures the weighted fraction of its preserved states that satisfy proposition $\ell$:
\begin{equation}
\label{eq:soft-label-weighted}
    \widehat{P}(\ell\mid v)
    =
    \frac{
    \sum_{s\in\mathcal{D}_v}
    K_v(s)\,\mathbf{1}\{\ell\in L_\phi(s)\}
    }{
    \sum_{s\in\mathcal{D}_v}
    K_v(s)
    },
\end{equation}
where $K_v(s)\geq0$ is the weight of support state $s$. For region predicates, the indicator tests whether $\Pi(s)\in\mathcal R_\ell$. When a guard term contains $\neg\ell$, the planner admits $v$ only if the estimate has a positive denominator and does not exceed $\tau_{\mathrm{soft}}$. At anchors, the planner instead checks the complete witness label set and rejects any anchor whose witness satisfies a forbidden proposition. Appendix~\ref{app:successor-oracle} combines these rules in the product successor relation.

By default, we use $K_v(s)\equiv1$, giving each preserved state equal weight within the cluster. The weighted definition also allows greater emphasis on states near the representative, for example through $K_v(s)=K_\sigma(\|\psi(s)-v\|_2)$ with $K_\sigma(r)=\exp(-r^2/(2\sigma^2))$ and $\sigma>0$.

\subsection{Anchor witness construction}
\label{app:anchor-nodes}

Anchors provide concrete witnesses for positive literals. A cluster may contain satisfying states even when tracking its representative reaches a state where the proposition is false. An anchor instead targets the embedding of a specific observed state whose labels satisfy the requirement.

For each proposition $\ell$, SAGAS scans the offline dataset and collects candidate states
\[
\mathcal{S}_{\ell}^{\mathrm{cand}}
=
\{\,s\in\mathcal{D}_{S}\mid \ell\in L_\phi(s)\,\}.
\]
If the pool is empty, the planner marks $\ell$ unavailable as a positive requirement under the current dataset.

For a nonempty pool, retrieval continues until $N_s$ connected anchors have been accepted, the per-proposition candidate-attempt limit is reached, or the pool is exhausted. Each candidate is embedded as
\[
a_{s_j^{(\ell)}}=\psi(s_j^{(\ell)}),
\qquad
s_j^{(\ell)}\in\mathcal{S}_{\ell}^{\mathrm{cand}},
\]
and tested for connections to the offline graph at the same temporal-distance scale:
\[
\mathcal{N}_{\mathrm{TD}}(a_{s_j^{(\ell)}})
=
\{\,v\in\mathcal{V}\mid \|a_{s_j^{(\ell)}}-v\|_2\le H_{\mathrm{TD}}\,\}.
\]
A candidate with a nonempty neighbor set is inserted as an anchor and connected to these neighbors, with edge weights given by their latent distances. Its support contains the witness state alone, and its label set is that state's complete assignment:
\[
a_{s_j^{(\ell)}}\in\mathcal{A}_{\ell},\qquad
\mathcal{D}_{a_{s_j^{(\ell)}}}=\{s_j^{(\ell)}\},\qquad
\lambda_{\mathcal{A}}(a_{s_j^{(\ell)}})=L_\phi(s_j^{(\ell)}).
\]
The full assignment allows an anchor to belong to several per-proposition sets, supporting overlapping predicates. Retrieval also includes propositions that occur only negatively, whose anchors can guide avoidance during execution. Candidates without temporal-distance neighbors are discarded; if none connects, the positive requirement remains unavailable.

\paragraph{Joint-guard retrieval.}
When a guard term requires several propositions simultaneously, separate witnesses for each proposition are insufficient. For a normalized guard term $\eta$, let $P_\eta$ and $N_\eta$ denote its sets of required and forbidden propositions. The joint candidate pool is
\[
\mathcal S_\eta^{\mathrm{cand}}
=\{s\in\mathcal D_S:P_\eta\subseteq L_\phi(s),\;
N_\eta\cap L_\phi(s)=\varnothing\}.
\]
For terms containing positive literals, SAGAS retrieves from this pool using the connection rule above. Accepted anchors can be reused by compatible guard terms and included in each applicable per-proposition set, so the final $\mathcal A_\ell$ may exceed the initial budget $N_s$. If no connected witness is found, the planner disables that term while retaining any supported alternatives. Terms without positive literals require no positive anchor.

\subsection{Support-aware automaton preprocessing}
\label{app:automaton-preprocess}

SAGAS translates $\phi$ with \texttt{ltl2ba}~\cite{gastin2001fast} and then restricts future transitions to those supported by the semantic graph. Before this restriction, the original nondeterministic B\"uchi automaton is
\[
\mathcal{B}_{\phi}=(Q,Q_0,\Sigma,\delta,F),
\]
where $Q$ is the finite state set, $Q_0\subseteq Q$ the initial-state set, $\Sigma=2^{\mathcal{AP}_\phi}$ the alphabet of proposition assignments, $\delta:Q\times\Sigma\to 2^Q$ the transition function, and $F\subseteq Q$ the accepting set. Under standard B\"uchi semantics~\cite{baier2008principles}, a run $q_0q_1\ldots$ over an infinite word $\sigma_0\sigma_1\ldots\in\Sigma^\omega$ satisfies $q_0\in Q_0$ and $q_{t+1}\in\delta(q_t,\sigma_t)$ for all $t\ge0$. It is accepting when it visits $F$ infinitely often, i.e., $\mathrm{Inf}(q_0q_1\ldots)\cap F\neq\emptyset$.

Boolean guards represent the transitions without enumerating every proposition assignment. For each state pair $(q,q')$, the guard $g_{q,q'}$ is satisfied by exactly the assignments that enable the transition:
\[
\mathsf{Sat}(g_{q,q'})
=
\{\sigma\in\Sigma\mid q'\in\delta(q,\sigma)\}.
\]
Thus, $q'\in\delta(q,\sigma)$ iff $\sigma\models g_{q,q'}$. To check alternative ways of satisfying a guard independently, SAGAS converts it into disjunctive normal form:
\[
g_{q,q'} \equiv \bigvee_{\eta\in\mathrm{DNF}(g_{q,q'})}\eta,
\qquad
\eta=
\bigwedge_{\ell\in P_\eta}\ell
\wedge
\bigwedge_{\ell\in N_\eta}\neg\ell,
\]
where $P_\eta$ and $N_\eta$ contain the propositions required to be true and false, respectively. The successor oracle and runtime monitor use these same sets to check the selected term.

Preprocessing treats each satisfiable DNF term as a separate transition and discards contradictory terms. A term containing positive literals is retained only when a connected anchor witnesses the complete term; any term removed for lack of support is reported to the user. For each state pair, the retained guard $\widetilde g_{q,q'}$ is the disjunction of the surviving terms, giving $\widetilde g_{q,q'}\Rightarrow g_{q,q'}$. This restriction applies to movement and dwell guards after the initial observation.

The initial assignment is already known from $s_0$, so the original NBA processes it using $Q_0$ and $\delta$ without requiring an offline anchor. State pruning then considers reachability from the resulting automaton states and the existence of return cycles through accepting states. The resulting support-restricted automaton $\widetilde{\mathcal B}_\phi$ has state set $\widetilde Q\subseteq Q$ and accepting set $\widetilde F=F\cap\widetilde Q$. Appendices~\ref{app:graph-automaton-sizes} and~\ref{app:reachable-products} report the automaton sizes before preprocessing and during product search.

Algorithm~\ref{alg:semantic-augmentation} summarizes semantic graph augmentation using the normalized guard terms.

\begin{algorithm}[!htbp]
\caption{Task-time semantic graph augmentation}
\label{alg:semantic-augmentation}
\small
\begin{algorithmic}[1]
\REQUIRE Latent graph $\mathcal{H}_{\mathrm{graph}}=(\mathcal{V},\mathcal{E},w)$, preserved supports $\{\mathcal D_v\}$, offline states $\mathcal D_S$, embedding $\psi$, initial state $s_0$, labeling function $L_\phi$, normalized guard terms $\mathcal G_\phi$, temporal-distance scale $H_{\mathrm{TD}}$, initial per-proposition anchor budget $N_s$ and candidate-attempt limit
\ENSURE Semantic graph $\mathcal{H}^{\mathrm{sem}}_{\phi}$, start node $v_0$, anchor sets and witness label sets, soft-label estimates, unavailable propositions $\mathcal{U}_{\mathrm{unav}}$
\STATE Initialize anchor sets $\mathcal{A}_\ell\leftarrow\emptyset$ for all $\ell\in\mathcal{AP}_\phi$
\STATE Estimate soft labels $\widehat{P}(\ell\mid v)$ for all original graph nodes $v$ and propositions $\ell$ using their raw supports and $L_\phi$
\FORALL{$\ell\in\mathcal{AP}_\phi$}
    \STATE Retrieve candidate states from $\mathcal{D}_{S}$ satisfying $\ell$ under $L_\phi$
    \WHILE{$|\mathcal{A}_\ell|<N_s$, untried candidates remain, and the candidate-attempt limit is not reached}
        \STATE Embed the next candidate as $a_{s_j^{(\ell)}}=\psi(s_j^{(\ell)})$
        \IF{$a_{s_j^{(\ell)}}$ has temporal-distance neighbors within $H_{\mathrm{TD}}$}
            \STATE Insert it as an anchor with singleton support, witness label set $L_\phi(s_j^{(\ell)})$, and local graph edges within $H_{\mathrm{TD}}$
            \STATE $\mathcal A_\ell\leftarrow\mathcal A_\ell\cup\{a_{s_j^{(\ell)}}\}$
        \ENDIF
    \ENDWHILE
\ENDFOR
\FORALL{DNF terms $\eta$ containing positive literals}
    \STATE Retrieve connected witnesses satisfying all positive and no negative literals
    \STATE Insert reusable joint anchors and update the applicable anchor sets $\mathcal A_\ell$; record $\eta$ unsupported if no witness connects
\ENDFOR
\STATE Set $\mathcal{U}_{\mathrm{unav}}\leftarrow\{\ell\in\mathcal{AP}_\phi\mid\mathcal A_\ell=\emptyset\}$ from the final anchor sets
\STATE Embed $s_0$ as $v_0=\psi(s_0)$ and connect it using the same temporal-distance scale
\IF{$v_0$ has no incident graph edge}
    \STATE Connect $v_0$ to the nearest existing graph node as an initialization heuristic
\ENDIF
\STATE Form $\mathcal{H}^{\mathrm{sem}}_{\phi}$ from the augmented graph, anchors, soft labels, and $v_0$
\STATE \textbf{return} $\mathcal{H}^{\mathrm{sem}}_{\phi}$ and its semantic auxiliary data
\end{algorithmic}
\end{algorithm}

\subsection{Semantic product transition rules}
\label{app:successor-oracle}

The successor oracle combines graph connectivity with the semantic checks that enable automaton transitions. It generates movement successors on demand during search and also checks whether an accepting endpoint admits a dwell suffix at the same waypoint, as described in Section~\ref{sec:planning}.

Let $\mathcal{H}^{\mathrm{sem}}_{\phi}=(\mathcal{V}_{\phi},\mathcal{E}_{\phi},w_{\phi})$ be the semantic graph and $\widetilde{\mathcal B}_\phi$ the support-restricted automaton from Appendix~\ref{app:automaton-preprocess}, with state set $\widetilde Q$, accepting set $\widetilde F$, and retained guards $\widetilde g_{q,q'}$. Semantic evidence depends on the node type: original nodes $\mathcal V$ use soft labels $\widehat P$, inserted anchors $\mathcal{A}=\bigcup_{\ell\in\mathcal{AP}_\phi}\mathcal{A}_{\ell}$ use witness assignments $\lambda_{\mathcal A}$, and the temporary start node $v_0$ uses $L_\phi(s_0)$. The implicit product system is
\[
\mathcal{P}_{\phi}=(\mathcal{Z}_{\phi},\mathcal{E}_{\mathcal{P}},c_{\mathcal{P}}),
\qquad
\mathcal{Z}_{\phi}=\mathcal{V}_{\phi}\times\widetilde Q .
\]
Its initial states pair $v_0$ with the automaton states reached after processing $L_\phi(s_0)$:
\[
Q_{\mathrm{init}}
=
\{q'\in\widetilde Q\mid \exists q_0\in Q_0,\ q'\in\delta(q_0,L_\phi(s_0))\},
\qquad
\mathcal{Z}_0=\{(v_0,q):q\in Q_{\mathrm{init}}\}.
\]
A movement from $z=(v,q)$ to $z'=(v',q')$ requires a semantic-graph edge and at least one retained guard term enabled at the destination:
\begin{equation}
\label{eq:product-transition}
(z,z')\in\mathcal{E}_{\mathcal{P}}^{\mathrm{mov}}
\Longleftrightarrow
(v,v')\in\mathcal{E}_{\phi}
\;\wedge\;
\exists \eta\in\mathrm{DNF}(\widetilde g_{q,q'})
\text{ such that }
\mathrm{Enable}_{\phi}(v,v',\eta)=\mathrm{true}.
\end{equation}
The cost of an enabled movement transition is inherited from the semantic graph:
\begin{equation}
c_{\mathcal{P}}(z,z')=w_{\phi}(v,v').
\end{equation}
For a conjunctive term $\eta$ with required and forbidden proposition sets $P_\eta$ and $N_\eta$, the enablement check separates the positive and negative requirements:
\begin{equation}
\mathrm{Enable}_{\phi}(v,v',\eta)
=
\operatorname{PosOK}(v',\eta)
\wedge
\operatorname{NegOK}(v',\eta).
\end{equation}
The negative-literal check depends on the type of successor node:
\begin{equation}
\operatorname{NegOK}(v',\eta)
=
\begin{cases}
\mathrm{true}, & N_\eta=\emptyset,\\
\displaystyle
\bigwedge_{\ell\in N_\eta}
\left(\widehat{P}(\ell\mid v')\le\tau_{\mathrm{soft}}\right),
& v'\in\mathcal{V}^{+},\\[1.2ex]
\lambda_{\mathcal{A}}(v')\cap N_\eta=\emptyset,
& v'\in\mathcal{A},\\
\displaystyle
L_\phi(s_0)\cap N_\eta=\emptyset,
& v'=v_0,\\
\mathrm{false}, & \text{otherwise}.
\end{cases}
\end{equation}
Here $\mathcal V^{+}$ contains original graph nodes with $\sum_{s\in\mathcal D_v}K_v(s)>0$, for which the soft-label estimate is defined. A guard term without negative literals passes this check automatically. Positive literals, in contrast, require a connected anchor whose witness satisfies them all:
\[
\operatorname{PosOK}(v',\eta)
=[P_\eta=\varnothing]\;\vee\;
[v'\in\mathcal A\ \land\ P_\eta\subseteq\lambda_{\mathcal A}(v')].
\]
Together, the two checks evaluate the complete term at an anchor and apply the soft-label screen at an original node.

At an accepting endpoint, the additional suffix-only dwell relation is
\[
\mathcal E_{\mathcal P}^{\mathrm{dwell}}
=\left\{(z,z):\begin{array}{l}
z=(u,q),\ q\in\widetilde F,\ \exists\eta\in\mathrm{DNF}(\widetilde g_{q,q}):\\
\operatorname{PosOK}(u,\eta)\wedge\operatorname{NegOK}(u,\eta)
\end{array}\right\}.
\]
A dwell keeps the same waypoint and has zero graph cost, without requiring a self-edge in the semantic graph. The complete relation is $\mathcal E_{\mathcal P}=\mathcal E_{\mathcal P}^{\mathrm{mov}}\cup\mathcal E_{\mathcal P}^{\mathrm{dwell}}$: prefix and nonempty return searches expand movement edges, while dwell is tested separately at accepting endpoints.

During expansion of $(v,q)$, the oracle examines its semantic-graph neighbors and outgoing retained guards, returning the successors that satisfy Eq.~\eqref{eq:product-transition}. Prefix and return searches share this procedure, and the dwell test reuses its semantic checks at the current waypoint.

\subsection{Prefix--suffix search procedure}
\label{app:prefix-suffix-search}

SAGAS selects a complete lasso by comparing prefixes together with their return suffixes. It first collects accepting prefixes, then searches for a suffix at each retained endpoint, using the same successor oracle for both stages.

\paragraph{Prefix candidate collection.}
The prefix search runs A* from $\mathcal Z_0$ toward the accepting set $\mathcal{Z}_F=\mathcal{V}_{\phi}\times\widetilde F$, using product-edge costs $c_{\mathcal P}$. When an accepting state $z_f=(u,q_f)$ is removed from the priority queue for expansion, its predecessor chain yields a candidate prefix. Retaining up to $K$ distinct accepting endpoints allows the final choice to account for differences in suffix cost. Since the search stores one predecessor chain per product state, each endpoint contributes at most one prefix.

\paragraph{Suffix cycle construction.}
At each retained endpoint $z_f=(u,q_f)$, an enabled dwell provides a zero-cost graph suffix. If no dwell is available, Dijkstra search starts from the enabled movement successors of $z_f$ and seeks the lowest-cost nonempty path back to $z_f$. A prefix becomes a complete candidate only when one of these suffixes is found.

\paragraph{Candidate selection.}
Each complete candidate $\rho=(\rho_{\mathrm{pre}},\rho_{\mathrm{suf}})$ retains its selected guard terms and has waypoint projection $\Gamma=\operatorname{proj}_{\mathcal V_\phi}(\rho)$. Following the prefix--suffix objective~\cite{luo2021abstraction}, SAGAS compares candidates by
\[
J_{\mathrm{graph}}(\Gamma)
=
\lambda\,\mathrm{Cost}(\Gamma_{\mathrm{pre}})
+
(1-\lambda)\,\mathrm{Cost}(\Gamma_{\mathrm{suf}}),
\]
where $\mathrm{Cost}$ sums semantic-graph edge weights for movements and assigns zero to a dwell. The planner returns the lowest-cost candidate under this objective.

The budget $K$ balances search effort against exploring additional accepting endpoints that may yield a lower-cost complete lasso. Appendix~\ref{app:search-accuracy} evaluates this trade-off by comparing feasibility, lasso cost, and runtime across candidate budgets and against exhaustive finite-product search.

\paragraph{Prefix-search heuristic.}
At a product state $(v,q)$, the A* heuristic estimates the remaining cost of reaching an accepting product state by considering the first positive requirement along a possible path. To identify these requirements, let $\mathcal C_0(q)$ contain $q$ and the states reachable from it in the preprocessed automaton using only guard terms with no positive literals. From these states, we collect into $\mathcal B_+(q)$ the propositions appearing positively in outgoing guard terms whose destination can still reach an accepting state. These propositions are candidates for the first positive requirement on a path to acceptance.

If $q\in\widetilde F$ or $\mathcal C_0(q)\cap\widetilde F\neq\emptyset$, acceptance is possible in the automaton without another positive requirement, and we use $h(v,q)=0$. We also use zero when $\mathcal B_+(q)=\emptyset$. Otherwise, the heuristic is the distance to the nearest anchor for any of the candidate propositions:
\begin{equation}
    h(v,q)=\min_{\ell \in \mathcal{B}_+(q)} \operatorname{dist}_{\phi}(v,\mathcal{A}_\ell),
\end{equation}
where $\operatorname{dist}_{\phi}(v,\mathcal A_\ell)$ is the shortest-path distance from $v$ to any anchor in $\mathcal A_\ell$, computed on the semantic graph without automaton constraints. These distances are precomputed for each relevant anchor set and reused during search.

When the heuristic is nonzero, every accepting product path must eventually take a first transition requiring a positive proposition $\ell\in\mathcal B_+(q)$. That transition ends at an anchor in $\mathcal A_\ell$, so the graph cost accumulated by then is at least $\operatorname{dist}_{\phi}(v,\mathcal A_\ell)$ and hence at least $h(v,q)$. The remaining edges have nonnegative costs, making $h(v,q)$ a lower bound on the total cost of reaching acceptance. The zero cases satisfy this bound directly, so the heuristic is admissible.

Algorithm~\ref{alg:product-search} gives the complete prefix--suffix search procedure.

\begin{algorithm}[!htbp]
\caption{Product-space prefix--suffix search}
\label{alg:product-search}
\small
\begin{algorithmic}[1]
\REQUIRE Semantic graph $\mathcal{H}^{\mathrm{sem}}_{\phi}=(\mathcal{V}_{\phi},\mathcal{E}_{\phi},w_{\phi})$ with $\{\mathcal A_\ell\}$, $\lambda_{\mathcal A}$, $\widehat P$, and start node $v_0$; original NBA $\mathcal{B}_{\phi}$, initial label $L_\phi(s_0)$, unavailable requirements, soft-label threshold $\tau_{\mathrm{soft}}$, prefix budget $K$, trade-off $\lambda$
\ENSURE Product plan $\rho^\star$, including product states and guard terms, with waypoint projection $\operatorname{proj}_{\mathcal V_\phi}(\rho^\star)$, or failure
\STATE $Q_{\mathrm{aligned}}\leftarrow\bigcup_{q_0\in Q_0}\delta(q_0,L_\phi(s_0))$
\STATE Restrict future guards by support, retain aligned initial states, and precompute $h(v,q)$
\STATE Set $Q_{\mathrm{init}}=Q_{\mathrm{aligned}}\cap\widetilde Q$ and $\mathcal Z_0=\{(v_0,q):q\in Q_{\mathrm{init}}\}$
\STATE Run A* on the implicit semantic product graph from $\{(v_0,q):q\in Q_{\mathrm{init}}\}$, retaining up to $K$ accepting prefix candidates
\STATE Initialize candidate plan set $\mathcal C\leftarrow\emptyset$
\FORALL{product prefix $\rho_{\mathrm{pre}}$ ending at $z_f=(u,q_f)$}
    \STATE Initialize $\rho_{\mathrm{suf}}\leftarrow\mathrm{none}$
    \IF{a feasible suffix-only symbolic dwell option exists at $z_f$}
    \STATE Set $\rho_{\mathrm{suf}}$ to the dwell at $z_f$, retaining the enabled self-loop guard
\ELSE
    \STATE Search the same implicit product graph for a nonempty return cycle through $z_f$ and set $\rho_{\mathrm{suf}}$ with its guard trace if one is found
\ENDIF
    \IF{$\rho_{\mathrm{suf}}$ is found}
        \STATE Add $(\rho_{\mathrm{pre}},\rho_{\mathrm{suf}})$ to $\mathcal C$
    \ENDIF
\ENDFOR
\IF{$\mathcal C=\emptyset$}
    \STATE \textbf{return} no accepting lasso found within the candidate budget
\ENDIF
\STATE Select $\rho^\star\in\arg\min_{\rho\in\mathcal C}J_{\mathrm{graph}}(\operatorname{proj}_{\mathcal V_\phi}(\rho))$
\STATE \textbf{return} $\rho^\star$ with its waypoint projection $\operatorname{proj}_{\mathcal V_\phi}(\rho^\star)$ and selected guard terms
\end{algorithmic}
\end{algorithm}

\subsection{Plan execution and runtime monitoring}
\label{app:execution-details}

The frozen low-level policy follows the waypoints of $\rho^\star$, while the monitor uses the observed trajectory to decide when to advance along the product plan. The plan retains its automaton states and selected guard terms alongside the waypoint projection $\Gamma=\operatorname{proj}_{\mathcal V_\phi}(\rho^\star)$. Repeating the suffix unfolds this projection into
\[
\Gamma_{\mathrm{pre}}\oplus\Gamma_{\mathrm{suf}}^\omega
=(v_0,v_1,\ldots),
\]
where $v_0$ is the start.

\paragraph{Waypoint selection and tracking.}
Let $k\ge1$ index the next unconfirmed waypoint and $k'\ge k$ the selected tracking target. Waypoints inherited from the offline graph guide intermediate motion and may be skipped when a later target lies within $H_{\mathrm{TD}}$. Selected anchors and accepting endpoints are mandatory semantic checkpoints that the tracker cannot bypass. For the current state $s_t$ and target $v_{k'}$, the nominal tracking displacement is
\begin{equation}
\Delta_t(s_t,v_{k'})=v_{k'}-h_t,
\end{equation}
where $h_t=\psi(s_t)$ denotes the current latent state. This displacement supplies the nominal direction to the frozen low-level policy. The monitor confirms the target using the checks below before advancing $k$ to $k'+1$; the indices continue across repeated suffix cycles.

\paragraph{Monitoring and progress confirmation.}
While the policy tracks a selected waypoint, the monitor follows the automaton progress induced by the observed state labels. Consider a segment from the last confirmed product state $(v_i,q_i)$ to the selected target $(v_j,q_j)$, where $j$ may exceed $i+1$ if intermediate waypoints are skipped. Starting from $S_i(t_i)=\{q_i\}$ after processing the source label, the monitor processes each subsequent observation using the original NBA:
\[
S_i(t+1)=\bigcup_{q\in S_i(t)}\delta(q,L_\phi(s_{t+1})).
\]
Here, $t_i$ is the source confirmation time, and $S_i(t)$ contains the automaton states reachable from $q_i$ under the labels observed since that confirmation.

During tracking, let $\mathcal N_{\mathrm{act}}(t)$ contain the propositions required to be false by the currently active guards of the plan. The monitor checks
\begin{equation}
L_\phi(s_t)\cap\mathcal N_{\mathrm{act}}(t)=\emptyset.
\end{equation}
This set is updated as the automaton progresses and also supplies the forbidden labels for guard-aware steering. A violation of these active negative requirements terminates execution.

Target confirmation combines physical arrival with the automaton progress recorded in $S_i$. For an anchor target $a_s$ with selected incoming guard $\eta$, the reach and positive-label checks are
\[
\|\psi(s_t)-a_s\|_2 < \varepsilon_{\mathrm{anc}},
\qquad
P_{\eta}\subseteq L_\phi(s_t),
\]
where $P_\eta$ contains the positive literals of $\eta$. A non-anchor target uses the waypoint-tracking reach condition. In either case, confirmation at time $t_j$ requires the reach and guard checks to pass and $q_j\in S_i(t_j)$, so the observed segment connects the planned automaton endpoints. The next segment starts from $\{q_j\}$ and processes only subsequent observations, without consuming the endpoint label again. If the target remains unconfirmed when the local tracking budget expires, execution terminates. Lemma~\ref{lem:monitor-soundness} establishes that successive confirmations preserve a valid run of the original NBA over the observed labels.

After completing the prefix, the tracker repeats the suffix under the same monitoring rules. Each cycle is confirmed only when the suffix endpoint is reached and the labels observed during that cycle admit a return to the planned accepting automaton state. A dwell cycle is likewise confirmed using a new physical observation. Appendix~\ref{app:conditional-proof} formalizes the resulting execution guarantee.

\paragraph{Guard-aware steering.}
\label{app:runtime-steering}
To help the executor avoid forbidden regions between waypoints in navigation, guard-aware steering adds bounded repulsion to the nominal displacement $\Delta_t(s_t,v_{k'})$. Anchors of active forbidden propositions serve as avoidance landmarks. For each $\ell\in\mathcal N_{\mathrm{act}}(t)$ with $\mathcal A_\ell\neq\emptyset$, the nearest such anchor is
\begin{equation}
v_{\ell}^{\mathrm{near}}(h_t)
=
\arg\min_{v\in\mathcal{A}_{\ell}}\|h_t-v\|_2 .
\end{equation}
The contributions from these anchors form the repulsion vector
\begin{equation}
r_t =
\sum_{\substack{\ell\in\mathcal N_{\mathrm{act}}(t)\\ \mathcal{A}_{\ell}\neq\emptyset}}
\beta
\left(1-\frac{\|h_t-v_{\ell}^{\mathrm{near}}(h_t)\|_2}{\rho_{\mathrm{rep}}}\right)_{+}^{p}
\frac{h_t-v_{\ell}^{\mathrm{near}}(h_t)}
{\|h_t-v_{\ell}^{\mathrm{near}}(h_t)\|_2+\epsilon},
\end{equation}
where $\rho_{\mathrm{rep}}$ is the repulsion radius, $\beta$ sets its strength, $p$ controls how sharply it increases near an anchor, and $(x)_+=\max(x,0)$. Clipping this vector to a maximum norm $r_{\max}$ gives the policy direction
\begin{equation}
\widetilde d_t =
\mathrm{normalize}\!\left(\Delta_t(s_t,v_{k'})+\mathrm{clip}_{r_{\max}}(r_t)\right),
\end{equation}
where clipping rescales $r_t$ only when its norm exceeds $r_{\max}$, and normalization uses a small numerical tolerance. The resulting direction combines motion toward the target with bounded deflection away from forbidden anchors. When steering is disabled or no active forbidden proposition has anchors, we set $r_t=0$ and the policy receives the normalized waypoint displacement.

Algorithm~\ref{alg:execution} gives the full execution procedure.

\begin{algorithm}[!htbp]
\caption{Plan execution and runtime monitoring}
\label{alg:execution}
\small
\begin{algorithmic}[1]
\REQUIRE Product plan $\rho^\star$ with waypoint projection $\Gamma=\operatorname{proj}_{\mathcal V_\phi}(\rho^\star)$, planned automaton states $\{q_i\}$ and guard terms $\{\eta_i\}$, original NBA $\mathcal B_\phi$, anchors $\mathcal A$, embedding $\psi$, frozen low-level policy $\pi_{\mathrm{low}}$, thresholds $H_{\mathrm{TD}},\varepsilon_{\mathrm{anc}}$, labeling function $L_\phi$, local budget $H_{\mathrm{low}}$, steering setting
\ENSURE Executed trajectory $\tau$ and execution outcome, including completion and violation statistics
\STATE Unfold the prefix followed by repeated suffix cycles into waypoint and guard streams, starting at $(v_0,q_0)$
\STATE For a dwell suffix, repeat the endpoint and its guard at successive waypoint indices
\STATE Initialize next unconfirmed index $k\leftarrow1$, selected target index $k'\leftarrow1$, and local tracking timer $c\leftarrow0$
\STATE Initialize $S\leftarrow\{q_0\}$, where $(v_0,q_0)$ starts $\rho^\star$ after processing $L_\phi(s_0)$
\FOR{$t=0,1,\ldots$ until timeout, violation, or the prescribed finite suffix-evaluation criterion is met}
    \STATE Embed current state $h_t\leftarrow\psi(s_t)$
    \STATE On each new observation after the last confirmation, update $S\leftarrow\bigcup_{q\in S}\delta(q,L_\phi(s_t))$
    \STATE Determine active obligations from automaton progress; terminate if no feasible branch remains
    \STATE Let $\mathcal{N}_{\mathrm{act}}(t)$ be the union of forbidden labels in the active guard obligations
    \IF{$L_\phi(s_t)\cap\mathcal{N}_{\mathrm{act}}(t)\neq\emptyset$}
        \STATE Record a semantic violation and terminate the rollout as failed
    \ENDIF
    \STATE If the target's reach and guard checks pass but $q_{k'}\notin S$, report an unrealized transition and terminate
    \STATE On a fresh observation, count a suffix cycle when its accepting endpoint passes the reach, guard, and automaton checks
    \IF{target anchor $v_{k'}$ is close and all its required positive labels are verified under $L_\phi(s_t)$}
        \STATE Mark the checkpoint verified; set $k\leftarrow k'+1$ and reset $S\leftarrow\{q_{k'}\}$, $c\leftarrow0$
    \ELSIF{non-anchor target $v_{k'}$ is reached and all preceding semantic checkpoints are confirmed}
        \STATE Mark waypoint progress verified; set $k\leftarrow k'+1$ and reset $S\leftarrow\{q_{k'}\}$, $c\leftarrow0$
    \ENDIF
    \STATE Select the furthest waypoint $k'\ge k$ within $H_{\mathrm{TD}}$, up to and including the next pending semantic checkpoint; default to $k'=k$
    \STATE Compute $\widetilde d_t$ by normalizing waypoint displacement plus bounded repulsion when steering is enabled; otherwise normalize waypoint displacement alone
    \STATE Apply action $a_t\sim\pi_{\mathrm{low}}(\cdot\mid s_t,\widetilde d_t)$ and observe $s_{t+1}$
    \STATE Increment $c$; if $c>H_{\mathrm{low}}$ before target confirmation, record the current product plan as unrealized and terminate as failed
    \STATE Record task-space labels, anchor completions, suffix-cycle completions, and any violation
\ENDFOR
\STATE \textbf{return} trajectory $\tau$ and execution outcome
\end{algorithmic}
\end{algorithm}

\subsection{Complexity analysis}
\label{app:complexity}

Task-time computation has two main components: grounding predicates on offline states and searching the semantic product graph.

Semantic preprocessing evaluates predicates on the preserved node supports and the full anchor-retrieval pool. Let $N_{\mathrm{sup}}=|\bigcup_{v\in\mathcal V}\mathcal D_v|$ and $N_{\mathrm{data}}=|\mathcal D_S|$ denote their respective sizes. Soft-label computation requires $O(|\mathcal{AP}_\phi|N_{\mathrm{sup}})$ predicate calls, while labeling the full retrieval pool requires $O(|\mathcal{AP}_\phi|N_{\mathrm{data}})$ calls, reusing support assignments where available.

Let $G$ count the normalized guard terms examined during joint-witness retrieval, with at most $b_\eta$ literals per term. Given the state assignments, a direct scan requires at most $O(Gb_\eta N_{\mathrm{data}})$ label checks. Checking each candidate's connectivity adds a range search and processing of its local neighbors. The anchor budget, candidate-attempt limit, and pool size bound the number of candidates examined. After DNF conversion, guard preprocessing is linear in the normalized representation.

For product search, let $n=|\mathcal{V}_{\phi}|$ and $m=|\mathcal{E}_{\phi}|$ denote the semantic-graph size, $r=|\widetilde Q|$ the number of preprocessed automaton states, and $d_B=\max_{q\in\widetilde Q}|\mathrm{Out}(q)|$ the maximum number of outgoing guard terms from an automaton state. The product contains at most
\begin{equation}
    N_{\mathcal{P}} \le n r
\end{equation}
states. Each movement successor combines a graph edge with an outgoing guard term, so the number $M_{\mathcal P}$ of generated movement edges is bounded by
\begin{equation}
    M_{\mathcal{P}} \le m r d_B .
\end{equation}
With a binary heap, one A* or Dijkstra search has worst-case time $O(M_{\mathcal{P}}\log N_{\mathcal{P}})$, plus guard-evaluation costs. Guard normalization, support pruning, and soft-label computation precede search, so each guard check uses stored values and costs time proportional to its number of literals.

The prefix heuristic requires distances to $b$ relevant anchor sets. These can be precomputed by $b$ multi-source shortest-path searches on the semantic graph, taking $O(b\,m\log n)$ time with a binary heap.

SAGAS runs one prefix search and up to $K$ suffix return searches, one for each retained accepting endpoint. The resulting worst-case search time is
\begin{equation}
    O\!\left((K+1) M_{\mathcal{P}}\log N_{\mathcal{P}}\right).
\end{equation}
In practice, the graph-distance heuristic can reduce product-state expansion, and an enabled dwell removes the need for a return search at that endpoint. Dwell testing adds no product states and checks at most $d_B$ self-loop terms per endpoint, separately from the movement-edge bound.

For a fixed task and network, the computational cost of execution and monitoring grows linearly with the rollout horizon: each step computes the state embedding, updates the monitor and tracking target, and queries the policy, with guard-aware steering applied when enabled.

\FloatBarrier
\section{Soundness of synthesis and monitored execution}
\label{app:conditional-proof}

The proof of Proposition~\ref{prop:conditional-main} connects the synthesized plan to the labels observed during execution. We first show that the plan's guard sequence admits accepting runs of the original NBA. We then show that the monitor preserves a valid run over the actual observations whenever it confirms a target. Infinitely many confirmed suffix cycles make this run accepting, establishing satisfaction of the task formula.

Throughout, $L_\phi$ is the task's labeling function and $\mathcal B_\phi=(Q,Q_0,\Sigma,\delta,F)$ is the original NBA, with $\mathcal L(\mathcal B_\phi)=\mathcal L(\phi)$ and $\Sigma=2^{\mathcal{AP}_\phi}$. The support-restricted automaton retains states $\widetilde Q\subseteq Q$ and accepting states $\widetilde F=F\cap\widetilde Q$.

\paragraph{Soundness of the planned guard sequence.}
Let $\rho=(\rho_{\mathrm{pre}},\rho_{\mathrm{suf}})$ be a returned product plan, unfolded as
\[
\rho_{\mathrm{pre}}\oplus\rho_{\mathrm{suf}}^\omega=(v_0,q_0)(v_1,q_1)\cdots.
\]
The prefix ends at $(v_p,q_p)$ with $q_p\in\widetilde F$, and the suffix returns to this product state on each repetition. Let $\eta_i$ be the guard term selected for the transition from $(v_i,q_i)$ to $(v_{i+1},q_{i+1})$, including the self-loop guard for a dwell suffix. Product initialization consumes $L_\phi(s_0)$ before search, so $q_0\in\delta(q^{\mathrm{init}},L_\phi(s_0))$ for some $q^{\mathrm{init}}\in Q_0$.

\begin{lemma}[Soundness of the synthesized guard sequence]
\label{lem:product-soundness}
For the returned plan $\rho$, the set of infinite words $\widehat\sigma\in\Sigma^\omega$ satisfying
\[
\widehat\sigma_0=L_\phi(s_0),\qquad
\widehat\sigma_{i+1}\models\eta_i\quad\text{for all }i\ge0
\]
is nonempty, and every such word is accepted by $\mathcal B_\phi$.
\end{lemma}

\noindent\textit{Proof.}
Automaton preprocessing expresses each original guard as a disjunction of satisfiable conjunctive terms and retains the supported alternatives. The movement and dwell rules select among these terms, so each planned transition satisfies $\eta_i\Rightarrow\widetilde g_{q_i,q_{i+1}}\Rightarrow g_{q_i,q_{i+1}}$. Consequently, a word satisfying the selected guards gives $q_{i+1}\in\delta(q_i,\widehat\sigma_{i+1})$ at every transition, as in the standard product construction~\cite{baier2008principles,luo2021abstraction}. Together with the initialization transition on $\widehat\sigma_0$, these transitions form a run from $q^{\mathrm{init}}\in Q_0$. The run visits $q_p\in\widetilde F\subseteq F$ on every suffix repetition and is therefore accepting.

To construct a word satisfying the guards, choose one satisfying assignment for each term in the finite prefix and suffix, prepend $L_\phi(s_0)$, and repeat the suffix assignments indefinitely. Each retained term is satisfiable, so this construction establishes nonemptiness. \hfill$\square$

\paragraph{Soundness of confirmed execution segments.}
During physical execution, the monitor reads $\sigma_t=L_\phi(s_t)$ at each observation time $t$. These labels are obtained from the actual rollout $\tau_\Gamma$, where $\Gamma=\operatorname{proj}_{\mathcal V_\phi}(\rho)$. A segment between selected waypoints can span several physical steps. The monitor follows the original NBA over all labels in that segment and checks that they connect its planned automaton endpoints.

\begin{definition}[Confirmation of targets and suffix cycles]
\label{def:execution-realization}
Let $\kappa_0=0<\kappa_1<\cdots$ be the plan indices of successive confirmed targets, and let $t_r$ be the confirmation time of target $\kappa_r$, with $t_0=0$ for the initial product state. This sequence may be finite or infinite; gaps between indices correspond to skipped intermediate waypoints.
Starting from a confirmed target, the monitor initializes $S_r(t_r)=\{q_{\kappa_r}\}$ and updates it with each subsequent observation:
\[
S_r(t+1)=\bigcup_{q\in S_r(t)}\delta(q,\sigma_{t+1}).
\]
Target $\kappa_{r+1}$ is confirmed at time $t_{r+1}$ when the reach and guard checks in Appendix~\ref{app:execution-details} pass and $q_{\kappa_{r+1}}\in S_r(t_{r+1})$. The next segment begins with the singleton $\{q_{\kappa_{r+1}}\}$ and continues with observations after $t_{r+1}$.

After the prefix, a suffix cycle is confirmed when these checks establish a return to its accepting endpoint $(v_p,q_p)$. Each cycle includes at least one new physical observation, including when the waypoint remains unchanged in a dwell suffix.
\end{definition}

Write $\delta^*(U,w)$ for the set of automaton states reachable from $U\subseteq Q$ after reading a finite label word $w$. The following lemma identifies the monitor's state set with this reachability relation and connects successive confirmations to a run over the observed trajectory.

\begin{lemma}[Soundness of monitored progress]
\label{lem:monitor-soundness}
During the segment starting at target $\kappa_r$, the monitor in Algorithm~\ref{alg:execution} maintains
\[
S_r(t)=\delta^*\!\left(\{q_{\kappa_r}\},\sigma_{t_r+1}\cdots\sigma_t\right).
\]
Whenever target $\kappa_{r+1}$ is confirmed, the observed segment admits a run from $q_{\kappa_r}$ to $q_{\kappa_{r+1}}$. These segment runs concatenate with the initialization transition to form a run of the original NBA over all labels observed up to that confirmation.
\end{lemma}

\noindent\textit{Proof.}
At $t=t_r$, the segment word is empty and both sides equal $\{q_{\kappa_r}\}$. Each update applies $\delta$ to every currently reachable state using the next observed label. Induction on the observation time therefore establishes the equality throughout the segment. At confirmation, membership of $q_{\kappa_{r+1}}$ in this set gives a run between the two planned automaton states over precisely $\sigma_{t_r+1}\cdots\sigma_{t_{r+1}}$.

The initialization transition covers $\sigma_0$ and ends at $q_0$. Each confirmed segment extends the preceding run from its final automaton state, and the next segment begins with the following observation. Concatenating these runs therefore yields a run from an original initial state over the full label sequence observed so far. \hfill$\square$

\paragraph{LTL satisfaction through repeated suffix execution.}
The segment-wise result extends to infinite executions because every confirmed suffix cycle returns to the same accepting automaton state.

\begin{proposition}[LTL satisfaction from verified recurrence]
\label{prop:conditional-satisfaction}
Let $\tau_\Gamma$ be an execution of a lasso returned by SAGAS. If execution completes the prefix and the monitor confirms infinitely many suffix cycles according to Definition~\ref{def:execution-realization}, then $\tau_\Gamma\models\phi$.
\end{proposition}

\noindent\textit{Proof.}
Lemma~\ref{lem:monitor-soundness} provides compatible finite runs between consecutive confirmed targets. Concatenating them with the initialization transition constructs a run over the physical label sequence. Since every suffix cycle includes a new observation, infinitely many confirmed cycles extend this run over the entire infinite trajectory. The infinitely many returns to $q_p\in\widetilde F\subseteq F$ make the run accepting, and $\mathcal L(\mathcal B_\phi)=\mathcal L(\phi)$ gives $\tau_\Gamma\models\phi$. \hfill$\square$

Together, these results prove Proposition~\ref{prop:conditional-main}.

\section{Scope and limitations}
\label{app:limitations}
The logical guarantees in Appendix~\ref{app:conditional-proof} concern the planned guards and the labels observed during execution. Completing a task also requires the offline data to support the necessary motions and the learned executor to perform them reliably.

\textbf{Predicate evaluation and offline support.}
SAGAS assumes that a predicate evaluator is available at task time and uses it to label states preserved from the offline dataset. To ground a guard term containing positive literals, the planner needs a dataset state that satisfies the complete term and can be connected to the graph as an anchor. Sparse coverage can leave a term without such a witness; disabling that alternative may then prevent a plan from being found even when the task is physically feasible.

\textbf{Learned reachability and execution.}
Graph edges describe estimated reachability between latent waypoint regions. Executing an edge, however, requires the frozen policy to reach its target from the particular physical state produced by preceding actions. Errors in the learned reachability estimate or in waypoint tracking can therefore cause execution to fail even when the planner has found an accepting lasso. The failures analyzed in Appendix~\ref{app:failure-analysis} motivate more reliable local tracking and suffix plans whose transitions can be repeated consistently.

\textbf{Planning cost and search budget.}
SAGAS returns the lowest graph-cost lasso among the candidates retained under budget $K$. A finite budget can omit an accepting endpoint with a cheaper complete lasso, so increasing $K$ trades additional computation for the opportunity to find a better plan (Appendix~\ref{app:search-accuracy}). The objective uses learned temporal-distance costs as estimates of execution length. A guarantee of optimal physical execution would additionally require assumptions relating these costs to realized step counts.

\textbf{Negative constraints.}
When a guard requires $\neg\ell$, the soft-label screen admits an original graph node when the weighted fraction of its preserved states satisfying $\ell$ is at most $\tau_{\mathrm{soft}}$. With a positive threshold, some of those states may therefore violate the guard. Even when all preserved states satisfy $\neg\ell$, the motion between waypoints may encounter states where $\ell$ holds. The threshold controls how nodes are screened from the data; it does not bound the probability of a violation during execution. The monitor checks active negative obligations on the states actually reached, while navigation additionally uses guard-aware steering to bias motion away from forbidden regions. Assessing recorded transition segments during planning could help account for violations between waypoints.

\textbf{Execution failures and recovery.}
SAGAS terminates a rollout when a guard is violated or waypoint tracking times out. A future extension could replan after a timeout, using the current physical state and the automaton states consistent with the observed trajectory to seek another feasible continuation of the task.

\textbf{Recurrent specifications.}
We evaluate recurrent tasks by requiring the executor to complete the prefix and traverse the suffix $M$ times within the rollout budget (Appendix~\ref{app:evaluation-metrics}). This finite-lasso criterion tests whether the planned behavior can be sustained over the evaluation horizon. Satisfaction of the infinite-horizon specification follows when the monitor continues to confirm suffix cycles indefinitely, as established in Proposition~\ref{prop:conditional-satisfaction}.

\textbf{Environment scope.}
The evaluation uses state-based predicates and a fixed environment topology for each dataset. Generalization to unseen layouts or changing topology remains untested and would require adapting the reachability interface to the new connectivity. Visual or language-based predicate grounding and planning under partial observability are further directions for extending the framework.

\textbf{Deployment scope.}
Moving from the simulated benchmarks to physical robots would require validating the learned executor under real operating conditions and integrating independent safety monitors and fail-safe controllers.

\section{Experimental protocols and implementation details}
\label{app:exp-details}

The navigation benchmark compares end-to-end LTL planning across robot dynamics and offline dataset regimes. The overlapping-region and Scene suites examine predicate generalization through simultaneous region membership and joint object configurations. Appendix~\ref{app:exp-results} reports the results for all three suites.

\subsection{Navigation environments and datasets}
\label{app:navigation-environments}
\begin{figure}[htbp]
    \centering
    \begin{subfigure}[t]{0.28\linewidth}
        \centering
        \includegraphics[height=1.2in]{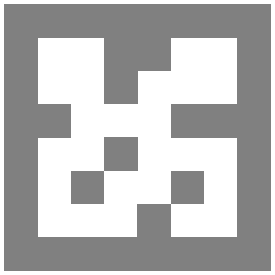}
        \caption{\emph{Medium}}
    \end{subfigure}
    \hfill
    \begin{subfigure}[t]{0.32\linewidth}
        \centering
        \includegraphics[height=1.2in]{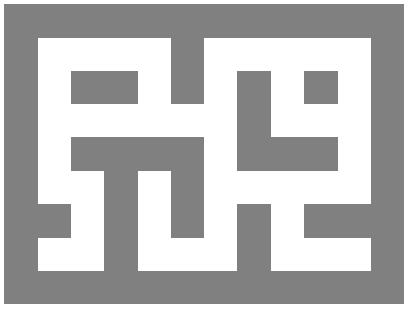}
        \caption{\emph{Large}}
    \end{subfigure}
    \hfill
    \begin{subfigure}[t]{0.38\linewidth}
        \centering
        \includegraphics[height=1.2in]{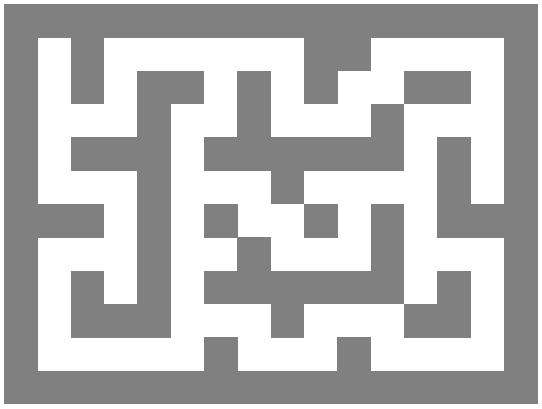}
        \caption{\emph{Giant}}
    \end{subfigure}
    \caption{Maze layouts used in the OGBench locomotion domains. Gray regions indicate walls and white regions indicate navigable space.}
    \label{fig:maze-layouts}
\end{figure}

We use the \texttt{antmaze} and \texttt{humanoidmaze} locomotion domains from OGBench~\cite{park2025ogbench}.
AntMaze has an 8-DoF agent with 29-dimensional state observations. Its eight benchmark settings combine \emph{medium}, \emph{large}, and \emph{giant} maze layouts with the available \texttt{navigate}, \texttt{stitch}, and \texttt{explore} datasets.
HumanoidMaze has a 21-DoF agent with 69-dimensional state observations and uses the same family of layouts with the available \texttt{navigate} and \texttt{stitch} datasets.
Figure~\ref{fig:maze-layouts} shows the layouts. The methods learn from offline trajectories and receive no obstacle maps.

All datasets contain task-agnostic trajectories collected independently of the test formulas. The AntMaze data regimes differ in the behavior they provide: \texttt{navigate} contains noisy-expert maze traversals, \texttt{stitch} contains short fragments that require composition at task time, and \texttt{explore} contains random trajectories with broad coverage but low-quality behavior.

\subsection{Navigation task generation}
\label{app:navigation-task-generation}
We generate LTL$_{-\mathrm{X}}$ tasks by randomly instantiating and composing the templates in Table~\ref{tab:LTL_templates}.
The difficulty groups in Table~\ref{tab:ltl_difficulty_levels} control the number of propositions and the number of template instances combined by conjunction.
Each proposition corresponds to a geometric region centered in a sampled free maze cell, with regions constrained to be pairwise disjoint.

Each environment--difficulty pair contains 100 tasks shared by all methods. The generator checks that each formula is syntactically well formed and that its automaton has a nonempty language. Every task that passes these checks is retained and evaluated by all methods.

\begin{table}[H]
\centering
\caption{Parameterized LTL task templates used by the generator.}
\label{tab:LTL_templates}
\small
\begin{tabular}{p{0.47\linewidth}p{0.45\linewidth}}
\toprule
\textbf{LTL formula structure} & \textbf{Semantic description} \\
\midrule
$\F \ell_i$ & \textbf{Reach:} Eventually visit region $i$. \\
$\G (\neg \ell_i)$ & \textbf{Safety:} Always avoid region $i$. \\
$\F (\ell_1 \wedge \F (\ell_2 \wedge \dots \wedge \F \ell_m))$ & \textbf{Sequence:} Visit $1 \to 2 \to \dots \to m$. \\
$\F \ell_1 \wedge \F \ell_2 \wedge \dots \wedge \F \ell_m$ & \textbf{Coverage:} Visit $m$ regions in any order. \\
$\neg \ell_{\text{avoid}} \,\U\, \ell_{\text{goal}}$ & \textbf{Conditional:} Avoid until goal is reached. \\
$\G \F (\ell_1 \wedge \F (\ell_2 \wedge \dots \wedge \F \ell_m))$ & \textbf{Patrol:} Infinitely revisit $1 \to \dots \to m$. \\
$\F (\ell_1 \vee \ell_2 \vee \dots \vee \ell_m)$ & \textbf{Choice:} Visit at least one region in set. \\
$\F\G \ell_i$ & \textbf{Persistence:} Eventually stay in region $i$. \\
$\F(\ell_1 \wedge \F(\ell_2 \wedge \F\G \ell_3))$ & \textbf{Sequence-to-persist:} Visit a sequence and eventually remain in the final region. \\
$\bigwedge_i \F \ell_i \wedge \bigwedge_{i>1}(\neg \ell_i \,\U\, \ell_{i-1})$ & \textbf{Strict order:} Visit multiple regions while enforcing a prescribed order. \\
$\bigwedge_i \F \ell_i \wedge \bigwedge_{j>1}(\neg \ell_1 \,\U\, \ell_j)$ & \textbf{Last visit:} Visit all regions while delaying region $1$ until the others are reached. \\
\bottomrule
\end{tabular}
\end{table}

\begin{table}[H]
\centering
\caption{Difficulty groups for generated LTL test cases.}
\label{tab:ltl_difficulty_levels}
\small
\begin{tabular}{lll}
\toprule
\textbf{Difficulty} & \textbf{Logical composition} & \textbf{Labeled regions} \\
\midrule
Easy & One template instance, no conjunction of multiple templates & At most 5 \\
Medium & Conjunction of up to 3 template instances & At most 5 \\
Hard & Conjunction of up to 4 template instances & At most 8 \\
\bottomrule
\end{tabular}
\end{table}

\subsection{Compared methods}
\label{app:compared-methods}
SAGAS, \LFGAS{}, and \GAGAS{} use the same latent graph and frozen policy within each dataset. They differ in how they select the high-level plan, allowing a comparison of planning strategies under a common reachability interface.

\LFGAS{} (Logic-first + GAS) first selects an accepting prefix--suffix sequence in the B\"uchi automaton, favoring sequences with fewer transitions. It maps the positive requirements to anchors and connects them by shortest paths in the latent graph, using the same node-level forbidden-label pruning as SAGAS. Graph costs thus guide motion between anchors, while the automaton sequence is chosen independently of those costs.

\GAGAS{} (Greedy anchor + GAS) selects the locally feasible outgoing B\"uchi transition with the lowest current GAS shortest-path cost to its required proposition, using the same forbidden-label pruning. Transitions without positive literals are taken in place when their negative requirements pass the check. The greedy rule constructs the prefix and a suffix returning to the same accepting graph node and B\"uchi state.

\APHIQL{} (Automaton path + HIQL goal reaching) adapts LTL-instructed goal reaching~\cite{qiu2023instructing} using a separate frozen HIQL~\cite{park2023hiql} policy. It first selects an accepting prefix--suffix label sequence in the automaton and chooses dataset states from the required regions as goals. The policy then reaches these goals in order while monitoring forbidden labels. This provides an end-to-end comparison with a different goal-conditioned architecture.

\subsection{Evaluation protocol and metrics}
\label{app:evaluation-metrics}
All reusable components are learned from the offline dataset and frozen before evaluation, including the HIQL policy used by \APHIQL{}. For each test case, every method receives the same formula, task-space labeling function, and initial state.

The navigation benchmark and product-size experiments initialize product search at an initial state of the NBA. The overlap and Scene suites first advance the NBA with $L_\phi(s_0)$, following Section~\ref{sec:planning}. In all three suites, final verification reads the physical label sequence from $L_\phi(s_0)$ onward.

Labels are evaluated on the executed states. Success rate (SR) is the fraction of tasks meeting the finite-lasso criterion: prefix completion and $M$ verified suffix traversals, with $M=2$ unless otherwise specified. A rollout that reaches its environment-step limit before meeting this criterion counts as a failure.

Normalized capped cost (NCC) evaluates the prefix and first valid suffix traversal under a common budget.
For a rollout that completes both, let \(T_{\mathrm{pre}}\) and \(T_{\mathrm{suf}}\) denote the respective environment-step counts.
Its NCC is
\[
\mathrm{NCC}
=
\frac{
\min\{\lambda T_{\mathrm{pre}}+(1-\lambda)T_{\mathrm{suf}},\, C_{\max}\}
}{C_{\max}},
\]
where \(C_{\max}\) is the capped evaluation budget shared by all methods within a setting.
Navigation uses $\lambda=0.5$ and the setting-specific budget $C_{\max}$, giving the equivalent expression $\min\{T_{\mathrm{pre}}+T_{\mathrm{suf}},2C_{\max}\}/(2C_{\max})$. We assign NCC $1$ when the prefix or first valid suffix traversal is not completed. Lower NCC therefore reflects shorter executions and fewer capped failures.

In navigation, SR requires $M=2$ suffix traversals, whereas NCC uses the one-cycle criterion. A rollout that completes the first suffix cycle but fails during the second counts as an SR failure, although its NCC can remain below $1$.

\subsection{Parameters and computing resources}
\label{app:implementation-details}

Table~\ref{tab:implementation-task-time} lists the navigation parameters shared by SAGAS, \LFGAS{}, and \GAGAS{}. Backbone training follows the GAS implementation for the corresponding OGBench domains~\cite{baek2025graph}; the methods reuse the same trained graph and policy within each dataset.

\begin{table}[H]
\centering
\caption{Task-time semantic graph augmentation, planning, and execution parameters for the navigation benchmark.}
\label{tab:implementation-task-time}
\footnotesize
\setlength{\tabcolsep}{3.5pt}
\begin{tabular}{p{0.35\linewidth}p{0.25\linewidth}p{0.25\linewidth}}
\toprule
\textbf{Parameter} & \textbf{AntMaze} & \textbf{HumanoidMaze} \\
\midrule
Task cases per environment--difficulty pair & 100 & 100 \\
\midrule
Anchor source & Dataset retrieval & Dataset retrieval \\
Initial anchors per proposition $N_s$ & 3 & 3 \\
Maximum candidate attempts per proposition & 50 & 50 \\
Soft-label threshold $\tau_{\mathrm{soft}}$ & 0.05 & 0.20 \\
\midrule
Temporal-distance scale $H_{\mathrm{TD}}$ & 8 & 32 \\
Prefix candidates $K$ & 5 & 5 \\
Prefix--suffix trade-off $\lambda$ & 0.5 & 0.5 \\
Suffix traversals for finite evaluation $M$ & 2 & 2 \\
Maximum rollout steps & 8000 & 12000 / 16000 / 20000 \\
\midrule
Guard-aware steering radius $\rho_{\mathrm{rep}}$ & 64 & 64 \\
Guard-aware steering gain $\beta$ & 0.1 & 0.01 \\
Guard-aware steering power $p$ & 1.0 & 1.0 \\
Guard-aware steering max norm $r_{\max}$ & 0.2 & 0.02 \\
\bottomrule
\end{tabular}
\end{table}

The three HumanoidMaze rollout budgets correspond to \texttt{medium}, \texttt{large}, and \texttt{giant} layouts, respectively. Navigation experiments ran on an Ubuntu 24.04 server with dual Intel Xeon Gold 6426Y CPUs, 256 GB RAM, and one NVIDIA RTX 3090 GPU with 24 GB memory. The following protocols specify the settings used for the overlap and Scene suites.

\subsection{Overlapping-region tasks and evaluation}\label{app:overlap-protocol}
All four task families use the intersection of regions $a$ and $b$ as a
target. The agent must reach the overlap ($\F(a\wedge b)$), avoid $c$ until
it reaches the overlap ($\neg c\,\U\,(a\wedge b)$), revisit it repeatedly
($\G\F(a\wedge b)$), or reach it and then visit $d$
($\F((a\wedge b)\wedge\F d)$). In AntMaze Giant-Stitch and Large-Explore,
we sample 40 configurations per family in each environment, giving 320 tasks.
Region shapes include circles, convex polygons, and non-convex polygons. Before planning, we
check that the target intersection is nonempty and the initial position is
clear, then fix the regions and initial state for each task.

We use the AntMaze settings in Table~\ref{tab:implementation-task-time} for
the number of prefix candidates $K$, guard-aware steering, and maximum
rollout steps. For overlapping regions, we use a soft-label threshold of
$\tau_{\mathrm{soft}}=0.01$. We run each task once and report finite-lasso
SR as defined in Appendix~\ref{app:evaluation-metrics}.

\Needspace{17\baselineskip}
\subsection{Scene manipulation tasks and evaluation}\label{app:scene-protocol}
\begin{wrapfigure}[13]{r}{0.25\linewidth}
\centering
\includegraphics[width=\linewidth]{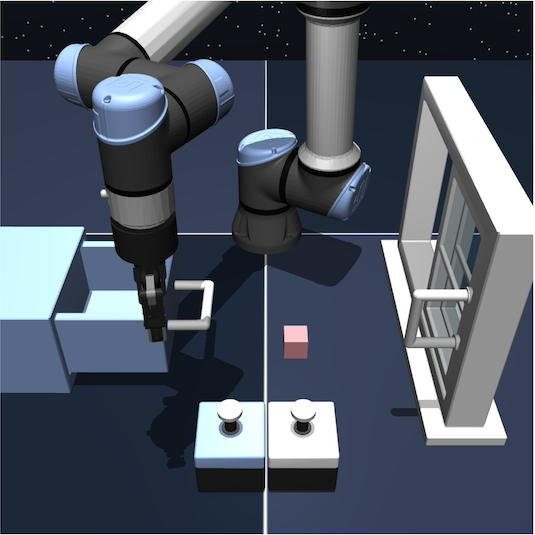}
\captionsetup{justification=raggedright,singlelinecheck=false}
\caption{Scene workspace, cropped from the
\href{https://github.com/seohongpark/ogbench/blob/master/assets/env_teaser.png}{OGBench overview}~\cite{park2025ogbench}.}
\label{fig:ogbench-scene}
\end{wrapfigure}
In OGBench \texttt{scene-play-v0}, a robot arm manipulates a cube, a drawer,
a window, and two buttons (Figure~\ref{fig:ogbench-scene}). Each button
toggles the lock on either the drawer or the window, so moving one of these
objects can require first pressing its button. The offline dataset consists
of scripted interactions with these objects, collected independently of the
LTL tasks. We use the released GAS Scene representation and low-level policy
and build a reachability graph that retains the dataset states assigned to
each node.

\paragraph{Predicates over joint object configurations.}
We define the task predicates on a 7D mixed discrete--continuous space:
the cube's 3D position, the drawer and window positions, and the two binary
button states.

The predicates $b_0$ and $b_1$ indicate whether the corresponding button
is on. We label the drawer open ($d_o$) at positions at or below $-0.10$
and closed ($d_c$) at or above $-0.08$. For the window, the open predicate
($w_o$) holds at or above $0.12$ and the closed predicate ($w_c$) at or
below $0.08$. Between its two thresholds, the drawer or window is labeled
neither open nor closed.

We choose three fixed cube positions $A$, $B$, and $C$ from the offline
data. The predicates $c_A$, $c_B$, and $c_C$ each require the cube to be
within $0.06$ of the corresponding position; $c_{A+}$ uses a wider radius
of $0.10$ around $A$. For the cube-in-drawer predicate $c_D$, the cube must lie
inside a box that moves with the drawer handle:
$x\in[0.21,0.45]$, $y-y_{\mathrm{handle}}\in[-0.27,-0.07]$, and
$z\in[0,0.15]$.

Table~\ref{tab:scene-configurations} defines the ten joint configurations.
Each row specifies the object predicates that must hold simultaneously.
\begin{table}[H]
\centering\small
\caption{Joint object configurations used in the Scene tasks.}
\label{tab:scene-configurations}
\begin{tabular}{lll}
\toprule
Symbol & Main feature & Full joint condition\\
\midrule
$D_r$ & Drawer ready & $b_0\wedge d_c\wedge w_c\wedge c_{A+}$\\
$D_o$ & Drawer open & $b_0\wedge d_o\wedge w_c\wedge c_{A+}$\\
$W_r$ & Window ready & $b_1\wedge d_c\wedge w_c\wedge c_{A+}$\\
$W_o$ & Window open & $b_1\wedge d_c\wedge w_o\wedge c_{A+}$\\
$C_A$ & Cube at A, objects closed & $c_A\wedge d_c\wedge w_c$\\
$C_B$ & Cube at B, objects open & $c_B\wedge d_o\wedge w_o$\\
$C_{CD}$ & Cube at C, drawer open & $c_C\wedge b_1\wedge d_o\wedge w_c$\\
$C_{CW}$ & Cube at C, window open & $c_C\wedge b_1\wedge d_c\wedge w_o$\\
$C_D$ & Cube stored, window closed & $c_D\wedge b_0\wedge d_c\wedge w_c$\\
$C_{DW}$ & Cube stored, window open & $c_D\wedge b_0\wedge d_c\wedge w_o$\\
\bottomrule
\end{tabular}
\end{table}
\FloatBarrier

\paragraph{Fixed task design.}
Each of the six task families contains four hand-written LTL formulas that
combine different predicates. Table~\ref{tab:scene-formulas} gives one
example per family. In the two-stage example, $\F(D_r\wedge\F D_o)$, the
drawer must first be closed and later open. At both stages, button~0 must
be on, the window closed, and the cube near $A$.

We pair each of the 24 formulas with all five official OGBench Scene
initial configurations, giving 20 tasks per family and 120 in total.
The initial configurations supply the starting states; the LTL formulas
specify the goals.
\begin{table}[H]
\centering\small
\caption{Representative Scene formulas, one from each of the six task families.}
\label{tab:scene-formulas}
\begin{tabular}{ll}
\toprule
Task family & Representative formula\\
\midrule
Single-stage configuration & $\F D_o$\\
Ordered two-stage manipulation & $\F(D_r\wedge\F D_o)$\\
Ordered three-stage manipulation & $\F(C_A\wedge\F(C_{CD}\wedge\F C_{CW}))$\\
Choice then configuration & $\F((D_r\vee W_r)\wedge\F D_o)$\\
Negative-until & $\neg c_D\,\U\,C_{CD}$\\
Recurrent suffix & $\G\F(D_o\wedge\F W_o)$\\
\bottomrule
\end{tabular}
\end{table}
\FloatBarrier

\paragraph{Execution and verification.}
When selecting anchors, we give priority to states immediately following an
observed change in the relevant object and retain examples with different
configurations of the other objects. We use a soft-label threshold of
$\tau_{\mathrm{soft}}=0.01$ for negative literals. Task-time planning is
timed from predicate-evaluator initialization to the end of product search,
including semantic graph augmentation (Table~\ref{tab:scene-timing}).

We run each task once, with at most 8,000 rollout steps and no guard-aware
steering. An independent verifier evaluates the same predicates on the
recorded state trajectory to check finite-lasso success. For recurrent
tasks, this requires completing the prefix and one suffix traversal
($M=1$). We report SR over all 120 tasks.

\FloatBarrier

\section{Detailed task results}
\label{app:exp-results}

\subsection{Navigation benchmark results}
\label{app:navigation-results}
Tables~\ref{tab:antmaze-ltl-main} and~\ref{tab:humanoid-ltl-main} report the AntMaze and HumanoidMaze results, with 100 generated LTL tasks per environment--difficulty pair. SR and NCC follow the definitions in Appendix~\ref{app:evaluation-metrics}.

SAGAS leads the aggregate AntMaze results with $71.8\%$ SR and $0.30$ NCC, compared with $67.0\%$ and $0.35$ for \LFGAS{}. Its advantage is most pronounced in \texttt{giant-stitch} and \texttt{large-explore}, where logical choices interact with reachability and coverage. HumanoidMaze has the same aggregate ordering: SAGAS reaches $57.4\%$ SR and $0.45$ NCC, followed by \LFGAS{} at $54.3\%$ and $0.49$. Some individual HumanoidMaze settings favor a baseline, while SAGAS leads overall and on the Hard-task summary in Table~\ref{tab:ltl-domain-summary}.

\raggedbottom
\begin{table}[H]
\centering
\small
\caption{AntMaze LTL benchmark results. Each environment--difficulty setting contains 100 test cases. SR is finite-lasso execution success rate; NCC is normalized capped cost for prefix completion plus one suffix traversal. Dark/light gray shading marks the best/second-best method within each setting; higher SR and lower NCC are better.}
\label{tab:antmaze-ltl-main}
\resizebox{\linewidth}{!}{%
\begin{tabular}{llrrrrrrrr}
\toprule
Environment & Difficulty & \multicolumn{2}{c}{SAGAS} & \multicolumn{2}{c}{\LFGAS{}} & \multicolumn{2}{c}{\GAGAS{}} & \multicolumn{2}{c}{\APHIQL{}} \\
\cmidrule(lr){3-4} \cmidrule(lr){5-6} \cmidrule(lr){7-8} \cmidrule(lr){9-10}
 &  & SR (\%) & NCC & SR (\%) & NCC & SR (\%) & NCC & SR (\%) & NCC \\
\midrule
Medium-Navigate & Easy & \cellcolor{gray!12}78.0 & \cellcolor{gray!30}$0.21\pm0.37$ & \cellcolor{gray!30}79.0 & \cellcolor{gray!12}$0.21\pm0.37$ & 68.0 & $0.35\pm0.45$ & 70.0 & $0.27\pm0.41$ \\
 & Medium & \cellcolor{gray!30}83.0 & \cellcolor{gray!30}$0.17\pm0.34$ & \cellcolor{gray!12}74.0 & $0.27\pm0.41$ & 60.0 & $0.42\pm0.48$ & 70.0 & \cellcolor{gray!12}$0.26\pm0.40$ \\
 & Hard & \cellcolor{gray!30}65.0 & \cellcolor{gray!30}$0.38\pm0.45$ & 60.0 & \cellcolor{gray!12}$0.42\pm0.46$ & 54.0 & $0.48\pm0.47$ & \cellcolor{gray!12}61.0 & $0.42\pm0.45$ \\
 & \emph{All} & \cellcolor{gray!30}75.3 & \cellcolor{gray!30}$0.25\pm0.40$ & \cellcolor{gray!12}71.0 & \cellcolor{gray!12}$0.30\pm0.42$ & 60.7 & $0.42\pm0.47$ & 67.0 & $0.32\pm0.43$ \\
\addlinespace
Large-Navigate & Easy & \cellcolor{gray!12}73.0 & \cellcolor{gray!12}$0.27\pm0.39$ & 67.0 & $0.31\pm0.42$ & 59.0 & $0.41\pm0.46$ & \cellcolor{gray!30}77.0 & \cellcolor{gray!30}$0.26\pm0.39$ \\
 & Medium & 63.0 & $0.37\pm0.44$ & 67.0 & $0.38\pm0.44$ & \cellcolor{gray!12}68.0 & \cellcolor{gray!12}$0.37\pm0.44$ & \cellcolor{gray!30}71.0 & \cellcolor{gray!30}$0.32\pm0.42$ \\
 & Hard & \cellcolor{gray!30}55.0 & \cellcolor{gray!30}$0.50\pm0.45$ & 49.0 & $0.54\pm0.45$ & 45.0 & $0.58\pm0.45$ & \cellcolor{gray!12}51.0 & \cellcolor{gray!12}$0.51\pm0.45$ \\
 & \emph{All} & \cellcolor{gray!12}63.7 & \cellcolor{gray!12}$0.38\pm0.44$ & 61.0 & $0.41\pm0.44$ & 57.3 & $0.45\pm0.46$ & \cellcolor{gray!30}66.3 & \cellcolor{gray!30}$0.36\pm0.44$ \\
\addlinespace
Giant-Navigate & Easy & \cellcolor{gray!12}61.0 & $0.41\pm0.45$ & \cellcolor{gray!30}62.0 & \cellcolor{gray!30}$0.39\pm0.43$ & 55.0 & $0.47\pm0.46$ & \cellcolor{gray!30}62.0 & \cellcolor{gray!12}$0.41\pm0.43$ \\
 & Medium & \cellcolor{gray!30}57.0 & \cellcolor{gray!12}$0.43\pm0.43$ & \cellcolor{gray!12}56.0 & \cellcolor{gray!30}$0.42\pm0.43$ & 53.0 & $0.50\pm0.45$ & 47.0 & $0.53\pm0.44$ \\
 & Hard & \cellcolor{gray!30}48.0 & \cellcolor{gray!30}$0.53\pm0.44$ & 37.0 & $0.65\pm0.43$ & \cellcolor{gray!12}45.0 & \cellcolor{gray!12}$0.60\pm0.44$ & 37.0 & $0.68\pm0.42$ \\
 & \emph{All} & \cellcolor{gray!30}55.3 & \cellcolor{gray!30}$0.45\pm0.44$ & \cellcolor{gray!12}51.7 & \cellcolor{gray!12}$0.49\pm0.44$ & 51.0 & $0.52\pm0.45$ & 48.7 & $0.54\pm0.44$ \\
\addlinespace
Medium-Stitch & Easy & \cellcolor{gray!12}81.0 & \cellcolor{gray!12}$0.21\pm0.37$ & 80.0 & $0.23\pm0.39$ & 75.0 & $0.28\pm0.42$ & \cellcolor{gray!30}82.0 & \cellcolor{gray!30}$0.20\pm0.34$ \\
 & Medium & \cellcolor{gray!30}85.0 & \cellcolor{gray!30}$0.17\pm0.34$ & \cellcolor{gray!12}81.0 & \cellcolor{gray!12}$0.19\pm0.35$ & 77.0 & $0.25\pm0.40$ & 69.0 & $0.32\pm0.43$ \\
 & Hard & \cellcolor{gray!30}71.0 & \cellcolor{gray!30}$0.30\pm0.41$ & \cellcolor{gray!12}65.0 & \cellcolor{gray!12}$0.35\pm0.44$ & 56.0 & $0.46\pm0.47$ & 61.0 & $0.40\pm0.44$ \\
 & \emph{All} & \cellcolor{gray!30}79.0 & \cellcolor{gray!30}$0.23\pm0.38$ & \cellcolor{gray!12}75.3 & \cellcolor{gray!12}$0.26\pm0.40$ & 69.3 & $0.33\pm0.44$ & 70.7 & $0.31\pm0.41$ \\
\addlinespace
Large-Stitch & Easy & \cellcolor{gray!30}74.0 & \cellcolor{gray!30}$0.27\pm0.39$ & \cellcolor{gray!12}73.0 & \cellcolor{gray!12}$0.28\pm0.40$ & 65.0 & $0.38\pm0.45$ & 67.0 & $0.34\pm0.43$ \\
 & Medium & \cellcolor{gray!30}81.0 & \cellcolor{gray!30}$0.21\pm0.35$ & \cellcolor{gray!12}79.0 & \cellcolor{gray!12}$0.23\pm0.35$ & 73.0 & $0.29\pm0.40$ & 61.0 & $0.40\pm0.45$ \\
 & Hard & \cellcolor{gray!30}62.0 & \cellcolor{gray!30}$0.41\pm0.43$ & 49.0 & \cellcolor{gray!12}$0.50\pm0.45$ & \cellcolor{gray!12}52.0 & $0.50\pm0.45$ & 45.0 & $0.57\pm0.45$ \\
 & \emph{All} & \cellcolor{gray!30}72.3 & \cellcolor{gray!30}$0.30\pm0.40$ & \cellcolor{gray!12}67.0 & \cellcolor{gray!12}$0.33\pm0.42$ & 63.3 & $0.39\pm0.44$ & 57.7 & $0.43\pm0.45$ \\
\addlinespace
Giant-Stitch & Easy & \cellcolor{gray!30}61.0 & \cellcolor{gray!30}$0.43\pm0.43$ & 56.0 & $0.49\pm0.43$ & \cellcolor{gray!12}60.0 & \cellcolor{gray!12}$0.46\pm0.44$ & 20.0 & $0.82\pm0.36$ \\
 & Medium & \cellcolor{gray!30}56.0 & \cellcolor{gray!30}$0.46\pm0.43$ & \cellcolor{gray!12}50.0 & \cellcolor{gray!12}$0.53\pm0.43$ & 45.0 & $0.58\pm0.44$ & 10.0 & $0.91\pm0.27$ \\
 & Hard & \cellcolor{gray!30}54.0 & \cellcolor{gray!30}$0.50\pm0.42$ & \cellcolor{gray!12}41.0 & \cellcolor{gray!12}$0.64\pm0.42$ & 35.0 & $0.68\pm0.41$ & 8.0 & $0.93\pm0.23$ \\
 & \emph{All} & \cellcolor{gray!30}57.0 & \cellcolor{gray!30}$0.46\pm0.43$ & \cellcolor{gray!12}49.0 & \cellcolor{gray!12}$0.55\pm0.43$ & 46.7 & $0.57\pm0.44$ & 12.7 & $0.89\pm0.29$ \\
\addlinespace
Medium-Explore & Easy & \cellcolor{gray!30}92.0 & \cellcolor{gray!12}$0.11\pm0.27$ & \cellcolor{gray!30}92.0 & \cellcolor{gray!30}$0.11\pm0.25$ & \cellcolor{gray!12}81.0 & $0.21\pm0.38$ & 53.0 & $0.51\pm0.46$ \\
 & Medium & \cellcolor{gray!30}90.0 & \cellcolor{gray!30}$0.13\pm0.29$ & \cellcolor{gray!12}82.0 & \cellcolor{gray!12}$0.19\pm0.36$ & 73.0 & $0.28\pm0.43$ & 63.0 & $0.42\pm0.44$ \\
 & Hard & \cellcolor{gray!30}85.0 & \cellcolor{gray!30}$0.18\pm0.35$ & \cellcolor{gray!12}80.0 & \cellcolor{gray!12}$0.24\pm0.38$ & 74.0 & $0.28\pm0.42$ & 44.0 & $0.60\pm0.45$ \\
 & \emph{All} & \cellcolor{gray!30}89.0 & \cellcolor{gray!30}$0.14\pm0.30$ & \cellcolor{gray!12}84.7 & \cellcolor{gray!12}$0.18\pm0.34$ & 76.0 & $0.26\pm0.41$ & 53.3 & $0.51\pm0.46$ \\
\addlinespace
Large-Explore & Easy & \cellcolor{gray!30}83.0 & \cellcolor{gray!30}$0.20\pm0.35$ & \cellcolor{gray!12}79.0 & \cellcolor{gray!12}$0.22\pm0.36$ & 74.0 & $0.29\pm0.42$ & 17.0 & $0.83\pm0.37$ \\
 & Medium & \cellcolor{gray!30}86.0 & \cellcolor{gray!30}$0.16\pm0.31$ & \cellcolor{gray!12}83.0 & \cellcolor{gray!12}$0.20\pm0.34$ & 80.0 & $0.22\pm0.37$ & 15.0 & $0.84\pm0.35$ \\
 & Hard & \cellcolor{gray!30}80.0 & \cellcolor{gray!30}$0.26\pm0.36$ & 66.0 & $0.38\pm0.43$ & \cellcolor{gray!12}68.0 & \cellcolor{gray!12}$0.36\pm0.43$ & 11.0 & $0.90\pm0.28$ \\
 & \emph{All} & \cellcolor{gray!30}83.0 & \cellcolor{gray!30}$0.21\pm0.35$ & \cellcolor{gray!12}76.0 & \cellcolor{gray!12}$0.26\pm0.39$ & 74.0 & $0.29\pm0.41$ & 14.3 & $0.86\pm0.34$ \\
\midrule
\emph{Overall} & \emph{All} & \cellcolor{gray!30}71.8 & \cellcolor{gray!30}$0.30\pm0.41$ & \cellcolor{gray!12}67.0 & \cellcolor{gray!12}$0.35\pm0.43$ & 62.3 & $0.40\pm0.45$ & 48.8 & $0.53\pm0.46$ \\
\bottomrule
\end{tabular}
}
\end{table}

\begin{table}[H]
\centering
\small
\caption{HumanoidMaze LTL benchmark results. Each environment--difficulty setting contains 100 test cases. SR is finite-lasso execution success rate; NCC is normalized capped cost for prefix completion plus one suffix traversal. Dark/light gray shading marks the best/second-best method within each setting; higher SR and lower NCC are better.}
\label{tab:humanoid-ltl-main}
\resizebox{\linewidth}{!}{%
\begin{tabular}{llrrrrrrrr}
\toprule
Environment & Difficulty & \multicolumn{2}{c}{SAGAS} & \multicolumn{2}{c}{\LFGAS{}} & \multicolumn{2}{c}{\GAGAS{}} & \multicolumn{2}{c}{\APHIQL{}} \\
\cmidrule(lr){3-4} \cmidrule(lr){5-6} \cmidrule(lr){7-8} \cmidrule(lr){9-10}
 &  & SR (\%) & NCC & SR (\%) & NCC & SR (\%) & NCC & SR (\%) & NCC \\
\midrule
Medium-Navigate & Easy & \cellcolor{gray!30}81.0 & \cellcolor{gray!12}$0.24\pm0.36$ & \cellcolor{gray!12}78.0 & \cellcolor{gray!30}$0.24\pm0.35$ & 76.0 & $0.29\pm0.39$ & 75.0 & $0.29\pm0.38$ \\
 & Medium & \cellcolor{gray!12}66.0 & \cellcolor{gray!12}$0.36\pm0.41$ & \cellcolor{gray!12}66.0 & $0.38\pm0.41$ & 55.0 & $0.48\pm0.45$ & \cellcolor{gray!30}77.0 & \cellcolor{gray!30}$0.30\pm0.37$ \\
 & Hard & \cellcolor{gray!30}67.0 & \cellcolor{gray!30}$0.36\pm0.40$ & 57.0 & \cellcolor{gray!12}$0.45\pm0.42$ & 55.0 & $0.52\pm0.44$ & \cellcolor{gray!12}61.0 & $0.48\pm0.42$ \\
 & \emph{All} & \cellcolor{gray!30}71.3 & \cellcolor{gray!30}$0.32\pm0.39$ & 67.0 & \cellcolor{gray!12}$0.35\pm0.40$ & 62.0 & $0.43\pm0.44$ & \cellcolor{gray!12}71.0 & $0.36\pm0.40$ \\
\addlinespace
Medium-Stitch & Easy & 68.0 & $0.31\pm0.41$ & \cellcolor{gray!12}72.0 & \cellcolor{gray!30}$0.30\pm0.39$ & 69.0 & $0.33\pm0.42$ & \cellcolor{gray!30}74.0 & \cellcolor{gray!12}$0.30\pm0.39$ \\
 & Medium & \cellcolor{gray!12}70.0 & \cellcolor{gray!12}$0.32\pm0.39$ & \cellcolor{gray!30}74.0 & \cellcolor{gray!30}$0.31\pm0.38$ & 61.0 & $0.42\pm0.44$ & 67.0 & $0.37\pm0.40$ \\
 & Hard & \cellcolor{gray!30}64.0 & \cellcolor{gray!30}$0.39\pm0.41$ & 54.0 & \cellcolor{gray!12}$0.50\pm0.43$ & \cellcolor{gray!12}56.0 & $0.50\pm0.44$ & 51.0 & $0.53\pm0.42$ \\
 & \emph{All} & \cellcolor{gray!30}67.3 & \cellcolor{gray!30}$0.34\pm0.40$ & \cellcolor{gray!12}66.7 & \cellcolor{gray!12}$0.37\pm0.41$ & 62.0 & $0.42\pm0.44$ & 64.0 & $0.40\pm0.41$ \\
\addlinespace
Large-Navigate & Easy & \cellcolor{gray!12}69.0 & $0.35\pm0.42$ & \cellcolor{gray!30}73.0 & \cellcolor{gray!30}$0.32\pm0.39$ & 66.0 & $0.39\pm0.43$ & \cellcolor{gray!12}69.0 & \cellcolor{gray!12}$0.35\pm0.40$ \\
 & Medium & \cellcolor{gray!30}66.0 & \cellcolor{gray!30}$0.37\pm0.41$ & 61.0 & $0.43\pm0.43$ & 55.0 & $0.48\pm0.45$ & \cellcolor{gray!12}62.0 & \cellcolor{gray!12}$0.42\pm0.42$ \\
 & Hard & \cellcolor{gray!30}57.0 & \cellcolor{gray!30}$0.51\pm0.43$ & 44.0 & $0.59\pm0.43$ & 45.0 & $0.58\pm0.43$ & \cellcolor{gray!12}48.0 & \cellcolor{gray!12}$0.56\pm0.43$ \\
 & \emph{All} & \cellcolor{gray!30}64.0 & \cellcolor{gray!30}$0.41\pm0.42$ & 59.3 & $0.44\pm0.43$ & 55.3 & $0.48\pm0.44$ & \cellcolor{gray!12}59.7 & \cellcolor{gray!12}$0.44\pm0.43$ \\
\addlinespace
Large-Stitch & Easy & \cellcolor{gray!30}75.0 & \cellcolor{gray!30}$0.30\pm0.39$ & 66.0 & $0.39\pm0.43$ & 61.0 & $0.43\pm0.44$ & \cellcolor{gray!12}74.0 & \cellcolor{gray!12}$0.37\pm0.38$ \\
 & Medium & \cellcolor{gray!12}64.0 & \cellcolor{gray!30}$0.42\pm0.43$ & 60.0 & \cellcolor{gray!12}$0.42\pm0.43$ & 55.0 & $0.50\pm0.45$ & \cellcolor{gray!30}67.0 & $0.44\pm0.39$ \\
 & Hard & \cellcolor{gray!12}49.0 & \cellcolor{gray!30}$0.56\pm0.43$ & \cellcolor{gray!30}50.0 & \cellcolor{gray!12}$0.58\pm0.42$ & 41.0 & $0.65\pm0.42$ & 45.0 & $0.63\pm0.38$ \\
 & \emph{All} & \cellcolor{gray!30}62.7 & \cellcolor{gray!30}$0.42\pm0.43$ & 58.7 & \cellcolor{gray!12}$0.46\pm0.43$ & 52.3 & $0.53\pm0.45$ & \cellcolor{gray!12}62.0 & $0.48\pm0.40$ \\
\addlinespace
Giant-Navigate & Easy & \cellcolor{gray!12}44.0 & \cellcolor{gray!30}$0.53\pm0.46$ & \cellcolor{gray!30}47.0 & $0.57\pm0.46$ & \cellcolor{gray!30}47.0 & \cellcolor{gray!12}$0.56\pm0.46$ & 26.0 & $0.75\pm0.38$ \\
 & Medium & 37.0 & $0.66\pm0.43$ & \cellcolor{gray!12}39.0 & \cellcolor{gray!12}$0.64\pm0.43$ & \cellcolor{gray!30}40.0 & \cellcolor{gray!30}$0.63\pm0.44$ & 22.0 & $0.79\pm0.36$ \\
 & Hard & \cellcolor{gray!30}23.0 & \cellcolor{gray!30}$0.77\pm0.39$ & \cellcolor{gray!12}20.0 & \cellcolor{gray!12}$0.81\pm0.35$ & 18.0 & $0.84\pm0.34$ & 12.0 & $0.89\pm0.28$ \\
 & \emph{All} & 34.7 & \cellcolor{gray!30}$0.65\pm0.43$ & \cellcolor{gray!30}35.3 & \cellcolor{gray!12}$0.67\pm0.43$ & \cellcolor{gray!12}35.0 & $0.68\pm0.43$ & 20.0 & $0.81\pm0.35$ \\
\addlinespace
Giant-Stitch & Easy & \cellcolor{gray!30}52.0 & \cellcolor{gray!30}$0.48\pm0.45$ & \cellcolor{gray!12}42.0 & \cellcolor{gray!12}$0.57\pm0.45$ & 38.0 & $0.61\pm0.46$ & 16.0 & $0.86\pm0.32$ \\
 & Medium & \cellcolor{gray!30}44.0 & \cellcolor{gray!30}$0.57\pm0.44$ & \cellcolor{gray!12}40.0 & \cellcolor{gray!12}$0.60\pm0.44$ & 35.0 & $0.65\pm0.43$ & 10.0 & $0.91\pm0.27$ \\
 & Hard & \cellcolor{gray!30}37.0 & \cellcolor{gray!30}$0.67\pm0.43$ & \cellcolor{gray!12}34.0 & $0.68\pm0.42$ & 32.0 & \cellcolor{gray!12}$0.68\pm0.42$ & 9.0 & $0.92\pm0.26$ \\
 & \emph{All} & \cellcolor{gray!30}44.3 & \cellcolor{gray!30}$0.57\pm0.45$ & \cellcolor{gray!12}38.7 & \cellcolor{gray!12}$0.62\pm0.44$ & 35.0 & $0.65\pm0.44$ & 11.7 & $0.89\pm0.28$ \\
\midrule
\emph{Overall} & \emph{All} & \cellcolor{gray!30}57.4 & \cellcolor{gray!30}$0.45\pm0.44$ & \cellcolor{gray!12}54.3 & \cellcolor{gray!12}$0.49\pm0.44$ & 50.3 & $0.53\pm0.45$ & 48.1 & $0.56\pm0.43$ \\
\bottomrule
\end{tabular}
}
\end{table}

\subsection{Paired comparisons and confidence intervals}\label{app:paired-statistics}
We compare methods on matched tasks to separate method differences from
variation across settings. The navigation benchmark contains 4,200 tasks
shared by four methods, giving 16,800 method--task outcomes.
Each environment--difficulty group has 100 tasks. Domain-level means average
these equally sized groups, while their standard deviations describe
cross-setting variation. Pairing outcomes for the same formula and initial
state within each setting and difficulty yields 2,400 AntMaze pairs and
1,800 HumanoidMaze pairs for each method comparison.

We obtain paired-bootstrap intervals by resampling matched records within
each fixed environment stratum and taking the 2.5\% and 97.5\% percentiles.
For SR, two-sided exact McNemar tests use the discordant outcomes, with Holm
correction across 24 comparisons: three baselines, two domains, and four task
groups (all difficulties combined, Easy, Medium, and Hard). The six aggregate
comparisons in Table~\ref{tab:paired-effects} belong to this family.
Paired-bootstrap intervals also quantify NCC differences, while Wilson
intervals summarize SR within each setting and difficulty.

\begin{table}[htbp]
\centering\footnotesize
\caption{Aggregate paired effects (SAGAS minus baseline). Brackets are 95\%
paired-bootstrap intervals; $p$ denotes the Holm-corrected $p$-value from a
two-sided exact McNemar test for SR.}
\label{tab:paired-effects}
\begin{tabular}{llccc}
\toprule
Domain & Baseline & $\Delta$SR (pp) & $\Delta$NCC & $p$\\
\midrule
AntMaze & \LFGAS & $+4.9\ [3.1,6.7]$ & $-0.046\ [-0.061,-0.030]$ & $2.2\times10^{-6}$\\
 & \GAGAS & $+9.5\ [7.7,11.5]$ & $-0.101\ [-0.118,-0.083]$ & $2.8\times10^{-21}$\\
 & \APHIQL & $+23.0\ [20.7,25.3]$ & $-0.224\ [-0.245,-0.204]$ & $5.3\times10^{-67}$\\
HumanoidMaze & \LFGAS & $+3.1\ [1.0,5.3]$ & $-0.034\ [-0.052,-0.016]$ & $0.032$\\
 & \GAGAS & $+7.1\ [5.1,9.2]$ & $-0.076\ [-0.095,-0.058]$ & $9.8\times10^{-10}$\\
 & \APHIQL & $+9.3\ [6.7,12.0]$ & $-0.111\ [-0.133,-0.089]$ & $4.5\times10^{-10}$\\
\bottomrule
\end{tabular}
\end{table}

All six aggregate comparisons favor SAGAS: its SR intervals lie above zero
and its NCC intervals below zero. For \LFGAS{} and \GAGAS{}, the graph and
low-level policy are held fixed, so the paired differences reflect high-level
planning choices under a common executor, including how reliably the selected
waypoint sequences can be realized.

Aggregate SAGAS SR is $1724/2400=71.83\%$ in AntMaze (Wilson 95\% CI
$[70.0\%,73.6\%]$) and $1033/1800=57.39\%$ in HumanoidMaze
($[55.1\%,59.7\%]$). Table~\ref{tab:cell-wilson} reports Wilson intervals for
each method, environment setting, and difficulty.
\begingroup
\scriptsize
\setlength{\tabcolsep}{3pt}
\begin{longtable}{llcccc}
\caption{Per-setting finite-lasso SR (\%) with Wilson 95\% intervals. Each method--difficulty cell contains 100 tasks.}\label{tab:cell-wilson}\\
\toprule
Setting & Difficulty & SAGAS & \LFGAS & \GAGAS & \APHIQL\\
\midrule
\endfirsthead
\toprule
Setting & Difficulty & SAGAS & \LFGAS & \GAGAS & \APHIQL\\
\midrule
\endhead
\bottomrule
\endfoot
Ant giant-nav. & Easy & $61\ [51.2,70.0]$ & $62\ [52.2,70.9]$ & $55\ [45.2,64.4]$ & $62\ [52.2,70.9]$\\
Ant giant-nav. & Medium & $57\ [47.2,66.3]$ & $56\ [46.2,65.3]$ & $53\ [43.3,62.5]$ & $47\ [37.5,56.7]$\\
Ant giant-nav. & Hard & $48\ [38.5,57.7]$ & $37\ [28.2,46.8]$ & $45\ [35.6,54.8]$ & $37\ [28.2,46.8]$\\
\addlinespace[2pt]
Ant giant-stitch & Easy & $61\ [51.2,70.0]$ & $56\ [46.2,65.3]$ & $60\ [50.2,69.1]$ & $20\ [13.3,28.9]$\\
Ant giant-stitch & Medium & $56\ [46.2,65.3]$ & $50\ [40.4,59.6]$ & $45\ [35.6,54.8]$ & $10\ [5.5,17.4]$\\
Ant giant-stitch & Hard & $54\ [44.3,63.4]$ & $41\ [31.9,50.8]$ & $35\ [26.4,44.7]$ & $8\ [4.1,15.0]$\\
\addlinespace[2pt]
Ant large-expl. & Easy & $83\ [74.5,89.1]$ & $79\ [70.0,85.8]$ & $74\ [64.6,81.6]$ & $17\ [10.9,25.5]$\\
Ant large-expl. & Medium & $86\ [77.9,91.5]$ & $83\ [74.5,89.1]$ & $80\ [71.1,86.7]$ & $15\ [9.3,23.3]$\\
Ant large-expl. & Hard & $80\ [71.1,86.7]$ & $66\ [56.3,74.5]$ & $68\ [58.3,76.3]$ & $11\ [6.3,18.6]$\\
\addlinespace[2pt]
Ant large-nav. & Easy & $73\ [63.6,80.7]$ & $67\ [57.3,75.4]$ & $59\ [49.2,68.1]$ & $77\ [67.8,84.2]$\\
Ant large-nav. & Medium & $63\ [53.2,71.8]$ & $67\ [57.3,75.4]$ & $68\ [58.3,76.3]$ & $71\ [61.5,79.0]$\\
Ant large-nav. & Hard & $55\ [45.2,64.4]$ & $49\ [39.4,58.7]$ & $45\ [35.6,54.8]$ & $51\ [41.3,60.6]$\\
\addlinespace[2pt]
Ant large-stitch & Easy & $74\ [64.6,81.6]$ & $73\ [63.6,80.7]$ & $65\ [55.3,73.6]$ & $67\ [57.3,75.4]$\\
Ant large-stitch & Medium & $81\ [72.2,87.5]$ & $79\ [70.0,85.8]$ & $73\ [63.6,80.7]$ & $61\ [51.2,70.0]$\\
Ant large-stitch & Hard & $62\ [52.2,70.9]$ & $49\ [39.4,58.7]$ & $52\ [42.3,61.5]$ & $45\ [35.6,54.8]$\\
\addlinespace[2pt]
Ant medium-expl. & Easy & $92\ [85.0,95.9]$ & $92\ [85.0,95.9]$ & $81\ [72.2,87.5]$ & $53\ [43.3,62.5]$\\
Ant medium-expl. & Medium & $90\ [82.6,94.5]$ & $82\ [73.3,88.3]$ & $73\ [63.6,80.7]$ & $63\ [53.2,71.8]$\\
Ant medium-expl. & Hard & $85\ [76.7,90.7]$ & $80\ [71.1,86.7]$ & $74\ [64.6,81.6]$ & $44\ [34.7,53.8]$\\
\addlinespace[2pt]
Ant medium-nav. & Easy & $78\ [68.9,85.0]$ & $79\ [70.0,85.8]$ & $68\ [58.3,76.3]$ & $70\ [60.4,78.1]$\\
Ant medium-nav. & Medium & $83\ [74.5,89.1]$ & $74\ [64.6,81.6]$ & $60\ [50.2,69.1]$ & $70\ [60.4,78.1]$\\
Ant medium-nav. & Hard & $65\ [55.3,73.6]$ & $60\ [50.2,69.1]$ & $54\ [44.3,63.4]$ & $61\ [51.2,70.0]$\\
\addlinespace[2pt]
Ant medium-stitch & Easy & $81\ [72.2,87.5]$ & $80\ [71.1,86.7]$ & $75\ [65.7,82.5]$ & $82\ [73.3,88.3]$\\
Ant medium-stitch & Medium & $85\ [76.7,90.7]$ & $81\ [72.2,87.5]$ & $77\ [67.8,84.2]$ & $69\ [59.4,77.2]$\\
Ant medium-stitch & Hard & $71\ [61.5,79.0]$ & $65\ [55.3,73.6]$ & $56\ [46.2,65.3]$ & $61\ [51.2,70.0]$\\
\addlinespace[2pt]
Humanoid giant-nav. & Easy & $44\ [34.7,53.8]$ & $47\ [37.5,56.7]$ & $47\ [37.5,56.7]$ & $26\ [18.4,35.4]$\\
Humanoid giant-nav. & Medium & $37\ [28.2,46.8]$ & $39\ [30.0,48.8]$ & $40\ [30.9,49.8]$ & $22\ [15.0,31.1]$\\
Humanoid giant-nav. & Hard & $23\ [15.8,32.2]$ & $20\ [13.3,28.9]$ & $18\ [11.7,26.7]$ & $12\ [7.0,19.8]$\\
\addlinespace[2pt]
Humanoid giant-stitch & Easy & $52\ [42.3,61.5]$ & $42\ [32.8,51.8]$ & $38\ [29.1,47.8]$ & $16\ [10.1,24.4]$\\
Humanoid giant-stitch & Medium & $44\ [34.7,53.8]$ & $40\ [30.9,49.8]$ & $35\ [26.4,44.7]$ & $10\ [5.5,17.4]$\\
Humanoid giant-stitch & Hard & $37\ [28.2,46.8]$ & $34\ [25.5,43.7]$ & $32\ [23.7,41.7]$ & $9\ [4.8,16.2]$\\
\addlinespace[2pt]
Humanoid large-nav. & Easy & $69\ [59.4,77.2]$ & $73\ [63.6,80.7]$ & $66\ [56.3,74.5]$ & $69\ [59.4,77.2]$\\
Humanoid large-nav. & Medium & $66\ [56.3,74.5]$ & $61\ [51.2,70.0]$ & $55\ [45.2,64.4]$ & $62\ [52.2,70.9]$\\
Humanoid large-nav. & Hard & $57\ [47.2,66.3]$ & $44\ [34.7,53.8]$ & $45\ [35.6,54.8]$ & $48\ [38.5,57.7]$\\
\addlinespace[2pt]
Humanoid large-stitch & Easy & $75\ [65.7,82.5]$ & $66\ [56.3,74.5]$ & $61\ [51.2,70.0]$ & $74\ [64.6,81.6]$\\
Humanoid large-stitch & Medium & $64\ [54.2,72.7]$ & $60\ [50.2,69.1]$ & $55\ [45.2,64.4]$ & $67\ [57.3,75.4]$\\
Humanoid large-stitch & Hard & $49\ [39.4,58.7]$ & $50\ [40.4,59.6]$ & $41\ [31.9,50.8]$ & $45\ [35.6,54.8]$\\
\addlinespace[2pt]
Humanoid medium-nav. & Easy & $81\ [72.2,87.5]$ & $78\ [68.9,85.0]$ & $76\ [66.8,83.3]$ & $75\ [65.7,82.5]$\\
Humanoid medium-nav. & Medium & $66\ [56.3,74.5]$ & $66\ [56.3,74.5]$ & $55\ [45.2,64.4]$ & $77\ [67.8,84.2]$\\
Humanoid medium-nav. & Hard & $67\ [57.3,75.4]$ & $57\ [47.2,66.3]$ & $55\ [45.2,64.4]$ & $61\ [51.2,70.0]$\\
\addlinespace[2pt]
Humanoid medium-stitch & Easy & $68\ [58.3,76.3]$ & $72\ [62.5,79.9]$ & $69\ [59.4,77.2]$ & $74\ [64.6,81.6]$\\
Humanoid medium-stitch & Medium & $70\ [60.4,78.1]$ & $74\ [64.6,81.6]$ & $61\ [51.2,70.0]$ & $67\ [57.3,75.4]$\\
Humanoid medium-stitch & Hard & $64\ [54.2,72.7]$ & $54\ [44.3,63.4]$ & $56\ [46.2,65.3]$ & $51\ [41.3,60.6]$\\
\addlinespace[2pt]
\end{longtable}
\endgroup

\subsection{Execution length on common-success tasks}
\label{app:common-success}
NCC summarizes execution cost over all tasks, including failures. To compare how many steps the methods take when they all succeed, we restrict Tables~\ref{tab:antmaze-ltl-common-success-len} and~\ref{tab:humanoid-ltl-common-success-len} to tasks on which every method shown meets the finite-lasso criterion.
Each row reports the number of such tasks and the mean number of environment steps each method takes to complete the prefix and first suffix traversal.

\begin{table}[H]
\centering
\footnotesize
\setlength{\tabcolsep}{3.5pt}
\caption{AntMaze common-success execution-length results. Each row is restricted to test cases solved by all displayed methods. $N$ is the number of shared successful cases; entries report mean$\pm$standard deviation of execution length in environment steps for prefix completion plus one suffix traversal. Dark/light gray shading marks the shortest/second-shortest method within each setting.}
\label{tab:antmaze-ltl-common-success-len}
\begin{tabular}{llrrrrr}
\toprule
Environment & Difficulty & $N$ & SAGAS & \LFGAS{} & \GAGAS{} & \APHIQL{} \\
\midrule
Medium-Navigate & Easy & 47 & \cellcolor{gray!30}$527\pm342$ & $587\pm367$ & \cellcolor{gray!12}$543\pm357$ & $617\pm352$ \\
 & Medium & 46 & \cellcolor{gray!30}$515\pm325$ & $557\pm335$ & \cellcolor{gray!12}$529\pm325$ & $628\pm440$ \\
 & Hard & 33 & \cellcolor{gray!30}$661\pm476$ & $731\pm470$ & \cellcolor{gray!12}$697\pm522$ & $886\pm584$ \\
 & \emph{All} & 126 & \cellcolor{gray!30}$557\pm378$ & $614\pm389$ & \cellcolor{gray!12}$578\pm400$ & $692\pm465$ \\
\addlinespace
Large-Navigate & Easy & 45 & $919\pm1306$ & $843\pm746$ & \cellcolor{gray!12}$796\pm733$ & \cellcolor{gray!30}$761\pm564$ \\
 & Medium & 39 & $865\pm556$ & $994\pm660$ & \cellcolor{gray!12}$862\pm602$ & \cellcolor{gray!30}$859\pm495$ \\
 & Hard & 21 & \cellcolor{gray!12}$1165\pm1004$ & $1325\pm1006$ & $1267\pm956$ & \cellcolor{gray!30}$1119\pm649$ \\
 & \emph{All} & 105 & $948\pm1020$ & $995\pm788$ & \cellcolor{gray!12}$915\pm753$ & \cellcolor{gray!30}$869\pm568$ \\
\addlinespace
Giant-Navigate & Easy & 30 & \cellcolor{gray!30}$905\pm631$ & $1033\pm636$ & \cellcolor{gray!12}$940\pm674$ & $1235\pm868$ \\
 & Medium & 24 & \cellcolor{gray!30}$1454\pm654$ & $1548\pm637$ & \cellcolor{gray!12}$1490\pm706$ & $1727\pm884$ \\
 & Hard & 15 & \cellcolor{gray!30}$1214\pm971$ & $1415\pm1291$ & \cellcolor{gray!12}$1289\pm1201$ & $1770\pm1602$ \\
 & \emph{All} & 69 & \cellcolor{gray!30}$1163\pm754$ & $1295\pm842$ & \cellcolor{gray!12}$1208\pm849$ & $1522\pm1086$ \\
\addlinespace
Medium-Stitch & Easy & 59 & \cellcolor{gray!12}$544\pm357$ & $558\pm317$ & \cellcolor{gray!30}$524\pm333$ & $733\pm492$ \\
 & Medium & 50 & \cellcolor{gray!30}$505\pm258$ & \cellcolor{gray!12}$543\pm261$ & $549\pm334$ & $832\pm501$ \\
 & Hard & 31 & \cellcolor{gray!30}$681\pm372$ & $793\pm476$ & \cellcolor{gray!12}$712\pm397$ & $950\pm572$ \\
 & \emph{All} & 140 & \cellcolor{gray!30}$561\pm333$ & $605\pm353$ & \cellcolor{gray!12}$575\pm354$ & $816\pm517$ \\
\addlinespace
Large-Stitch & Easy & 44 & \cellcolor{gray!30}$758\pm637$ & $856\pm649$ & \cellcolor{gray!12}$767\pm641$ & $852\pm733$ \\
 & Medium & 44 & \cellcolor{gray!30}$763\pm459$ & $888\pm495$ & \cellcolor{gray!12}$788\pm472$ & $951\pm574$ \\
 & Hard & 21 & \cellcolor{gray!30}$1189\pm590$ & \cellcolor{gray!12}$1289\pm632$ & $1302\pm744$ & $1498\pm719$ \\
 & \emph{All} & 109 & \cellcolor{gray!30}$843\pm582$ & $953\pm606$ & \cellcolor{gray!12}$878\pm631$ & $1017\pm706$ \\
\addlinespace
Giant-Stitch & Easy & 12 & \cellcolor{gray!12}$1105\pm1173$ & $1458\pm1307$ & \cellcolor{gray!30}$1056\pm1205$ & $1600\pm1241$ \\
 & Medium & 4 & \cellcolor{gray!30}$1413\pm1612$ & \cellcolor{gray!12}$1530\pm1536$ & \cellcolor{gray!30}$1413\pm1612$ & $1788\pm1382$ \\
 & Hard & 4 & \cellcolor{gray!30}$664\pm366$ & \cellcolor{gray!12}$673\pm376$ & \cellcolor{gray!12}$673\pm376$ & $1419\pm891$ \\
 & \emph{All} & 20 & \cellcolor{gray!12}$1079\pm1135$ & $1316\pm1222$ & \cellcolor{gray!30}$1051\pm1154$ & $1601\pm1155$ \\
\addlinespace
Medium-Explore & Easy & 47 & \cellcolor{gray!30}$315\pm242$ & $396\pm303$ & \cellcolor{gray!12}$380\pm392$ & $1298\pm1413$ \\
 & Medium & 50 & \cellcolor{gray!30}$407\pm220$ & $433\pm228$ & \cellcolor{gray!12}$433\pm269$ & $1619\pm1882$ \\
 & Hard & 33 & \cellcolor{gray!30}$456\pm267$ & $592\pm389$ & \cellcolor{gray!12}$545\pm386$ & $1544\pm1131$ \\
 & \emph{All} & 130 & \cellcolor{gray!30}$386\pm246$ & $460\pm310$ & \cellcolor{gray!12}$442\pm351$ & $1484\pm1548$ \\
\addlinespace
Large-Explore & Easy & 15 & \cellcolor{gray!12}$233\pm266$ & $257\pm258$ & \cellcolor{gray!30}$232\pm265$ & $420\pm304$ \\
 & Medium & 15 & \cellcolor{gray!30}$225\pm174$ & \cellcolor{gray!12}$263\pm194$ & $271\pm258$ & $1012\pm1420$ \\
 & Hard & 8 & $1014\pm1666$ & $1004\pm1658$ & \cellcolor{gray!12}$977\pm1671$ & \cellcolor{gray!30}$859\pm613$ \\
 & \emph{All} & 38 & \cellcolor{gray!30}$394\pm818$ & $416\pm809$ & \cellcolor{gray!12}$404\pm819$ & $746\pm972$ \\
\midrule
\emph{Overall} & \emph{All} & 737 & \cellcolor{gray!30}$688\pm661$ & $762\pm651$ & \cellcolor{gray!12}$709\pm637$ & $1034\pm960$ \\
\bottomrule
\end{tabular}
\end{table}

\begin{table}[H]
\centering
\footnotesize
\setlength{\tabcolsep}{3.5pt}
\caption{HumanoidMaze common-success execution-length results. Each row is restricted to test cases solved by all displayed methods. $N$ is the number of shared successful cases; entries report mean$\pm$standard deviation of execution length in environment steps for prefix completion plus one suffix traversal. Dark/light gray shading marks the shortest/second-shortest method within each setting.}
\label{tab:humanoid-ltl-common-success-len}
\begin{tabular}{llrrrrr}
\toprule
Environment & Difficulty & $N$ & SAGAS & \LFGAS{} & \GAGAS{} & \APHIQL{} \\
\midrule
Medium-Navigate & Easy & 56 & $1752\pm2487$ & \cellcolor{gray!30}$1687\pm2323$ & \cellcolor{gray!12}$1704\pm2371$ & $1925\pm1332$ \\
 & Medium & 42 & \cellcolor{gray!30}$2103\pm2222$ & $2634\pm2515$ & $2388\pm2546$ & \cellcolor{gray!12}$2343\pm1453$ \\
 & Hard & 34 & \cellcolor{gray!30}$1953\pm1457$ & $2251\pm1648$ & \cellcolor{gray!12}$2096\pm1611$ & $2527\pm1421$ \\
 & \emph{All} & 132 & \cellcolor{gray!30}$1916\pm2168$ & $2134\pm2258$ & \cellcolor{gray!12}$2023\pm2265$ & $2213\pm1407$ \\
\addlinespace
Medium-Stitch & Easy & 43 & \cellcolor{gray!30}$1576\pm1340$ & \cellcolor{gray!12}$1590\pm1230$ & $1667\pm1318$ & $1894\pm1665$ \\
 & Medium & 41 & \cellcolor{gray!30}$1805\pm1204$ & $1981\pm1396$ & \cellcolor{gray!12}$1938\pm1775$ & $2270\pm1350$ \\
 & Hard & 26 & \cellcolor{gray!30}$2049\pm1274$ & $2359\pm1521$ & \cellcolor{gray!12}$2218\pm1536$ & $3738\pm3205$ \\
 & \emph{All} & 110 & \cellcolor{gray!30}$1773\pm1277$ & $1917\pm1385$ & \cellcolor{gray!12}$1898\pm1553$ & $2470\pm2150$ \\
\addlinespace
Large-Navigate & Easy & 44 & $2529\pm2615$ & \cellcolor{gray!30}$2453\pm2369$ & \cellcolor{gray!12}$2489\pm2543$ & $2968\pm2519$ \\
 & Medium & 33 & $2993\pm3156$ & $3305\pm3103$ & \cellcolor{gray!30}$2893\pm3113$ & \cellcolor{gray!12}$2911\pm2443$ \\
 & Hard & 23 & \cellcolor{gray!30}$3352\pm2115$ & \cellcolor{gray!12}$3362\pm2001$ & $3391\pm2005$ & $4496\pm3116$ \\
 & \emph{All} & 100 & \cellcolor{gray!12}$2872\pm2701$ & $2943\pm2575$ & \cellcolor{gray!30}$2830\pm2638$ & $3301\pm2697$ \\
\addlinespace
Large-Stitch & Easy & 51 & \cellcolor{gray!30}$2745\pm2161$ & $3432\pm2979$ & \cellcolor{gray!12}$3009\pm2674$ & $4673\pm2983$ \\
 & Medium & 35 & \cellcolor{gray!30}$2711\pm2243$ & \cellcolor{gray!12}$3054\pm2556$ & $3117\pm2807$ & $5122\pm3650$ \\
 & Hard & 16 & \cellcolor{gray!30}$3582\pm1777$ & $4844\pm3596$ & \cellcolor{gray!12}$3710\pm1892$ & $5459\pm2820$ \\
 & \emph{All} & 102 & \cellcolor{gray!30}$2864\pm2138$ & $3524\pm2978$ & \cellcolor{gray!12}$3156\pm2605$ & $4950\pm3188$ \\
\addlinespace
Giant-Navigate & Easy & 14 & \cellcolor{gray!12}$1935\pm2252$ & \cellcolor{gray!30}$1650\pm1577$ & \cellcolor{gray!12}$1935\pm2252$ & $4651\pm4257$ \\
 & Medium & 10 & \cellcolor{gray!30}$3526\pm4211$ & \cellcolor{gray!12}$4120\pm4593$ & \cellcolor{gray!12}$4120\pm4593$ & $5211\pm2607$ \\
 & Hard & 3 & \cellcolor{gray!12}$1232\pm1320$ & \cellcolor{gray!30}$1143\pm584$ & \cellcolor{gray!12}$1232\pm1320$ & $2356\pm2733$ \\
 & \emph{All} & 27 & \cellcolor{gray!30}$2446\pm3093$ & \cellcolor{gray!12}$2508\pm3191$ & $2666\pm3363$ & $4604\pm3566$ \\
\addlinespace
Giant-Stitch & Easy & 10 & \cellcolor{gray!30}$1696\pm1485$ & $1956\pm1414$ & \cellcolor{gray!12}$1708\pm1476$ & $4088\pm4014$ \\
 & Medium & 7 & \cellcolor{gray!12}$2738\pm2147$ & $3031\pm2046$ & \cellcolor{gray!30}$2687\pm2131$ & $7303\pm7000$ \\
 & Hard & 6 & \cellcolor{gray!30}$1981\pm2216$ & \cellcolor{gray!12}$2024\pm2183$ & \cellcolor{gray!30}$1981\pm2216$ & $2510\pm1764$ \\
 & \emph{All} & 23 & \cellcolor{gray!12}$2087\pm1866$ & $2301\pm1813$ & \cellcolor{gray!30}$2077\pm1852$ & $4655\pm4929$ \\
\midrule
\emph{Overall} & \emph{All} & 494 & \cellcolor{gray!30}$2310\pm2215$ & $2565\pm2445$ & \cellcolor{gray!12}$2430\pm2386$ & $3300\pm2838$ \\
\bottomrule
\end{tabular}
\end{table}

We also form a separate common-success subset for each SAGAS--baseline pair and compare execution lengths on that subset. SAGAS uses on average 103.3 fewer environment steps than \LFGAS{} in AntMaze (95\% CI $[81.6,126.5]$) and 210.0 fewer in HumanoidMaze ($[102.1,313.8]$), again measured over the prefix and first suffix traversal.

\subsection{SAGAS and \APHIQL{} across AntMaze datasets}\label{app:regime-analysis}
To examine how relative performance varies across offline datasets, we compare SAGAS and \APHIQL{} separately on Navigate, Stitch, and Explore using the 2,400 matched AntMaze tasks from Appendix~\ref{app:paired-statistics}.
\begin{table}[htbp]
\centering\small
\caption{Success rates by AntMaze dataset type. $\Delta$SR is the difference in percentage points (SAGAS minus \APHIQL{}). Brackets give 95\% confidence intervals from 20,000 paired bootstrap replicates stratified by environment.}
\label{tab:regime-effects}
\begin{tabular}{lrrrr}
\toprule
Dataset type & Tasks & SAGAS SR & \APHIQL{} SR & $\Delta$SR (pp)\\
\midrule
Navigate & 900 & 64.78\% & 60.67\% & $+4.11\ [0.22,8.11]$\\
Stitch & 900 & 69.44\% & 47.00\% & $+22.44\ [18.56,26.33]$\\
Explore & 600 & 86.00\% & 33.83\% & $+52.17\ [48.17,56.33]$\\
All & 2,400 & 71.83\% & 48.83\% & $+23.00\ [20.67,25.33]$\\
\bottomrule
\end{tabular}
\end{table}
\APHIQL{} is most competitive on Navigate, whose offline data contain longer, more coherent trajectories.
SAGAS's SR advantage increases from 4.11 percentage points on Navigate to 22.44 on Stitch and 52.17 on Explore, where the data consist of fragmented or exploratory trajectories.

\FloatBarrier
\subsection{Overlapping-region and Scene manipulation results}
\label{app:predicate-results}

The two predicate suites test simultaneous proposition assignments and
joint object configurations. Their results use the respective protocols in
Appendices~\ref{app:overlap-protocol} and~\ref{app:scene-protocol}.

\paragraph{Overlapping regions.}
SAGAS finds accepting lassos for 292 of the 320 tasks across AntMaze
Giant-Stitch and Large-Explore. Independent post-rollout verification of the
saved state trajectories confirms finite-lasso success on 260 of the 320
tasks, giving $81.3\%$ SR (Wilson 95\% CI
$[76.6\%,85.1\%]$). Both environments achieve 130/160 successes, but
performance varies by task family: ordered overlap has the lowest SR in
Giant-Stitch, whereas negative-until has the lowest SR in Large-Explore
(Table~\ref{tab:predicate-family-results}).

\paragraph{Joint object configurations.}
SAGAS finds accepting lassos for all 120 Scene tasks, and independent
post-rollout verification confirms finite-lasso success on 81, giving
$67.5\%$ SR (Wilson 95\% CI
$[58.7\%,75.2\%]$). Table~\ref{tab:predicate-family-results} breaks these
results down by task family. Of the 39 failed
executions, 25 fail to reach a target and 11 exceed the local step budget.
The remaining failures are two negative-literal violations and one segment
whose observed labels admit no automaton run between its planned source
and target states.

\begin{table}[tbp]
\centering
\small
\caption{Predicate generalization. SR is finite-lasso execution success rate, determined by independent post-rollout verification. The suites use their stated suffix criteria and are reported separately. Per-family results are in Table~\ref{tab:predicate-family-results}.}
\label{tab:predicate-generalization}
\begin{tabular}{lrrr}
\toprule
Setting / task construction & Tasks & Accepting lassos & SR\\
\midrule
\multicolumn{4}{l}{\textbf{(a) AntMaze: overlapping regions, $M=2$}}\\
Giant-Stitch & 160 & 149/160 & 130/160 (81.3\%)\\
Large-Explore & 160 & 143/160 & 130/160 (81.3\%)\\
All overlapping-region tasks & 320 & 292/320 & 260/320 (81.3\%)\\
\midrule
\multicolumn{4}{l}{\textbf{(b) OGBench Scene: joint object configurations, $M=1$}}\\
24 formulas $\times$ five initial states & 120 & 120/120 & 81/120 (67.5\%)\\
\bottomrule
\end{tabular}
\end{table}

\begin{table}[H]
\centering\small
\caption{Per-family predicate-generalization results. SR is computed over all tasks in each family using independent post-rollout finite-lasso verification.}
\label{tab:predicate-family-results}
\begin{tabular}{lrrr}
\toprule
Task family & Tasks & Accepting lassos & SR\\
\midrule
\multicolumn{4}{l}{AntMaze Giant-Stitch ($M=2$)}\\
Overlap event & 40 & 36/40 & 33/40 (82.5\%)\\
Negative-until overlap & 40 & 37/40 & 34/40 (85.0\%)\\
Recurrent overlap & 40 & 37/40 & 34/40 (85.0\%)\\
Ordered overlap & 40 & 39/40 & 29/40 (72.5\%)\\
\midrule
\multicolumn{4}{l}{AntMaze Large-Explore ($M=2$)}\\
Overlap event & 40 & 38/40 & 36/40 (90.0\%)\\
Negative-until overlap & 40 & 31/40 & 28/40 (70.0\%)\\
Recurrent overlap & 40 & 37/40 & 32/40 (80.0\%)\\
Ordered overlap & 40 & 37/40 & 34/40 (85.0\%)\\
\midrule
\multicolumn{4}{l}{OGBench Scene ($M=1$)}\\
Single-stage configuration & 20 & 20/20 & 17/20 (85.0\%)\\
Ordered two-stage manipulation & 20 & 20/20 & 14/20 (70.0\%)\\
Ordered three-stage manipulation & 20 & 20/20 & 11/20 (55.0\%)\\
Choice, then configuration & 20 & 20/20 & 13/20 (65.0\%)\\
Negative-until & 20 & 20/20 & 11/20 (55.0\%)\\
Recurrent suffix & 20 & 20/20 & 15/20 (75.0\%)\\
\bottomrule
\end{tabular}
\end{table}

\section{Failure-mode analysis}
\label{app:failure-analysis}

To identify where tasks fail, we group the SAGAS outcomes from the navigation benchmark in Tables~\ref{tab:antmaze-ltl-main} and~\ref{tab:humanoid-ltl-main}.
Of the 4,200 cases, 2757 succeed and 1443 fail, giving an overall SR of $65.6\%$.
Table~\ref{tab:sagas-failure-modes} assigns each failure to the first stage at which progress toward the finite-lasso criterion stops.

\begin{table}[H]
\centering
\small
\caption{Aggregate SAGAS failure modes over 4,200 LTL test cases. Percentages are reported relative to failed cases and to all cases.}
\label{tab:sagas-failure-modes}
\begin{tabular}{lrrr}
\toprule
Failure mode & Count & Failed cases (\%) & All cases (\%) \\
\midrule
Prefix execution or semantic realization failure & 930 & 64.4 & 22.1 \\
Suffix-cycle or finite-lasso realization failure & 423 & 29.3 & 10.1 \\
No accepting prefix or suffix found & 90 & 6.2 & 2.1 \\
\bottomrule
\end{tabular}
\end{table}

More than $93\%$ of failures occur during physical execution after a candidate prefix or suffix has been found. Prefix-stage locomotion failure, such as stalling or overturning, is the most frequent cause (748 cases). The remaining execution failures involve incomplete first and second suffix traversals (201 and 136), incomplete prefix tracking (100), or failed semantic checks during the suffix and prefix (86 and 82). Search finds no accepting prefix in 69 cases and no suffix for the retained prefixes in 21 cases.

\begin{table}[H]
\centering
\small
\caption{Domain-wise SAGAS failure breakdown. Percentages in the last three columns are relative to failures within the corresponding domain.}
\label{tab:sagas-domain-failure}
\resizebox{\linewidth}{!}{%
\begin{tabular}{lrrrrrr}
\toprule
Domain & Cases & Successes & SR (\%) & Prefix failure & Suffix failure & Planning failure \\
\midrule
AntMaze & 2400 & 1724 & 71.8 & 417 (61.7\%) & 206 (30.5\%) & 53 (7.8\%) \\
HumanoidMaze & 1800 & 1033 & 57.4 & 513 (66.9\%) & 217 (28.3\%) & 37 (4.8\%) \\
\bottomrule
\end{tabular}%
}
\end{table}

Execution failures are more frequent in HumanoidMaze, while planning failures remain uncommon in both domains (Table~\ref{tab:sagas-domain-failure}). Giant mazes have the lowest SR, at $34.7\%$ for \texttt{humanoidmaze-giant-navigate}, $44.3\%$ for \texttt{humanoidmaze-giant-stitch}, $55.3\%$ for \texttt{antmaze-giant-navigate}, and $57.0\%$ for \texttt{antmaze-giant-stitch}. Longer waypoint chains expose the frozen executor to repeated tracking and suffix-stability challenges. Appendix~\ref{app:paired-statistics} compares SAGAS with the baselines on matched tasks.

Although every sampled proposition has connected anchors, prefix or suffix search fails in 90 cases under the chosen graph scale $H_{\mathrm{TD}}$ and candidate budget, particularly for formulas with more labels and negative or until constraints.

\section{Component and sensitivity analysis}
\label{app:ablations-diagnostics}

The following analyses examine how semantic screening, execution steering, and learned graph costs contribute to planning and execution. We then vary graph resolution, data availability, and predicate constraints to measure their effects on plan feasibility and computation.

\subsection{Soft-label screening and guard-aware steering}
\label{app:semantic-steering-ablation}
The paired ablations use the same 100 tasks per difficulty as the main benchmark on \texttt{antmaze-large-explore} and \texttt{humanoidmaze-medium-stitch}. The former tests semantic filtering under low-quality, broad-coverage data; the latter adds more demanding locomotion dynamics. Tasks, initial states, graph, and frozen policy are held fixed.

The \textbf{w/o Soft} variant removes negative-literal soft-label screening from product search while retaining anchors and execution rules. The \textbf{w/o Steering} variant keeps the selected plan and removes only the guard-aware steering term during execution. The two comparisons thus separate planning-time screening from execution-time adjustment.

\begin{table}[H]
\centering
\small
\setlength{\tabcolsep}{3.0pt}
\caption{Soft-label screening and guard-aware steering ablations. Each setting contains 100 tasks per difficulty; Average rows pool all 300 tasks. SR and NCC follow Appendix~\ref{app:evaluation-metrics}, with NCC reported as mean$\pm$standard error. Higher SR and lower NCC are better.}
\label{tab:soft-steering-ablation}
\resizebox{\linewidth}{!}{%
\begin{tabular}{llrrrrrr}
\toprule
Environment & Difficulty & \multicolumn{2}{c}{SAGAS} & \multicolumn{2}{c}{w/o soft} & \multicolumn{2}{c}{w/o steering} \\
\cmidrule(lr){3-4} \cmidrule(lr){5-6} \cmidrule(lr){7-8}
 & & SR (\%) & NCC & SR (\%) & NCC & SR (\%) & NCC \\
\midrule
Ant Large-Explore & Easy & $83.0$ & $0.198\pm0.035$ & $82.0$ & $0.208\pm0.036$ & $82.0$ & $0.209\pm0.036$ \\
 & Medium & $86.0$ & $0.162\pm0.031$ & $82.0$ & $0.200\pm0.035$ & $85.0$ & $0.174\pm0.032$ \\
 & Hard & $80.0$ & $0.263\pm0.036$ & $77.0$ & $0.289\pm0.038$ & $78.0$ & $0.269\pm0.037$ \\
 & Average & $83.0$ & $0.208\pm0.020$ & $80.3$ & $0.232\pm0.021$ & $81.7$ & $0.217\pm0.020$ \\
\midrule
Humanoid Medium-Stitch & Easy & $68.0$ & $0.312\pm0.041$ & $68.0$ & $0.312\pm0.041$ & $68.0$ & $0.314\pm0.041$ \\
 & Medium & $70.0$ & $0.316\pm0.039$ & $69.0$ & $0.325\pm0.040$ & $70.0$ & $0.317\pm0.039$ \\
 & Hard & $64.0$ & $0.387\pm0.041$ & $63.0$ & $0.396\pm0.041$ & $60.0$ & $0.426\pm0.043$ \\
 & Average & $67.3$ & $0.338\pm0.023$ & $66.7$ & $0.344\pm0.023$ & $66.0$ & $0.352\pm0.024$ \\
\bottomrule
\end{tabular}%
}
\end{table}

Removing soft-label screening lowers average AntMaze Large-Explore SR from $83.0\%$ to $80.3\%$ and raises NCC from $0.208$ to $0.232$, with the largest differences on Medium and Hard tasks (Table~\ref{tab:soft-steering-ablation}). Removing steering has a smaller effect in the same direction. In HumanoidMaze Medium-Stitch, soft labels mainly improve NCC, while steering is most useful on Hard tasks: without it, SR falls from $64.0\%$ to $60.0\%$ and NCC rises from $0.387$ to $0.426$. The full method achieves higher aggregate SR and lower NCC than either ablated variant in both environments.

\subsection{Graph cost and realized execution length}\label{app:surrogate-cost}
SAGAS ranks candidate plans using learned temporal-distance edge costs.
We examine whether the resulting graph cost reflects the number of
environment steps needed to execute a plan.

The analysis uses 140 successful SAGAS rollouts, with 20
difficulty-stratified tasks from each of seven settings. For each rollout,
we pair the planned graph cost with the realized number of environment
steps for the same prefix and first suffix traversal. We compute a
Spearman rank correlation separately for each setting.
\begin{table}[H]
\centering\small
\caption{Across-task association between graph cost and realized steps on
successful executions. Each setting contributes 20 trajectories.}
\label{tab:cost-association}
\begin{tabular}{lr}
\toprule
Setting & Spearman $\rho$\\
\midrule
AntMaze Giant-Navigate & 0.974\\
AntMaze Giant-Stitch & 0.887\\
AntMaze Large-Explore & 0.746\\
AntMaze Large-Navigate & 0.973\\
AntMaze Large-Stitch & 0.979\\
HumanoidMaze Giant-Stitch & 0.893\\
HumanoidMaze Medium-Stitch & 0.752\\
\bottomrule
\end{tabular}
\end{table}
Correlations range from $0.746$ to $0.979$, with a median of $0.893$
and a stratified-bootstrap 95\% interval of $[0.823,0.968]$ for the median.
Within each setting, higher graph costs generally correspond to more
execution steps, supporting the use of learned temporal-distance costs
as a proxy for execution length.

\subsection{Graph resolution and data subsampling}\label{app:horizon-sensitivity}
We examine how graph resolution and the amount of offline data used for
graph construction affect graph size, accepting-lasso discovery, and
planning time. For graph resolution, we compare temporal-distance scales
at 0.75, 1, and 1.25 times the domain defaults:
$H_{\mathrm{TD}}\in\{6,8,10\}$ in AntMaze Large-Stitch and Large-Explore,
and $H_{\mathrm{TD}}\in\{24,32,40\}$ in HumanoidMaze Medium-Stitch.
Each setting uses the same 30 Medium and 30 Hard tasks across scales,
with initial states, learned representation, semantic thresholds, and search
budget held fixed.

For data subsampling, we use a fixed seed to randomly select half of the
complete trajectories in AntMaze Large-Explore, retaining 2.5 million of the
five million states. We build an additional graph at the default scale from
this subset, which also supplies node support and anchor candidates.
The representation remains the one trained on the full dataset.
We evaluate the graph on the same 60 tasks as the full-data
graphs.
\begin{table}[H]
\centering\small
\caption{Graph resolution and data subsampling. Data denotes the fraction of
offline trajectories used for graph construction, node support, and anchor
retrieval. Accepting-lasso counts cover all 30 tasks per difficulty in each
condition. Time is median total planning time in seconds, including semantic
augmentation, automaton construction, graph copying, and search.}
\label{tab:horizon-sensitivity}
\begin{tabular}{rrrrrrrr}
\toprule
$H_{\mathrm{TD}}$ & Data & Nodes & Edges & \multicolumn{2}{c}{Accepting lassos} & \multicolumn{2}{c}{Time (s)}\\
\cmidrule(lr){5-6}\cmidrule(lr){7-8}
 & & & & Medium & Hard & Medium & Hard\\
\midrule
\multicolumn{8}{l}{AntMaze Large-Stitch (30 tasks per difficulty)}\\
6 & 100\% & 3,500 & 86,194 & 30/30 & 27/30 & 3.80 & 10.08\\
8 & 100\% & 1,563 & 29,166 & 30/30 & 28/30 & 1.81 & 4.10\\
10 & 100\% & 855 & 10,710 & 30/30 & 27/30 & 1.13 & 2.01\\
\midrule
\multicolumn{8}{l}{AntMaze Large-Explore (30 tasks per difficulty)}\\
6 & 100\% & 4,986 & 77,078 & 30/30 & 30/30 & 5.01 & 9.97\\
8 & 100\% & 2,525 & 34,774 & 30/30 & 29/30 & 3.61 & 5.97\\
10 & 100\% & 1,567 & 24,954 & 30/30 & 29/30 & 3.29 & 5.01\\
8 & 50\% & 1,798 & 22,402 & 30/30 & 29/30 & 1.89 & 3.37\\
\midrule
\multicolumn{8}{l}{HumanoidMaze Medium-Stitch (30 tasks per difficulty)}\\
24 & 100\% & 6,169 & 299,600 & 29/30 & 29/30 & 10.12 & 30.59\\
32 & 100\% & 2,188 & 107,362 & 30/30 & 29/30 & 4.51 & 11.43\\
40 & 100\% & 1,043 & 48,778 & 29/30 & 29/30 & 2.85 & 6.00\\
\bottomrule
\end{tabular}
\end{table}

Increasing $H_{\mathrm{TD}}$ produces coarser graphs with fewer nodes and
edges and reduces median planning time in all three settings
(Table~\ref{tab:horizon-sensitivity}). Accepting-lasso counts vary by only
one of the 60 tasks across scales within each setting. All requested
propositions retain connected anchors in every tested condition.

Halving the graph-construction data in Large-Explore reduces the node count
from 2,525 to 1,798 and median planning time from 4.45 to 2.45 s over the
60 tasks. The half-data and full-data default graphs agree on whether an
accepting lasso is found for every task, finding one for 59 of the 60.
For these tasks, the smaller graph and support pool therefore reduce
planning time while preserving the ability to find accepting lassos.

\subsection{Predicate constraints and support subsampling}
\label{app:grounding-sensitivity}

We examine how adding object-state constraints and reducing offline support
affect the availability of connected anchors and predicate grounding time.
Starting from the six Scene configurations
$D_r,D_o,W_r,W_o,C_{CD},C_{CW}$ in Table~\ref{tab:scene-configurations},
we first constrain only the cube's position (three coordinates), then add
the drawer and window positions (five), and finally one button state (six).
The graph, learned representation, and target values and thresholds for
the retained constraints remain fixed. Removing constraints makes some
configurations identical, leaving two, five, and six distinct predicates
at the respective levels. Each distinct predicate is evaluated once per
support subset.

The full support pool contains 2,864 distinct offline states, with at most
four retained per graph node. Using three random seeds, we draw nested
subsets containing 25\% and 50\% of the pool and compare them with the full
pool, evaluated once. Each subset supplies both node support and anchor
candidates. Grounding time covers evaluator initialization, predicate
evaluation, joint-witness retrieval, and anchor insertion.

\begin{table}[H]
\centering\small
\caption{Scene predicate grounding as object constraints and support size vary.
Coordinates gives the number of constrained object-state coordinates.
Connected reports how many of the six configurations have a connected
anchor under those constraints.
Matching-state counts and grounding times are medians over distinct
predicates. Ranges give the minimum and maximum of these medians across
the three subset-sampling seeds.}
\label{tab:scene-grounding}
\begin{tabular}{rrrrrr}
\toprule
Coordinates & Support & Predicates & Connected & Matching states & Time (s)\\
\midrule
3 & 25\% & 2 & 6/6 & 234.0--238.5 & 2.33--2.51\\
3 & 50\% & 2 & 6/6 & 462.5--474.0 & 4.56--5.26\\
3 & 100\% & 2 & 6/6 & 939.0 & 9.30\\
\midrule
5 & 25\% & 5 & 6/6 & 33.0--46.0 & 4.48--4.73\\
5 & 50\% & 5 & 6/6 & 71.0--82.0 & 8.64--9.01\\
5 & 100\% & 5 & 6/6 & 149.0 & 17.54\\
\midrule
6 & 25\% & 6 & 6/6 & 27.0--30.5 & 6.58--6.81\\
6 & 50\% & 6 & 6/6 & 58.5--67.5 & 12.78--13.23\\
6 & 100\% & 6 & 6/6 & 134.5 & 25.93\\
\bottomrule
\end{tabular}
\end{table}

More specific predicates have fewer matching support states and take longer
to ground (Table~\ref{tab:scene-grounding}). With full support, median
grounding time rises from 9.30 s for cube position alone to 17.54 s with
drawer and window constraints, and to 25.93 s when a button state is added.

Reducing support lowers grounding time at every constraint level.
For the six-coordinate predicates, the median falls from 25.93 s at full
support to 6.58--6.81 s at 25\% support. All six configurations retain
connected anchors at this lowest support level for each of the three seeds.

\FloatBarrier

\section{Planning time and search complexity}\label{app:search-scale}
We report task-time planning costs for navigation and Scene, then examine
the graph and automaton sizes that determine the product search space.
Controlled comparisons on fixed tasks measure the effects of search
strategy and candidate budget.

\subsection{Task-time planning}
\label{app:navigation-planning-time}
\paragraph{Navigation.}
We measure task-time planning as the combined time spent on semantic graph
augmentation and product search before execution.
Table~\ref{tab:sagas-planning-time-summary} summarizes this time for tasks
that SAGAS successfully executes, grouped by domain, task difficulty, and
maze size.

\begin{table}[H]
\centering
\small
\caption{Navigation task-time planning in seconds (mean$\pm$standard deviation over successful SAGAS executions). Times include semantic graph augmentation and product search; the Medium, Large, and Giant columns denote maze size.}
\label{tab:sagas-planning-time-summary}
\begin{tabular}{llrrr}
\toprule
Domain & Difficulty & Medium & Large & Giant \\
\midrule
AntMaze & Easy & $1.67\pm2.04$ & $1.86\pm2.45$ & $1.40\pm1.85$ \\
 & Medium & $1.87\pm1.96$ & $1.82\pm2.05$ & $1.58\pm1.94$ \\
 & Hard & $6.10\pm10.99$ & $11.84\pm27.70$ & $6.81\pm14.29$ \\
\midrule
HumanoidMaze & Easy & $4.88\pm7.36$ & $6.73\pm9.44$ & $11.45\pm19.57$ \\
 & Medium & $6.53\pm6.78$ & $6.90\pm10.13$ & $12.23\pm11.72$ \\
 & Hard & $20.05\pm39.82$ & $26.04\pm55.74$ & $52.58\pm214.41$ \\
\bottomrule
\end{tabular}
\end{table}

Hard tasks take longer to plan and show greater variation in planning time
than Easy and Medium tasks in both domains.
HumanoidMaze has longer average planning times than AntMaze at every
difficulty and maze size, with the highest mean on Giant Hard ($52.58$ s).

\paragraph{Scene.}
Table~\ref{tab:scene-timing} summarizes planning time over all 120 Scene
instances under the protocol in Appendix~\ref{app:scene-protocol}.
Semantic graph augmentation accounts for most of the computation, while
product search takes a median of 0.51 s. Appendix~\ref{app:grounding-sensitivity}
examines how predicate constraints and support size affect grounding time.
\begin{table}[H]
\centering\small
\caption{Scene task-time planning over all 120 instances. Times are in seconds;
p95 denotes the 95th percentile.}
\label{tab:scene-timing}
\begin{tabular}{lrrr}
\toprule
Stage & Median & p95 & Maximum\\
\midrule
Semantic graph augmentation & 39.61 & 54.07 & 55.34\\
Product search & 0.51 & 1.75 & 1.99\\
Total planning & 40.63 & 55.38 & 57.09\\
\bottomrule
\end{tabular}
\end{table}

\subsection{Offline graph and automaton sizes}
\label{app:graph-automaton-sizes}
Each navigation setting uses one offline graph for all 300 tasks, while each
task formula is translated into its own B\"uchi automaton. We report their
sizes separately to characterize the two components of product planning.

The 14 navigation graphs contain 944--7,950 nodes and 7,522--349,626 directed
edges, with one connected component per graph
(Table~\ref{tab:graph-scale}). The table also reports estimated dataset
sizes, compression ratios, and the number of original states retained per
node for semantic evaluation. Compression compares the estimated number of
dataset states with the number of graph nodes; retained support determines
the states available for task-time predicate evaluation.
\begin{table}[H]
\centering\footnotesize
\caption{Frozen navigation graphs. Dataset scale is estimated as
$\widehat N_{\mathcal D}=1+i_{\max}$, where $i_{\max}$ is the largest stored
zero-based dataset index. Compression is $\widehat N_{\mathcal D}/|\mathcal V|$.
Support entries give the median and 95th percentile of preserved states per node. All graphs have one
connected component; each symmetric connection is counted as two directed edges.}
\label{tab:graph-scale}
\begin{tabular}{lrrrrr}
\toprule
Setting & Est. states & Nodes & Edges & Compression & Support\\
\midrule
\multicolumn{6}{l}{AntMaze (29D state)}\\
Giant-Navigate & 1,000,000 & 978 & 7,522 & 1,022$\times$ & 27 / 244.8\\
Giant-Stitch & 1,000,000 & 2,080 & 35,900 & 481$\times$ & 22 / 123.0\\
Large-Explore & 5,000,000 & 2,514 & 34,390 & 1,989$\times$ & 18.5 / 129.0\\
Large-Navigate & 1,000,000 & 1,092 & 12,310 & 916$\times$ & 22 / 196.9\\
Large-Stitch & 1,000,000 & 1,563 & 29,166 & 640$\times$ & 24 / 169.0\\
Medium-Explore & 5,000,000 & 1,879 & 25,668 & 2,661$\times$ & 17 / 193.1\\
Medium-Navigate & 1,000,000 & 944 & 12,792 & 1,059$\times$ & 21 / 237.1\\
Medium-Stitch & 1,000,000 & 1,184 & 23,190 & 845$\times$ & 25 / 257.4\\
\midrule
\multicolumn{6}{l}{HumanoidMaze (69D state)}\\
Giant-Navigate & 4,000,000 & 5,982 & 198,296 & 669$\times$ & 61 / 415.8\\
Giant-Stitch & 4,000,000 & 7,950 & 349,626 & 503$\times$ & 83 / 428.0\\
Large-Navigate & 2,000,000 & 2,175 & 61,162 & 920$\times$ & 28 / 133.3\\
Large-Stitch & 2,000,000 & 3,185 & 138,408 & 628$\times$ & 50 / 214.0\\
Medium-Navigate & 2,000,000 & 1,771 & 58,180 & 1,129$\times$ & 33 / 115.0\\
Medium-Stitch & 2,000,000 & 2,190 & 107,818 & 913$\times$ & 65 / 289.0\\
\bottomrule
\end{tabular}
\end{table}

For the automata, we translate all 4,200 formulas with \texttt{ltl2ba} and
count states and transitions before support-aware preprocessing.
Each conjunctive term in a DNF guard is counted as a separate transition,
matching the guard terms evaluated by the product successor.
Table~\ref{tab:automaton-scale} reports these sizes by task difficulty and
across the full benchmark.
\begin{table}[H]
\centering\small
\caption{Automaton sizes before support-aware preprocessing for 4,200 formulas,
with 1,400 per difficulty. Entries are median / p95 / maximum; transitions
count guard-normalized DNF terms.}
\label{tab:automaton-scale}
\begin{tabular}{lrr}
\toprule
Difficulty & NBA states & DNF transitions\\
\midrule
Easy & 5 / 32 / 32 & 15 / 243 / 243\\
Medium & 6 / 29 / 33 & 19 / 243 / 340\\
Hard & 16 / 217 / 1,520 & 81 / 5,265 / 80,379\\
\midrule
All & 7 / 90 / 1,520 & 27 / 1,299 / 80,379\\
\bottomrule
\end{tabular}
\end{table}
Hard tasks have larger median automata and a longer tail of large instances,
while Easy and Medium tasks have similar sizes. The difficulty categories
are defined by template composition and proposition count, with logical
simplification performed during translation, so their automaton-size ranges
overlap.

At task time, support-aware preprocessing removes unsupported guards and
irrelevant states. The reachable product then depends on these retained
automaton transitions, the semantic graph, and the initial state. The next
section measures this task-specific product directly.

\subsection{Product-graph sizes and reachability}
\label{app:reachable-products}
To measure the search space induced by each task, we explicitly construct
the reachable product for 120 fixed tasks. These come from AntMaze
Giant-Stitch and Large-Explore, and HumanoidMaze Giant-Stitch and
Medium-Stitch. Each setting contributes ten tasks at each of the three
difficulty levels, giving 40 tasks per difficulty.

All counts use the task-specific semantic graph and the automaton after
support-aware preprocessing. The Cartesian state count
$|\mathcal V_\phi||\widetilde Q|$ pairs every semantic-graph node with every
retained automaton state and gives an upper bound on product size.
To obtain the reachable subgraph, we start from the initial product states
and exhaustively follow the movement transitions allowed by
Eq.~\eqref{eq:product-transition-main}. Initialization follows the navigation
protocol in Appendix~\ref{app:evaluation-metrics}. Accepting self-loops at a
fixed waypoint are tested separately as dwell suffixes and are excluded
from the movement-edge counts.
\begin{table}[H]
\centering\scriptsize
\caption{Product sizes over 40 tasks per difficulty. Entries are median / p95 /
maximum. Cartesian states gives $|\mathcal V_\phi||\widetilde Q|$ for the
augmented semantic graph and support-restricted automaton; reachable states
and edges describe the subgraph reached from the initial product states.}
\label{tab:product-scale-full}
\begin{tabular}{lrrr}
\toprule
Difficulty & Cartesian states & Reachable states & Reachable edges\\
\midrule
Easy & 12,630 / 83,251 / 159,240 & 12,600 / 81,659 / 127,392 & 437,414 / 2,211,939 / 5,618,768\\
Medium & 16,318 / 74,047 / 159,240 & 14,227 / 72,031 / 103,507 & 415,586 / 3,173,433 / 4,557,727\\
Hard & 33,534 / 262,726 / 414,336 & 23,178 / 256,878 / 378,726 & 805,823 / 11,338,466 / 18,863,198\\
\bottomrule
\end{tabular}
\end{table}
Hard tasks produce larger reachable products in both state and edge counts,
while Easy and Medium have similar sizes (Table~\ref{tab:product-scale-full}).
For each task, we also divide its reachable-state count by its Cartesian
state count. The median of this fraction is $99.90\%$ for Easy, $77.89\%$ for
Medium, and $79.85\%$ for Hard, showing how much of the Cartesian product
remains reachable under the task's transition constraints.

The semantic graphs themselves have comparable sizes across the three
groups: median/p95 node counts are $2362/7962$, $2361/7962$, and
$2367/7965$ for Easy, Medium, and Hard. The corresponding support-restricted
automata have $5/20$, $5/24$, and $9/90$ states. The Hard group thus combines
similarly sized semantic graphs with larger automata, consistent with its
larger reachable products.

\subsection{Search strategies and candidate budget}\label{app:search-accuracy}
We compare search strategies and prefix-candidate budgets on the 40 Hard
tasks from Appendix~\ref{app:reachable-products}. For each task, all variants
use the same semantic graph, support-restricted automaton, initial product
states, edge costs, and dwell-suffix test. An exhaustive reference computes
the shortest prefix and return suffix for every reachable accepting endpoint
and selects the pair with the lowest weighted lasso cost. It finds an
accepting lasso for 39 tasks and establishes infeasibility for the remaining
task in the fixed finite product.

\paragraph{Search strategy.}
We compare explicit-product Dijkstra, on-demand Dijkstra, and on-demand A*,
all with $K=5$. Explicit-product Dijkstra constructs the entire
start-reachable product before searching. The on-demand variants generate
successors as needed, and their comparison measures the effect of adding
the A* heuristic.

\begin{table}[H]
\centering\small
\caption{Controlled search comparison on 40 Hard tasks. Times cover product
construction and search, in seconds.}
\label{tab:search-variants}
\begin{tabular}{lrr}
\toprule
Variant & Median (s) & p95 (s)\\
\midrule
On-demand A*, $K=5$ & 4.28 & 58.22\\
On-demand Dijkstra, $K=5$ & 4.58 & 62.73\\
Explicit-product Dijkstra, $K=5$ & 5.02 & 74.32\\
Exhaustive accepting-endpoint reference & 5.33 & 77.59\\
\bottomrule
\end{tabular}
\end{table}

On-demand Dijkstra reduces both median and p95 runtime relative to
explicit-product Dijkstra, and on-demand A* reduces them further
(Table~\ref{tab:search-variants}).

\paragraph{Candidate budget.}
The budget $K$ controls how many accepting prefix endpoints receive a
suffix search. We vary $K$ in the on-demand A* planner and compare the
resulting lasso costs and feasibility outcomes with the exhaustive reference.
\begin{table}[H]
\centering\small
\caption{Candidate-budget sensitivity for on-demand A*. Agreement is with
the exhaustive reference: feasibility is evaluated on all 40 tasks,
including the one infeasible task, and exact cost on the 39 feasible tasks.}
\label{tab:prefix-budget}
\begin{tabular}{rrrrr}
\toprule
$K$ & Feasibility agreement & Exact-cost agreement & Median time (s) & p95 time (s)\\
\midrule
1 & 37/40 & 34/39 & 2.72 & 24.92\\
3 & 40/40 & 37/39 & 3.53 & 36.84\\
5 & 40/40 & 39/39 & 4.28 & 58.22\\
10 & 40/40 & 39/39 & 4.31 & 108.93\\
\bottomrule
\end{tabular}
\end{table}
At $K=1$, the planner misses three feasible lassos. Increasing the budget
to $K=3$ finds a lasso for every feasible task, although two costs exceed
their exact optima by at most $3.583\%$. Both $K=5$ and $K=10$ match the
reference on feasibility for all 40 tasks and on graph cost for all 39
feasible tasks (Table~\ref{tab:prefix-budget}). The default $K=5$ achieves
this agreement with a lower p95 runtime than $K=10$ (58.22 versus
108.93 s), balancing search effort and solution quality on this task set.

\section{Case studies}
\label{app:case-studies}

\subsection{Product-space planning versus decoupled planning}
\label{app:decoupled-case-study}
\begin{figure}[H]
    \centering
    \begin{subfigure}[c]{0.45\textwidth}
        \centering
        \includegraphics[width=\textwidth]{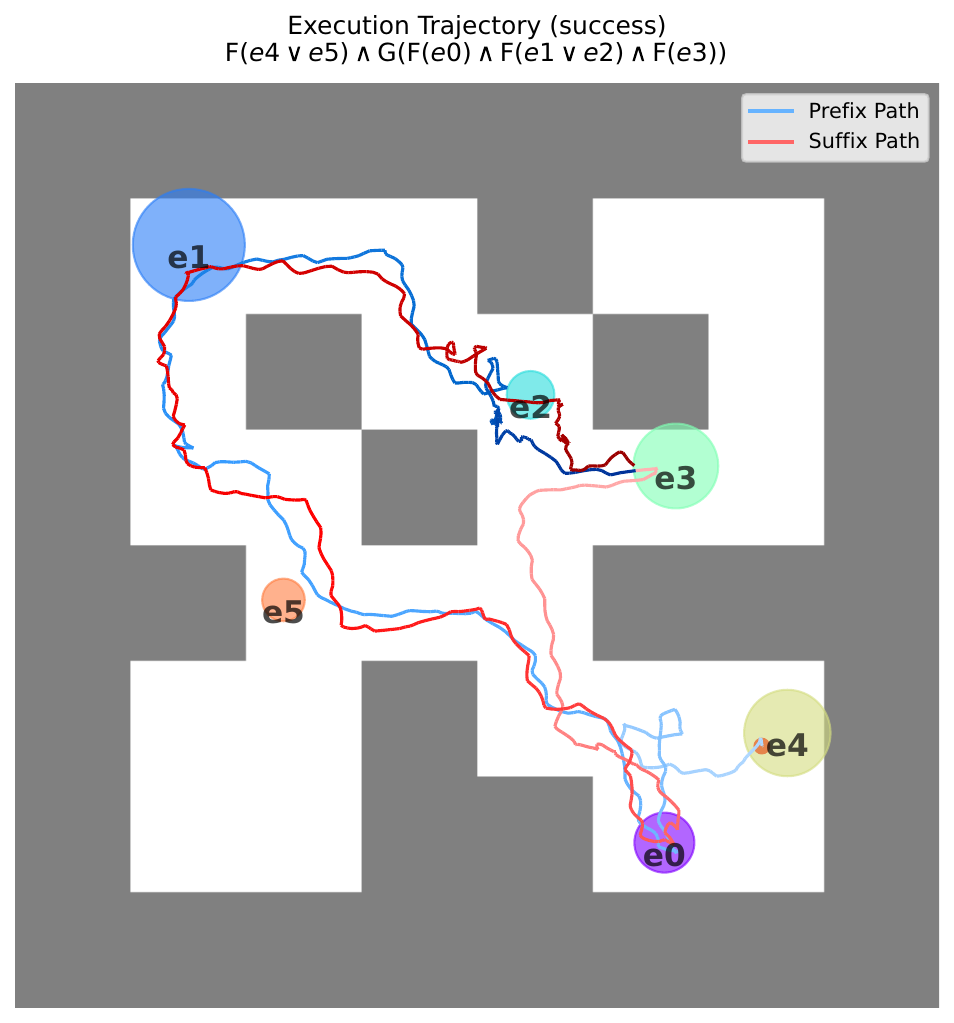}
        \caption{\LFGAS{}}
    \end{subfigure}
    \begin{subfigure}[c]{0.45\textwidth}
        \centering
        \includegraphics[width=\textwidth]{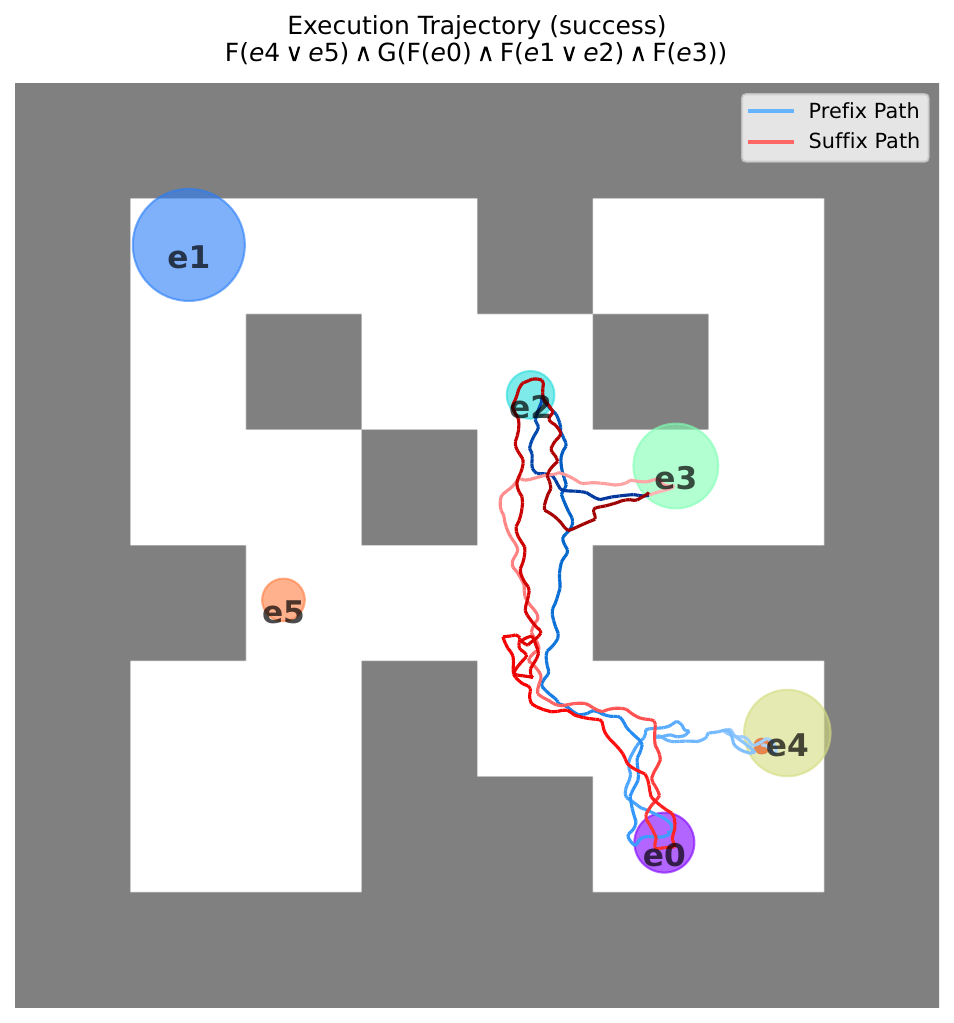}
        \caption{SAGAS}
    \end{subfigure}
    \caption{Executions of the disjunctive task in Eq.~\eqref{eq:app-disjunction-case}. Blue and red curves show the prefix and suffix, respectively.}
    \label{fig:case_2}
\end{figure}

The maze in Figure~\ref{fig:case_2} contains six labeled regions.
The robot must visit either $e_4$ or $e_5$ at least once and repeatedly
visit $e_0$, $e_3$, and either $e_1$ or $e_2$. This task is specified by
\begin{equation}
\label{eq:app-disjunction-case}
\phi_{\mathrm{disj}}
=
\F(e_4 \vee e_5)
\wedge
\G\bigl(\F e_0 \wedge \F(e_1 \vee e_2) \wedge \F e_3\bigr).
\end{equation}
The disjunctions allow several visiting patterns, whose costs depend on
where the selected regions lie in the maze. SAGAS and \LFGAS{} use the same
semantic graph and executor but choose these patterns differently.
\LFGAS{} first selects an accepting automaton sequence with few logical
transitions, then connects its witnesses by shortest graph paths.
SAGAS selects the witnesses and visiting order jointly through cost-aware
product search.

In Figure~\ref{fig:case_2}, \LFGAS{} takes a detour through the upper-left
region $e_1$. SAGAS instead uses $e_2$ for the recurrent disjunction,
yielding a shorter route between the patrol targets. This example shows
why the cost of reaching a region matters when choosing among logically
valid alternatives.

\subsection{Generalization across predicate geometries}
\label{app:predicate-case-studies}
\begin{figure}[htbp]
    \centering
    \begin{subfigure}[t]{0.47\textwidth}
        \centering
        \includegraphics[width=\textwidth]{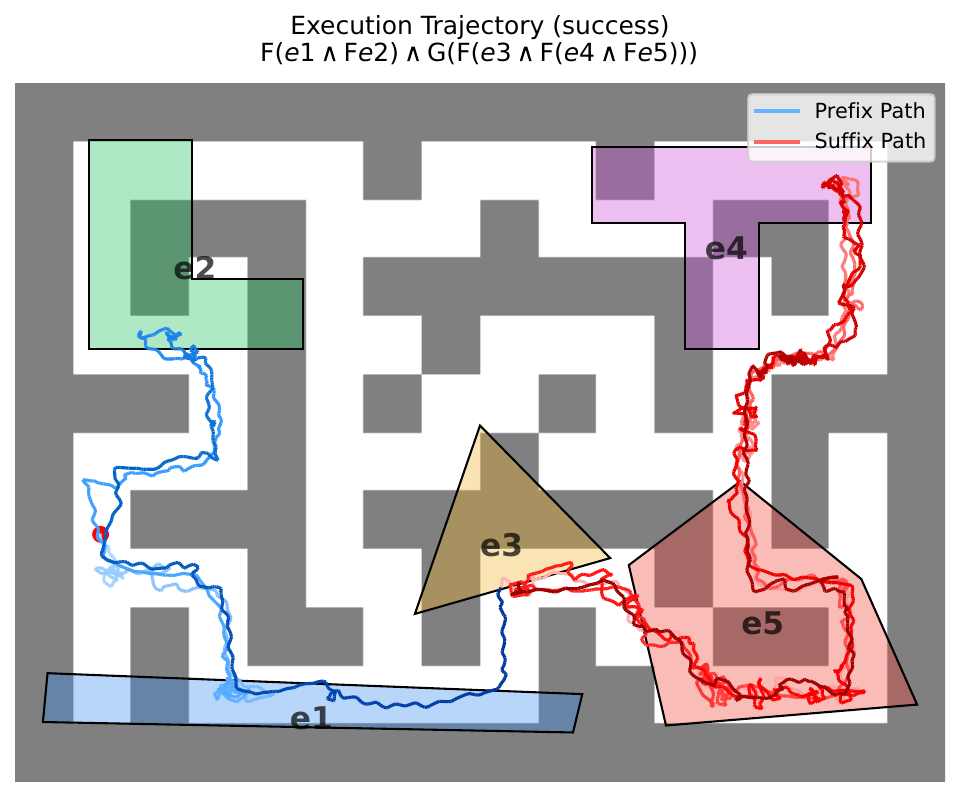}
        \caption{Ordered visits followed by a three-region patrol ($\phi_{\mathrm{rec}}$).}
    \end{subfigure}
    \hfill
    \begin{subfigure}[t]{0.47\textwidth}
        \centering
        \includegraphics[width=\textwidth]{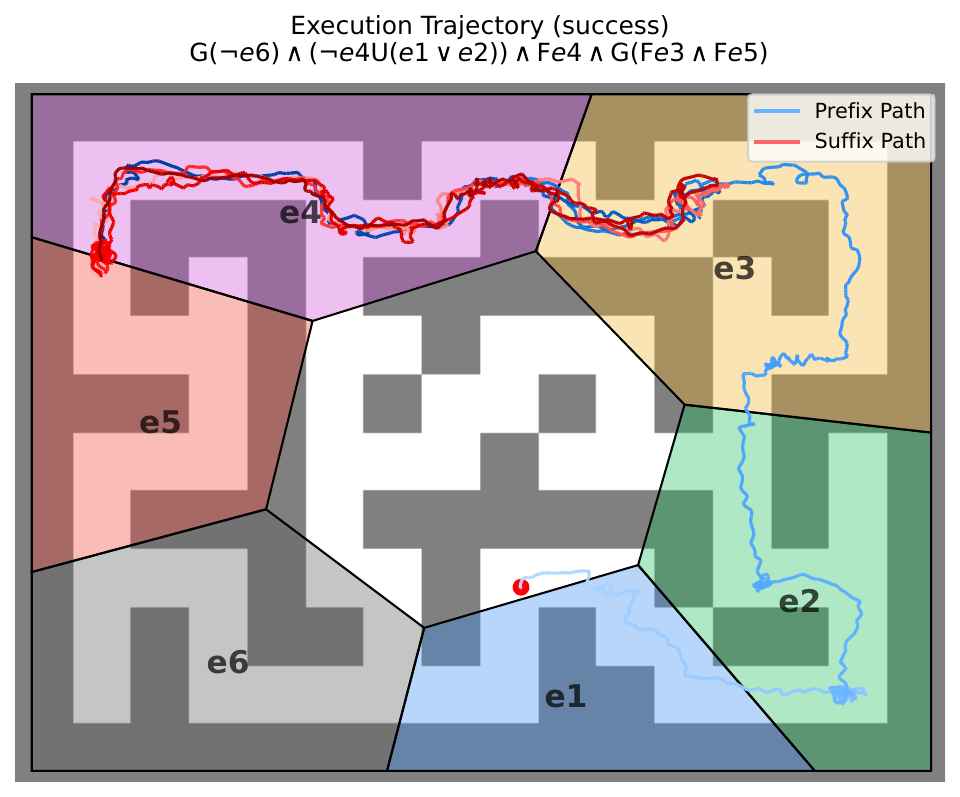}
        \caption{Conditional entry into $e_4$ and global avoidance of $e_6$ ($\phi_{\mathrm{neg}}$).}
    \end{subfigure}

    \vspace{0.6em}
    \begin{subfigure}[t]{0.47\textwidth}
        \centering
        \includegraphics[width=\textwidth]{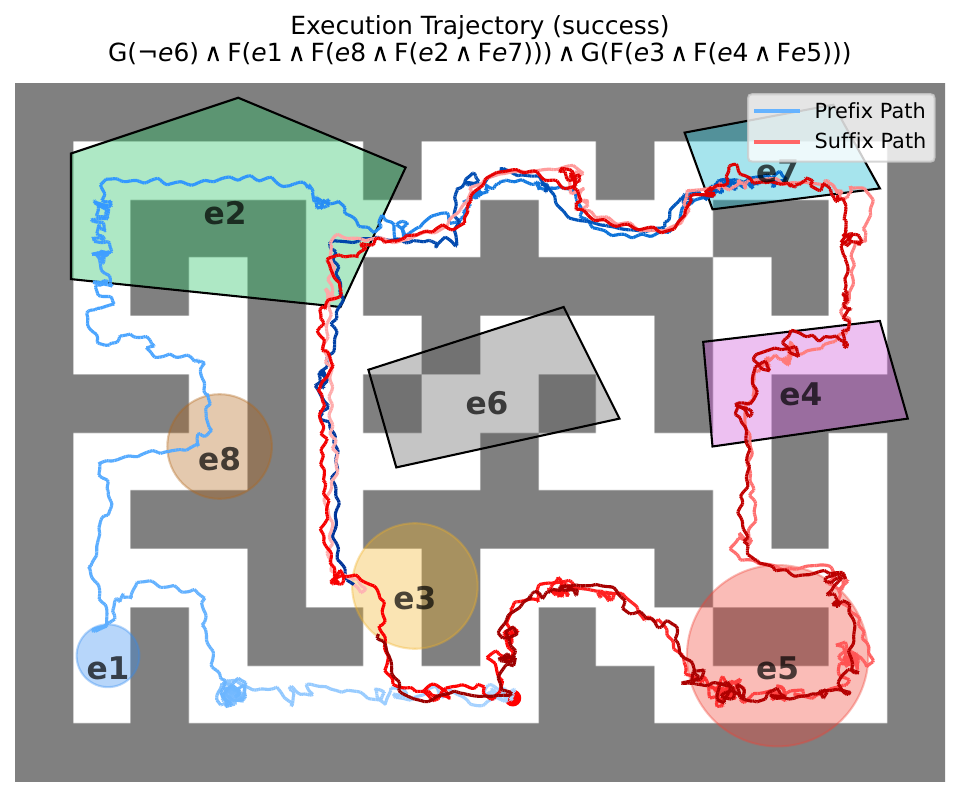}
        \caption{A four-stage prefix and recurrent patrol under an avoidance constraint ($\phi_{\mathrm{mix}}$).}
    \end{subfigure}
    \hfill
    \begin{subfigure}[t]{0.47\textwidth}
        \centering
        \includegraphics[width=\textwidth]{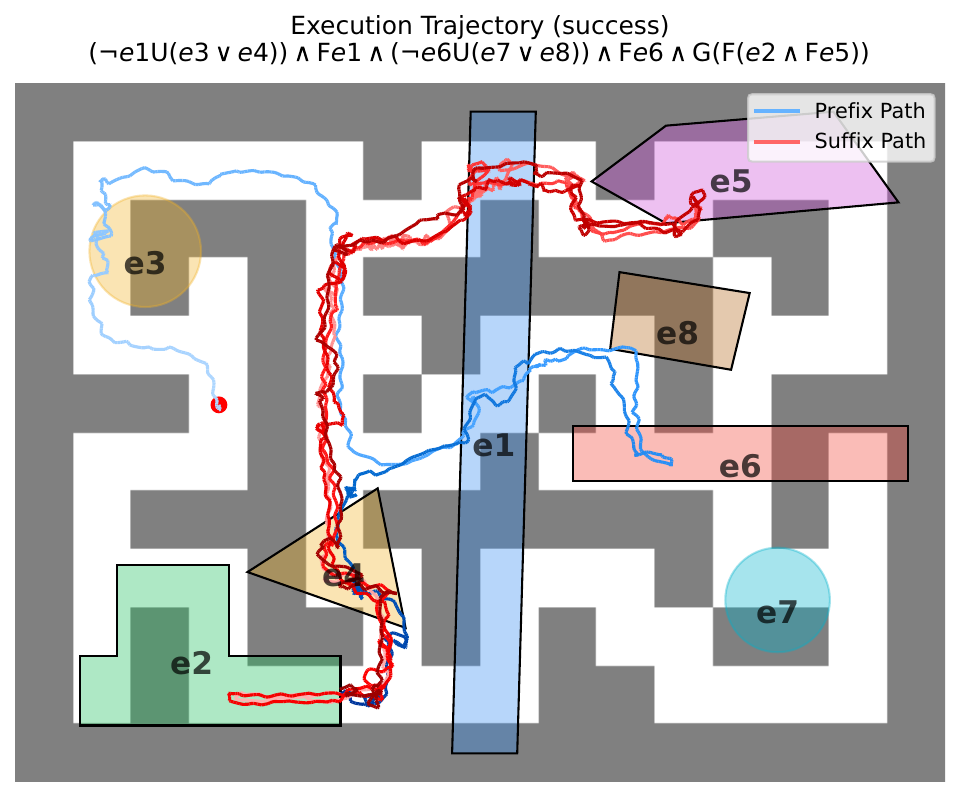}
        \caption{Two conditionally accessible strips and a two-region patrol ($\phi_{\mathrm{mode}}$).}
    \end{subfigure}
    \caption{Executions with heterogeneous predicate geometries for the four formulas in Eq.~\eqref{eq:app-case-study-formulas}. Blue and red curves show the prefix and suffix, respectively. Each execution satisfies the finite-lasso criterion.}
    \label{fig:app-mixed-case-studies}
\end{figure}

The four maze tasks in Figure~\ref{fig:app-mixed-case-studies} use new
labeling functions with the same offline graph and frozen executor.
Their regions vary in shape, size, and location, including circles,
triangles, elongated strips, and non-convex polygons. Some regions are
specified broadly enough to include walls or areas with sparse offline
coverage. By selecting graph-connected anchors from predicate-satisfying
offline states, SAGAS identifies data-supported targets within these
regions, allowing users to specify predicates without precisely delineating
their reachable portions.
The tasks combine these geometries with the following temporal requirements:

\begin{equation}
\begin{aligned}
\phi_{\mathrm{rec}}
&= \F(e_1 \land \F e_2)
   \land \G\F(e_3 \land \F(e_4 \land \F e_5)),\\
\phi_{\mathrm{neg}}
&= \G\neg e_6
   \land (\neg e_4 \,\U\, (e_1 \lor e_2))
   \land \F e_4
   \land \G(\F e_3 \land \F e_5),\\
\phi_{\mathrm{mix}}
&= \G\neg e_6
   \land \F(e_1 \land \F(e_8 \land \F(e_2 \land \F e_7)))
   \land \G\F(e_3 \land \F(e_4 \land \F e_5)),\\
\phi_{\mathrm{mode}}
&= (\neg e_1 \,\U\, (e_3 \lor e_4))
   \land \F e_1
   \land (\neg e_6 \,\U\, (e_7 \lor e_8))\\
&\quad \land \F e_6
   \land \G\F(e_2 \land \F e_5).
\end{aligned}
\label{eq:app-case-study-formulas}
\end{equation}

In Figure~\ref{fig:app-mixed-case-studies}(a), five polygonal regions define
two ordered visits followed by a recurrent patrol. For $\phi_{\mathrm{rec}}$,
the robot first visits the strip $e_1$ and then $e_2$ on the left of the
maze. It then travels to the right-hand regions and follows the
$e_3\rightarrow e_4\rightarrow e_5$ suffix loop.

The six sectors in Figure~\ref{fig:app-mixed-case-studies}(b) surround an
unlabeled starting area. Under $\phi_{\mathrm{neg}}$, $e_6$ is always
forbidden, while entry into $e_4$ becomes permissible once $e_1$ or $e_2$
has been reached. The robot must also visit $e_4$ and repeatedly return to
$e_3$ and $e_5$. The shown trajectory enters $e_2$ before $e_4$, then
patrols between $e_3$ and $e_5$ while avoiding $e_6$.

Figure~\ref{fig:app-mixed-case-studies}(c) combines eight regions in a task
with a longer ordered prefix. In the shown execution of $\phi_{\mathrm{mix}}$, the robot
visits $e_1$, $e_8$, $e_2$, and $e_7$ in sequence before entering the
$e_3\rightarrow e_4\rightarrow e_5$ patrol. The forbidden region $e_6$
must be avoided throughout both phases.

In Figure~\ref{fig:app-mixed-case-studies}(d), two elongated strips have
independent entry conditions. The task $\phi_{\mathrm{mode}}$ allows entry
into the vertical strip $e_1$ once $e_3$ or $e_4$ has been visited, and
into the horizontal strip $e_6$ once $e_7$ or $e_8$ has been visited.
The trajectory meets both conditions, visits the strips, and then
patrols between $e_2$ and $e_5$.

Across these cases, SAGAS satisfies different visiting orders and avoidance
requirements over new predicate geometries using the same learned graph
and executor.

\end{document}